\documentclass[pre,twocolumn,showpacs,superscriptaddress,preprintnumbers,floatfix]{revtex4-1}

\usepackage{etex}
\usepackage{ifpdf}
\usepackage[final,pagebackref]{hyperref}
\usepackage{dcolumn}
\usepackage{url}
\usepackage{xkeyval}
\usepackage{amsfonts}
\usepackage{amsmath}
\usepackage{amssymb}
\usepackage{amsthm}
\usepackage{mathtools}
\usepackage{etoolbox}
\usepackage{array}
\usepackage{bm}   % bold math
\usepackage{bbm}
\usepackage{verbatim}
\usepackage{xcolor}
\usepackage{setspace}
\usepackage{graphicx}
\usepackage{pgf}
\usepackage{tikz}
\usetikzlibrary{arrows,shapes,automata}
\theoremstyle{plain}
\theoremstyle{plain}
\theoremstyle{plain}
\theoremstyle{plain}
\theoremstyle{plain}
\theoremstyle{plain}
\theoremstyle{plain}\newtheorem{Def}{Definition}
\theoremstyle{plain}


\def\clap#1{\hbox to 0pt{\hss#1\hss}}

\def\mathrlap{\mathpalette\mathrlapinternal}

\def\mathrlapinternal#1#2{%
\rlap{$\mathsurround=0pt#1{#2}$}}

\newenvironment{itemize*}%
  {\begin{itemize}%
    \setlength{\itemsep}{0pt}%
    \setlength{\parskip}{0pt}%
	\setlength{\topsep}{0pt}%
	\setlength{\partopsep}{0pt}%
	\setlength{\parsep}{0pt}}%
  {\end{itemize}}

\newenvironment{enumerate*}%
  {\begin{enumerate}%
    \setlength{\itemsep}{0pt}%
    \setlength{\parskip}{0pt}%
	\setlength{\topsep}{0pt}%
	\setlength{\partopsep}{0pt}%
	\setlength{\parsep}{0pt}}%
  {\end{enumerate}}

\DeclareMathOperator*{\argmax}{argmax}

\newcommand{\eM}     {\mbox{$\epsilon$-machine}}
\newcommand{\eMs}    {\mbox{$\epsilon$-machines}}

\newcommand{\EMs}    {\mbox{$\epsilon$-Machines}}

\newcommand{\MeasAlphabet}	{\mathcal{X}}

\newcommand{\AllWords}[1][]{%
  \ifblank{#1}%
    {\MeasAlphabet^*}%
    {\ifstrequal{#1}{*}%
      {\MeasAlphabet^*}% #1 is *
      {\MeasAlphabet^{\bar{#1}}}% #1 is not *
    }%
}

\newcommand{\MeasSymbol}   { {X} }
\newcommand{\meassymbol}   { {x} }

\newcommand{\CausalState}	{ \mathcal{S} }

\newcommand{\causalstate}	{ \sigma }

\newcommand{\AlternateState}	{ R }
\newcommand{\Cmu}		{C_\mu}
\newcommand{\hmu}		{h_\mu}

\newcommand{\forward}{+}
\newcommand{\reverse}{-}
\newcommand{\forwardreverse}{\pm} % \pm

\newcommand{\FutureCausalState}	{ {\CausalState}^{\forward} }

\newcommand{\PastCausalState}	{ {\CausalState}^{\reverse} }

\newcommand{\lastindex}[2]{
  \edef\tempa{0}
  \edef\tempb{#2}
  \ifx\tempa\tempb
    \edef\tempc{#1}
  \else
    \edef\tempa{0}
    \edef\tempb{#1}
    \ifx\tempa\tempb
      \edef\tempc{#2}
    \else
      \edef\tempc{#1+#2}
    \fi
  \fi
  \tempc
}

\newcommand{\CSjoint}[1][,]{
   \edef\tempa{:}
   \edef\tempb{#1}
   \ifx\tempa\tempb
      \ensuremath{\FutureCausalState\!#1\PastCausalState}
   \else
      \ensuremath{\FutureCausalState#1\PastCausalState}
   \fi
}

\newif\ifpm 
\edef\tempa{\forwardreverse}
\edef\tempb{\pm}
\ifx\tempa\tempb
   \pmfalse
\else
   \pmtrue  
\fi

\renewcommand{\Pr}{\mathbb{P}}

\MHInternalSyntaxOn
\providecommand*\phantomword[3][c]{%
  \mathchoice
  {\MT_phantom_word:NNnn #1\displaystyle {#2}{#3}}%
  {\MT_phantom_word:NNnn #1\textstyle {#2}{#3}}%
  {\MT_phantom_word:NNnn #1\scriptstyle {#2}{#3}}%
  {\MT_phantom_word:NNnn #1\scriptscriptstyle {#2}{#3}}%
}
\def\MT_phantom_word:NNnn #1#2#3#4{%
  \@begin@tempboxa\hbox{$\m@th#2#4$}%
    \setlength\@tempdima{\widthof{$\m@th#2#3$}}%
    \hbox{\hb@xt@\@tempdima{\csname bm@#1\endcsname}}%
  \@end@tempboxa}
\MHInternalSyntaxOff

\usepackage{xkeyval}
\usepackage{etoolbox}

\makeatletter
\newlength\CMlength
\define@cmdkey[CM]{vector}[CM@@]{raise}{}
\define@cmdkey[CM]{vector}[CM@@]{pre}{}
\define@cmdkey[CM]{vector}[CM@@]{post}{}
\define@cmdkey[CM]{vector}[CM@@]{symbol}{}
\presetkeys[CM]{vector}{%
  raise=1.02,%
  pre={\scriptscriptstyle\,},%
  post={}%
}{}
\newcommand{\CMvector}[2][]{%
  \setkeys[CM]{vector}[]{#1}%
  \settoheight{\CMlength}{\ensuremath{#2}}%
  \setlength{\CMlength}{\CM@@raise\CMlength}%
  \ensuremath{%
    \mathrlap{\ensuremath{#2}}%
    \smash{\phantomword[c]%
      {\ensuremath{#2}}%
      {\raise \CMlength \hbox{\ensuremath{\CM@@pre\CM@@symbol\CM@@post}}}%
    }%
  }%
}
\makeatother

\newcommand{\leftrightharpoonup}{\mathrlap{\leftharpoonup}\phantomword[l]{\,\leftharpoonup}{\,\rightharpoonup}}

\newcommand{\CMlharpoon}[1]{\CMvector[symbol=\leftharpoonup,pre=\scriptscriptstyle]{#1}}
\newcommand{\CMrharpoon}[1]{\CMvector[symbol=\rightharpoonup,pre=\scriptscriptstyle]{#1}}
\newcommand{\CMlrharpoon}[1]{\CMvector[symbol=\leftrightharpoonup,pre=\scriptscriptstyle]{#1}}

\newcommand{\processfirst}[1]{#1\expandafter\processnext}
\newcommand{\processnext}[1]{%
  \ifx\listfinish#1\empty\else\listact{#1}\expandafter\processnext\fi}

\makeatletter
\def\BEsep{:}
\newcommand{\CMIndexedSymbol}[2]{%
  \protected\expandafter\def\csname #1\endcsname{%
    \@ifnextchar({\csname #1@i\endcsname}%
                 {\csname #1@i\endcsname()}%
  }%
  \expandafter\def\csname #1@i\endcsname(##1){%
    \ifstrequal{##1}{>}%
      {\def\CMtemp{\CMrharpoon{#2}}}% (>) --> right
      {% false
        \ifstrequal{##1}{<}%
          {\def\CMtemp{\CMlharpoon{#2}}}% (<) --> left
          {%
            \ifstrequal{##1}{<>}%
              {\def\CMtemp{\CMlrharpoon{#2}}}% (<>) --> left,right
              {\def\CMtemp{#2}}% () or anything else --> regular
          }%
      }%  
    \@ifnextchar[{\csname #1@ii\endcsname(\CMtemp)}%
                 {\CMtemp} %                                   \cmdname
  }%
  \expandafter\def\csname #1@ii\endcsname(##1)[##2]{%
    \@ifnextchar[{\csname #1@iii\endcsname({##1})[{##2}]}%
                 {##1_{##2}}%                                  \cmdname[i]
  }%
  \expandafter\def\csname #1@iii\endcsname(##1)[##2][##3]{%
    ##1_{##2\BEsep##3}%                                        \cmdname[i][j]
  }%
}

\newcommand{\CMSuperIndexedSymbol}[3]{%
  \protected\expandafter\def\csname #1\endcsname{%
    \@ifnextchar({\csname #1@i\endcsname}%
                 {\csname #1@i\endcsname()}%
  }%
  \expandafter\def\csname #1@i\endcsname(##1){%
    \ifstrequal{##1}{>}%
      {\def\CMtemp{\CMrharpoon{#2}}}% (>) --> right
      {% false
        \ifstrequal{##1}{<}%
          {\def\CMtemp{\CMlharpoon{#2}}}% (<) --> left
          {%
            \ifstrequal{##1}{<>}%
              {\def\CMtemp{\CMlrharpoon{#2}}}% (<>) --> left,right
              {\def\CMtemp{#2}}% () or anything else --> regular
          }%
      }%  
    \@ifnextchar[{\csname #1@ii\endcsname(\CMtemp)}%
                 {\CMtemp^{#3}} %                                   \cmdname
  }%
  \expandafter\def\csname #1@ii\endcsname(##1)[##2]{%
    \@ifnextchar[{\csname #1@iii\endcsname({##1})[{##2}]}%
                 {##1_{##2}^{#3}}%                                  \cmdname[i]
  }%
  \expandafter\def\csname #1@iii\endcsname(##1)[##2][##3]{%
    ##1_{##2\BEsep##3}^{#3}%                                        \cmdname[i][j]
  }%
}

\CMIndexedSymbol{MS}{\MeasSymbol}
\CMIndexedSymbol{ms}{\meassymbol}
\CMIndexedSymbol{CS}{\CausalState}
\CMIndexedSymbol{cs}{\causalstate}
\CMSuperIndexedSymbol{FCS}{\CausalState}{+}
\CMSuperIndexedSymbol{RCS}{\CausalState}{-}
\CMIndexedSymbol{AS}{\AlternateState}
\CMSuperIndexedSymbol{ASPrime}{\AlternateState}{\prime}

\newcommand{\arxiv}[1]{\href{http://arxiv.org/abs/#1}{\texttt{arXiv}:#1}}
\newcommand{\sfiwp}[1]{Santa Fe Institute Working Paper #1}

\newcommand{\HMM}{hidden Markov model}

\newcommand{\figref}[1]{Fig.~\ref{#1}}
\newcommand{\eqnref}[1]{Eq.~\eqref{#1}}

\newcommand{\Mi}{M_{i}}              % model topology i
\newcommand{\Mset}{\boldsymbol{\mathcal{M}}}   % set of machine topologies
\newcommand{\hiddenstate}{\causalstate}
\newcommand{\HiddenStateSet}	{ \boldsymbol{\CausalState} }
\usepackage{listings}
\lstdefinestyle{mypython}{
language=Python,                        % Code langugage
basicstyle=\small\ttfamily,             % Code font, Examples: \footnotesize, \ttfamily
keywordstyle=\color{green!50!black},    % Keywords font ('*' = uppercase)
commentstyle=\color{gray},              % Comments font
numbers=left,                           % Line nums position
numberstyle=\tiny,                      % Line-numbers fonts
stepnumber=1,                           % Step between two line-numbers
numbersep=5pt,                          % How far are line-numbers from code
backgroundcolor=\color{gray!10},        % Choose background color
frame=none,                             % A frame around the code
tabsize=2,                              % Default tab size
captionpos=b,                           % Caption-position = bottom
breaklines=true,                        % Automatic line breaking?
breakatwhitespace=false,                % Automatic breaks only at whitespace?
showspaces=false,                       % Dont make spaces visible
showtabs=false,                         % Dont make tabls visible
morekeywords={as},                      % Additional keywords
}

\begin{document}

\def\ourTitle{Bayesian Structural Inference for Hidden Processes}

\author{Christopher C.~Strelioff}
\email{strelioff@ucdavis.edu}
\affiliation{Complexity Sciences Center and Physics Department,
University of California at Davis, One Shields Avenue, Davis, CA 95616}

\author{James P. Crutchfield}
\email{chaos@ucdavis.edu}
\affiliation{Complexity Sciences Center and Physics Department,
University of California at Davis, One Shields Avenue, Davis, CA 95616}
\affiliation{Santa Fe Institute, 1399 Hyde Park Road, Santa Fe, NM 87501}

% ************************* ABSTRACT *************************
\def\ourAbstract{%
We introduce a Bayesian approach to discovering patterns in structurally complex
processes. The proposed method of Bayesian Structural Inference (BSI) relies on
a set of candidate unifilar \HMM\ (uHMM) topologies for inference of process
structure from a data series. We employ a recently developed exact
enumeration of topological \eMs. (A sequel then removes the topological
restriction.) This subset of the uHMM topologies has the added benefit that
inferred models are guaranteed to be \eMs, irrespective of estimated
transition probabilities. Properties of \eMs\ and uHMMs allow for the derivation
of analytic expressions for estimating transition probabilities, inferring
start states, and comparing the posterior probability of candidate model
topologies, despite process internal structure being only indirectly present in
data. We demonstrate BSI's effectiveness in estimating a process's randomness,
as reflected by the Shannon entropy rate, and its structure, as quantified by
the statistical complexity. We also compare using the posterior distribution
over candidate models and the single, maximum a posteriori model for point
estimation and show that the former more accurately reflects uncertainty in
estimated values. We apply BSI to in-class examples of finite- and
infinite-order Markov processes, as well to an out-of-class, infinite-state
hidden process.
}
% ************************************************************

\def\ourKeywords{%
  stochastic process, hidden Markov model,
  \texorpdfstring{\eM}{epsilon-machine}, causal states
}

\bibliographystyle{unsrt}

\hypersetup{
  pdfauthor={C. C. Strelioff and J. P. Crutchfield},
  pdftitle={\ourTitle},
  pdfsubject={\ourAbstract},
  pdfkeywords={\ourKeywords},
  pdfproducer={},
  pdfcreator={}
}

% Implementation of abstract
\begin{abstract}
\ourAbstract

\vspace{0.1in}
\end{abstract}


\preprint{\sfiwp{13-09-027}}
\preprint{\arxiv{1309.1392 [stat.ML]}}

\title{\ourTitle}
\date{\today}
\maketitle
%\tableofcontents
\setstretch{1.1}

%%
%\noindent
%\paragraph*{Popular Summary} Discovering structural patterns in experimental
%and observational data is
%fundamental to understanding the world around us. Today, we regularly analyze
%nucleotides in {DNA} segments, daily ups and downs in the stock market, and
%glyphs inscribed on ancient stone tablets. A question of primary importance
%in each such attempt: Are the observations random or is there structure in
%the nucleotide sequence, market swings, and pictographs? Pattern detection
%lies at the heart of our ability to predict, control, and diagnose natural
%and engineered systems.
%
%We adapt Bayesian statistical inference to find the, potentially hidden,
%pattern in a string of symbols. To do this, we use a recently developed method
%to exhaustively enumerate candidate structural models and use Bayes' Theorem to
%exactly calculate the most probable structure given the available data and set
%of models considered. Topological \eMs---the models we use---are a type of
%\HMM\ that allows us to capture, on the one hand, randomness via a single-state
%model, as well many different types of structure, on the other, using models
%with multiple states and edges.  In addition, information-theoretic tools allow
%us to quantify randomness and structure using the entropy rate and statistical
%complexity, respectively.
%
%The Bayesian Structure Inference method we introduce applies to any sequence
%of symbols, in space or time, from a finite alphabet. As a result, we expect
%applications in fields ranging from bioinformatics and dynamical systems to
%linguistics and crystallography.

%%
\section{Introduction}

Emergent patterns are a hallmark of complex, adaptive behavior, whether
exhibited by natural or designed systems. Practically, discovering and
quantifying the structures making up emergent patterns from a sequence of
observations lies at the heart of our ability to understand, predict, and
control the world. But, what are the statistical signatures of structure? A
common modeling assumption is that observations are independent and identically
distributed (IID). This is tantamount, though, to assuming a system is
structureless.  And so, pattern discovery depends critically on testing when
the IID assumption is violated.  Said more directly, successful pattern
discovery extracts the (typically hidden) mechanisms that create departures
from IID structurelessness.  In many applications, the search for structure is
made all the more challenging by limited available data. The very real
consequences, when pattern discovery is done incorrectly with finite data, are
that structure can be mistaken for randomness and randomness for structure.

Here, we develop an approach to pattern discovery that removes these confusions,
focusing on data series consisting of a sequence of symbols from a finite
alphabet. That is, we wish to discover temporal patterns, as they occur in
discrete-time and discrete-state time series. (The approach also applies to
spatial data exhibiting one-dimensional patterns.) Inferring structure from
data series of this type is integral to many fields of science ranging from
bioinformatics \cite{Yoon2009,Narlikar_etal2012}, dynamical systems
\cite{Davidchack_etal2000,Daw_etal2003,Kennel_etal2003,Stre07b}, and
linguistics \cite{Rao09,Lee_etal2010} to single-molecule spectroscopy
\cite{Kelly_etal2012, Li_etal2008}, neuroscience \cite{Graben_etal2000,
Haslinger2009}, and crystallography \cite{Varn06c,Varn12a}. Inferred structure
assumes a meaning distinctive to each field. For example, in single molecule
dynamics structure reflects stable molecular configurations, as well as the
rates and types of transition between them. In the study of coarse-grained
dynamical systems and linguistics, structure often reflects forbidden words and
relative frequencies of symbolic strings that make the language or dynamical
system functional.  Thus, the results of successful pattern discovery teach one
much more about a process than models that are only highly predictive.

Our goal is to infer structure using a finite data sample from some process 
of interest and a set of candidate \eM\ model topologies. This choice of model
class is made because \eMs\ provide optimal prediction as well as being a
minimal and unique representation \cite{Crut12a}. In addition, given an
\eM, structure and randomness can be quantified using the statistical complexity
$\Cmu$ and Shannon entropy rate $\hmu$. Previous efforts to infer \eMs\ from
finite data include \emph{subtree merging} (SM) \cite{Crut88a},
\emph{\eM\ spectral reconstruction} ($\epsilon$MSR) \cite{Varn02a}, and
\emph{causal-state splitting reconstruction} (CSSR) \cite{Shal02a,Shal04a}.
These methods produce a single, best-estimate of the appropriate \eM\ given
the available data.

The following develops a distinctively different approach to the problem of
structural inference---\emph{Bayesian Structural Inference} (BSI). BSI requires
a data series $D$ and a set of candidate unifilar hidden Markov model (uHMM)
topologies, which we denote $\Mset$.  However, for our present goal of
introducing BSI, we consider only a subset of unifilar \HMM s---the topological
\eMs---that are guaranteed to be \eMs\ irrespective of estimated transition
probabilities \cite{John10a}. Unlike the inference methods cited above, BSI's
output is not a single best-estimate. Instead, BSI determines the posterior
probability of each model topology conditioned on $D$ and $\Mset$. One result
is that many model topologies are viable candidates for a given data set. The
shorter the data series, the more prominent this effect becomes. We argue, in
this light, that the most careful approach to structural inference and
estimation is to use the complete set of model topologies according to their
posterior probability. Another consequence, familiar in a Bayesian setting, is
that principled estimates of uncertainty---including uncertainty in model
topology---can be straightforwardly obtained from the posterior distribution.

The methods developed here draw from several fields, ranging from computational
mechanics \cite{Crut12a} and dynamical systems \cite{Ott93a,Stro94a,Lind95a} to
methods of Bayesian statistical inference \cite{GelmanBook1995}. As a result,
elements of the following will be unfamiliar to some readers. To
create a bridge, we provide an informal overview of foundational concepts in
Sec.~\ref{sec:informal} before moving to BSI's technical details in Sec.
\ref{sec:StructuredProcess}.

\section{Process Structure, Model Topologies, and Finite Data}
\label{sec:informal}

\begin{figure*}[th]
\centering
\includegraphics[width=\textwidth]{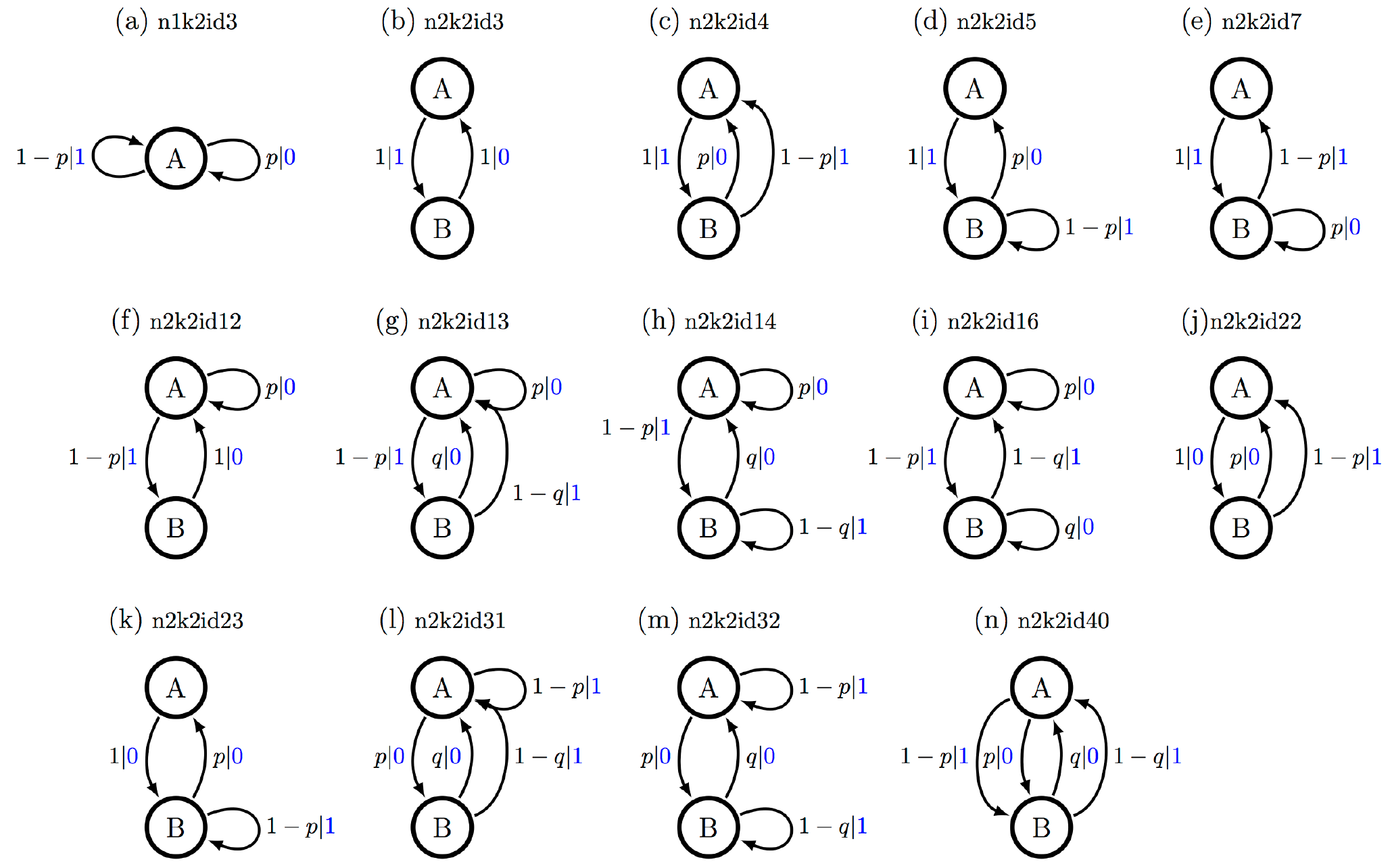}
\caption{All binary, unifilar hidden Markov model topologies with one or two
  states. Each topology, designated (a) through (n), also has a unique label
  that provides the number of states $n=1,2$, the alphabet size $k=2$, and a
  unique \emph{id} that comes from the algorithm used to enumerate all possible
  model topologies \cite{John10a}.  Model edges are labeled with a transition
  probability and output symbol using the format:
  \emph{probability} $\vert$ \emph{symbol}.
  }
\label{fig:uHMM}
\end{figure*}

To start, we offer a nontechnical introduction to structural inference to be
clear how we distinguish (i) a process and its inherent structure from (ii)
model topology and these from (iii) sampled data series. A \emph{process}
represents all possible behaviors of a system of interest. It is the object of
our focus.  Saying that we infer \emph{structure} means we want to find the
process's organization---the internal mechanisms that generate its observed
behavior.  However, in any empirical setting we only have samples of the
process's behavior in the form of finite \emph{data series}. A data series
necessarily provides an incomplete picture of the process due to the finite
nature of the observation. Finally, we use a \emph{model} or, more precisely, a
\emph{model topology} to express the process's structure. The model
topology---the set of states and transitions, their connections and observed
output symbols---explicitly represents the process's structure. Typically,
there are many model topologies that accurately describe the probabilistic
structure of a given process.  \EMs\ are special within the set of accurate models, however, in that they are the model topology that provides the unique
and minimal representation of process structure.

To ground this further, let's graphically survey different model topologies and
consider what processes they represent and how they generate finite data
samples. Figure \ref{fig:uHMM} shows models with one or two states that
generate binary processes---observed behavior is a sequence of $0$s and $1$s.
For example, the smallest model topology is shown in \figref{fig:uHMM}(a) and
represents the IID binary process. This model generates data by starting in
state $A$ and outputs a $0$ with probability $p$ and a $1$ with probability
$1-p$, always returning to state $A$.

A more complex model topology, shown in \figref{fig:uHMM}(g), has two states
and four edges. In this case, when the model is in state $A$ it generates a $0$
with probability $p$ and returns to state $A$, or it generates a $1$ with
probability $1-p$ and moves to state $B$.  When in state $B$, a $0$ is
generated with probability $q$ and $1$ with probability $1-q$, moving to state
$A$ in both cases. If $p \neq q$ this model topology represents a unique,
structured process. However, if $p=q$ the probability of generating a $0$ or
$1$ does not depend on states $A$ and $B$ and the resulting process is IID.
Thus, this model topology with $p=q$ becomes an overly verbose representation
of the IID process, which requires only a single state---the topology of
\figref{fig:uHMM}(a). This setting of the transition probabilities is an
example where a model topology describes the probabilistic behavior of a
process, but does not reflect the structure. In fact, the model topology in
\figref{fig:uHMM}(g) is not an \eM\ when $p=q$. Rather, the process structure
is properly represented by \figref{fig:uHMM}(a), which is.

This example and other cases where specific model topologies are not minimal
and unique representations of a process's structure motivate identifying
a subclass of model topologies. All model topologies in \figref{fig:uHMM} are
unifilar hidden Markov models (defined shortly). However, the six model
topologies with two states and four edges, \figref{fig:uHMM}(g-i, l-n), are not
minimal when $p=q$. As with the previous example, they all become overly
complex representations of the IID process for this parameter setting.
Excluding these uHMMs leaves a subset of topologies called \emph{topological
\eMs}, \figref{fig:uHMM}(a-f,j-k), that are guaranteed to be minimal and unique
representations of process structure for any transition probabilities setting,
other than $0$ or $1$. Partly to emphasize the role of process structure and
partly to simplify technicalities, in this first introduction to BSI we only
consider topological \eMs. A sequel lifts this restriction, adapting BSI to
work with all \eMs.

In this way, we see how a process's structure is expressed in model topology
and how possible ambiguities arise. This is the \emph{forward} problem of
statistical inference. Now, consider the complementary \emph{inverse} problem:
Given an observed data series, find the model topology that most effectively
describes the unknown process structure. In a Bayesian setting, the first step
is to identify those model topologies that can generate the observed data. As
just discussed, we do this by choosing a specific model topology and start
state and attempting to trace the hidden-state path through the model, using
the observed symbols to determine the edges to follow. If there is a path for
at least one start state, the model topology is a viable candidate. This
process is repeated for each model topology in a specified set, such as that
displayed in \figref{fig:uHMM}. The procedure that lists, and tests, model
topologies in a set of candidates we call \emph{enumeration}.

To clarify the procedure for tracing hidden-state paths let's consider a
specific example of observed data consisting of the short binary sequence:
\begin{equation}
  11101100111101111001
  ~.
\label{eg:observed_data}
\end{equation}
If tested against each candidate in \figref{fig:uHMM}, eight of the sixteen
model topologies are possible: (a, e, g-i, l-n). For example, using
\figref{fig:uHMM}(i) and starting in state $A$, the observed data is generated
by the hidden-state path:
\begin{equation}
  ABABBABBBABABBABABBBA
  ~.
\label{eg:state_path}
\end{equation}
One way to describe this path---one that is central to statistical
estimation---is to count the number of times each edge in the model was
traversed. Using $n(\hiddenstate \ms \vert \hiddenstate_{0})$ to denote the
number of times that symbol $\ms$ is generated using an edge from state
$\hiddenstate$ given that the sequence starts in state $\hiddenstate_{0}$, we
obtain: $n(A0\vert A)=0$, $n(A1\vert A)=7$, $n(B0\vert A)=6$, and $n(B1\vert
A)=7$, again assuming $\hiddenstate_{0}=A$. Similar paths and sets of edge
counts are found for the eight viable topologies cited above. These counts are
the basis for estimating a topology's transition and start-state
probabilities.  From these, one can then calculate the probability that each
model topology produced the observed data series---each candidate's posterior
probability.

By way of outlining what is to follow, let's formalize the procedure just
sketched in terms of the primary goal of estimating candidates' posterior
probabilities. First, Sec.~\ref{sec:StructuredProcess} recapitulates what is
known about the space of structured processes, reviewing how they are
represented as \eMs\ and how topological \eMs\ are exactly enumerated. Then,
Sec.~\ref{sec:BayesianInference} adapts Bayesian inference methods to this
model class, analyzing transition probability and start state estimation for a
single, known topology. Next, setting the context for comparing model
topologies, it explores the organization of the prior over the set $\Mset$ of
candidate models. Section \ref{sec:BayesianInference} closes with a discussion
of how to estimate various process statistics from functions of model
parameters.  Finally, Sec.~ \ref{sec:Examples} applies BSI to a series of
increasingly complex processes: (i) a finite-order Markov process, (ii) an
infinite-order Markov process, and, finally, (iii) an infinite-memory process.
Each illustrates BSI's effectiveness by emphasizing its ability to accurately
estimate a process's stored information (statistical complexity $\Cmu$) and
randomness (Shannon entropy rate $\hmu$).

\section{Structured Processes}
\label{sec:StructuredProcess}

We describe a system of interest in terms of its observed behavior, following
the approach of computational mechanics, as reviewed in \cite{Crut12a}. Again,
a \emph{process} is the collection of behaviors that the system produces. A
process's probabilistic description is a bi-infinite chain of random variables,
denoted by capital letters
$\ldots \, \MS_{t-2} \, \MS_{t-1} \, \MS_{t} \, \MS_{t+1} \, \MS_{t+2} \ldots$.
A realization is indicated by lowercase letters
$\ldots \, \ms_{t-2} \, \ms_{t-1} \, \ms_{t} \, \ms_{t+1} \, \ms_{t+2} \ldots$.
We assume the value $\ms_{t}$ belongs to a discrete alphabet $\MeasAlphabet$.
We work with blocks $\MS_{t:t^\prime} = \MS_t \ldots \MS_{t^\prime-1}$, where
the first index is inclusive and the second exclusive.

\EMs\ were originally defined in terms of prediction, in the so-called history 
formulation \cite{Crut88a,Crut12a}. Given a past realization
$\ms_{-\infty:t} = \ldots \ms_{t-2} \, \ms_{t-1}$ and future random variables 
$\MS_{t:\infty} = \MS_{t} \, \MS_{t+1} \ldots$, the conditional distributions
$\Pr(X_{t:\infty} \vert \meassymbol_{-\infty:t})$ define the predictive
equivalence relation over pasts: 
\begin{equation}
\ms_{-\infty:t} \sim \ms_{-\infty:t^{\prime}} \Leftrightarrow
  \Pr(X_{t:\infty} \vert \ms_{-\infty:t})
  \! = \!
  \Pr(X_{t^{\prime}:\infty} \vert \ms_{-\infty:t^{\prime}})
  .
\label{eqn:equivalence}
\end{equation}
Within the history formulation, a process determines the \eM\ topology through 
$\sim$: The \emph{causal states} $\HiddenStateSet$ are its equivalence classes
and these, in turn, induce state-transition dynamics \cite{Crut12a}. This way
of connecting a process and its \eM\ influenced previous approaches to structural
inference \cite{Crut88a,Shal98a,Shal04a}. 

The \eM\ generator formulation, an alternative, was motivated by the
problem of synchronization \cite{Trav10a,Trav10b}. There, an \eM\ topology
defines the process that can be generated by it. Recently, the generator and
history formulations were proven to be equivalent \cite{Trav11a}. Although, the
history view is sometimes more intuitive, the generator view is useful in a
variety of applications, especially the approach to structural inference
developed here.

Following \cite{Trav10a,Trav10b,Trav11a}, we start with four definitions that
delineate the model classes relevant for temporal pattern discovery.

\begin{Def}
\label{def:HMM}
A \emph{finite-state, edge-labeled hidden Markov model} (HMM) consists of:
\begin{enumerate*}
  \item A finite set of hidden states
  $\HiddenStateSet = \left\{ 
                     \hiddenstate_{1}, \ldots ,\hiddenstate_{N} \right\}$.
  \item A finite output alphabet $\MeasAlphabet$.
  \item A set of $N \times N$ symbol-labeled transition matrices
  $T^{(\meassymbol)}$, $\meassymbol \in \MeasAlphabet$, where
  $T^{(\meassymbol)}_{i,j}$ is the probability of transitioning from state
  $\hiddenstate_{i}$ to state $\hiddenstate_{j}$ and emitting symbol
  $\meassymbol$. The corresponding overall state-to-state transition matrix is
  denoted $T = \sum_{\meassymbol \in \MeasAlphabet} T^{(\meassymbol)}$.
\end{enumerate*}
\end{Def}

\begin{Def}
\label{def:uHMM}
A \emph{finite-state, edge-labeled, unifilar HMM} (uHMM) is a finite-state,
edge-labeled HMM with the following property:
\begin{itemize*}
    \item \emph{Unifilarity}: For each state $\hiddenstate_{i} \in
  \HiddenStateSet$ and each symbol $\meassymbol \in \MeasAlphabet$ there is at
  most one outgoing edge from state $\hiddenstate_{i}$ that outputs symbol
  $\meassymbol$.
\end{itemize*}
\end{Def}

\begin{Def}
\label{def:geneM}
A \emph{finite-state \eM} is a uHMM with the following property:
\begin{itemize*}
    \item \emph{Probabilistically distinct states}: For each pair of distinct
  states $\hiddenstate_{k}, \hiddenstate_{j} \in \HiddenStateSet$ there exists
  some finite word $w = \meassymbol_{0} \meassymbol_{1} \ldots
  \meassymbol_{L-1}$ such that:
  \[
    \Pr( w \vert \hiddenstate_{0} = \hiddenstate_{k} ) \neq
    \Pr( w \vert \hiddenstate_{0} = \hiddenstate_{j} )
  ~.
  \]
\end{itemize*}
\end{Def}

\begin{Def}
\label{def:topeM}
A \emph{topological \eM} is a finite-state \eM\ where the transition
probabilities for leaving each state are equal for all outgoing edges.
\end{Def}

These definitions provide a hierarchy in the model topologies to be
considered.  The most general set (Def.~\ref{def:HMM}) consists of
finite-state, edge-labeled HMM topologies with few restrictions. These are
similar to models employed in many machine learning and bioinformatics
applications; see, e.g., \cite{Yoon2009}. Using Def.~\ref{def:uHMM}, the class
of HMMs is further restricted to be unifilar. The inference methods developed
here apply to all model topologies in this class, as well as all more
restricted subclasses. As a point of reference, \figref{fig:uHMM} shows all
binary, full-alphabet (able to generate both $0$s and $1$s) uHMM topologies
with one or two states.  If all states in the model are probabilistically
distinct, following Def.~\ref{def:geneM}, these model topologies are also valid
generator \eMs.  Whether a uHMM is also a valid \eM\ often depends on the
specific transition probabilities for the machine; see Sec.~\ref{sec:informal}
for an example. This dependence motivates the final restriction to topological
\eMs\ (Def.~\ref{def:topeM}), which are guaranteed to be minimal even if
transition probabilities are equal.

Here, we employ the set of topological \eMs\ for structural inference.
Although specific settings of the transition probabilities are used to
\emph{define the set of allowed model topologies} this does not affect the
actual inference procedure.  For example, in \figref{fig:uHMM} only (a-f, j-k)
are topological \eMs. However, the set of topological \eMs\ does exclude a
variety of model topologies that might be useful for general time-series
inference.  For example, when Def.~\ref{def:topeM} is applied, all processes
with full support (all words allowed) reduce to a single-state model. However,
broadening the class of topologies beyond the set considered here is
straightforward and so we address extending the present methods to them in a
sequel. The net result emphasizes structure arising from the distribution's
support and guarantees that inferred models can be interpreted as valid \eMs.
And, the goal is to present BSI's essential ideas for one class of structured
processes---the topological \eMs.

\begin{table}
%\begin{ruledtabular}
\begin{tabular}{cr}
\hline
\hline
States & ~~~~\EMs\      \\ \noalign{\smallskip}
 $n$   & $F_{n,2}$  \\ \noalign{\smallskip}
\hline
1      & 1  \\
%       & 1  \\ \hline \noalign{\smallskip}
2      & 7  \\
%       & 1  \\
%       & 6  \\ \hline \noalign{\smallskip}
3      & 78 \\
%       & 2  \\
%       & 22 \\
%       & 54 \\ \hline \noalign{\smallskip}
4      & 1,388 \\
%       & 3   \\
%       & 68  \\
%       & 403 \\
%       & 914 \\ \hline \noalign{\smallskip}
5      & 35,186 \\
%       & 6      \\
%       & 192    \\
%       & 2,228  \\
%       & 10,886 \\
%       & 21,874 \\
\hline
\hline
\end{tabular}
%\end{ruledtabular}
\caption{Size $F_{n,2}$ of the enumerated library of full-alphabet, binary
  topological \eMs\ from one to five states.
  Reproduced with permission from \protect\cite{John10a}.
  }
\label{tab:topeM}
\end{table}

The set of topological \eMs\ can be exactly and efficiently enumerated
\cite{John10a}, motivating the use of this model class as our first example
application of BSI. Table \ref{tab:topeM} lists the number $F_{n,k}$ of
full-alphabet topologies with $n = 1, \ldots, 5$ states and alphabet size $k =
2$. Compare this table with the model topologies in \figref{fig:uHMM}, where
all $n = 1$ and $n = 2$ uHMMs are shown.  Only \figref{fig:uHMM}(a-f,j-k) are
topological \eMs, accounting for the difference between the eight models in the
table above and the fourteen in \figref{fig:uHMM}.  For comparison, the library
has been enumerated up to eight states, containing approximately $2 \times
10^9$ distinct topologies. However, for the examples to follow we employ all
$36,660$ binary model topologies up to and including five states as the
candidate basis for structural inference.

\section{Bayesian Inference}
\label{sec:BayesianInference}

Previously, we developed methods for $k$th-order Markov chains to infer models
of discrete stochastic processes and coarse-grained continuous chaotic
dynamical systems \cite{Stre07a,Stre07b}. There, we demonstrated that correct
models for in-class data sources could be effectively and parsimoniously
estimated. In addition, we showed that the hidden-state nature of out-of-class
data sources could be extracted via model comparison between Markov orders as
a function of data series length. Notably, we also found that the entropy rate
can be accurately estimated, even when out-of-class data was considered.

The following extends the Markov chain methods to the topologically richer
model class of unifilar hidden Markov models. The starting point depends on the
unifilar nature of the {HMM} topologies considered here
(Def.~\ref{def:uHMM})---transitions from each state have a unique emitted symbol
and destination state. As we demonstrated in Sec.~\ref{sec:informal}
unifilarity also means that, given an assumed start state, an observed data
series corresponds to at most one path through the hidden states. The ability
to directly connect observed data and hidden-state paths is not possible in the
more general class of {HMMs} (Def.~\ref{def:HMM}) because they can have many,
often exponentially many, possible hidden paths for a single observed data
series. In contrast, as a result of unifilarity, our analytic methods previously
developed for ``nonhidden'' Markov chains \cite{Stre07a} can be applied to
infer {uHMMs} and \eMs\ by adding a latent (hidden) variable for the unknown
start state. We note in passing that for the more general class of {HMMs},
including nonunifilar topologies, there are two approaches to statistical
inference. The first is to convert them to a {uHMM} (if possible), using mixed
states \cite{Crut08b,Elli13a}. The second is to use more conventional
computational methods, such as Baum-Welch \cite{Rabner1989}.

Setting aside these alternatives for now, we formalize the connection between
observed data series and a candidate {uHMM} topology discussed in Sec.
\ref{sec:informal}. We assume that a data series $D_{0:T} = \ms_{0} \ms_{1}
\ldots \ms_{T-2} \ms_{T-1}$ of length $T$ has been obtained from the process of
interest, with $\ms_{t}$ taking values in a discrete alphabet $\MeasAlphabet$.
When a specific model topology and start state are assumed, a hidden-state
sequence corresponding to the observed data can sometimes, but not always, be
found. We denote a hidden state at time $t$ as $\hiddenstate_{t}$ and a
hidden-state sequence corresponding to $D_{0:T}$ as $S_{0:T+1} =
\hiddenstate_{0} \hiddenstate_{1} \ldots \hiddenstate_{T-1} \hiddenstate_{T}$.
Note that the state sequence is longer than the observed data series since the
start and final states are included. Using this notation, an observed symbol
$\ms_{t}$ is emitted when transitioning from state $\hiddenstate_{t}$ to
$\hiddenstate_{t+1}$. For example, using the observed data in
\eqnref{eg:observed_data}, a hidden-state path corresponding to
\eqnref{eg:state_path} can be obtained by assuming topology
\figref{fig:uHMM}(i) and start state $A$.

We can now write out the probability of an observed data series. We assume a
stationary uHMM topology $\Mi$ with a set of hidden states $\hiddenstate_{i}
\in \HiddenStateSet_{i}$.  We add the subscript $i$ to make it clear that we
are analyzing a set of distinct, enumerated model topologies.  As demonstrated
in the example from Sec.~\ref{sec:informal}, edge counts $n(\hiddenstate_{i}
\ms \vert \hiddenstate_{i,0})$ are obtained by tracing the hidden-state path
given an assumed start state $\hiddenstate_{i,0}$.  Putting this all together,
the probability of observed data $D_{0:T}$ and corresponding state-path
$S_{0:T+1}$ is:
\begin{eqnarray} \label{eqn:likelihood}
  \Pr(S_{0:T+1}, D_{0:T} )
  & = &
  p( \hiddenstate_{i,0} ) \\ 
  & \times & 
  \prod_{\hiddenstate_{i} \in \HiddenStateSet_{i}} 
  \prod_{\ms \in \MeasAlphabet}
  p( \ms \vert \hiddenstate_{i} )^{
   n(\hiddenstate_{i} \ms \vert \hiddenstate_{i,0})
   } ~. \nonumber
\end{eqnarray}
A slight manipulation of \eqnref{eqn:likelihood} lets us write the
probability of observed data and hidden dynamics, given an assumed start
state $\hiddenstate_{i,0}$, as:
\begin{equation} \label{eqn:likelihood_given_ss}
  \Pr(S_{0:T+1}, D_{0:T} \vert \hiddenstate_{i,0} )
  =
  \prod_{\hiddenstate_{i} \in \HiddenStateSet_{i}} 
  \prod_{\ms \in \MeasAlphabet}
  p( \ms \vert \hiddenstate_{i} )^{
   n(\hiddenstate_{i} \ms \vert \hiddenstate_{i,0})
   } ~.
\end{equation}
The development of \eqnref{eqn:likelihood_given_ss} and the simple example
provided in Sec.~\figref{sec:informal} lay the groundwork for our
application of Bayesian methods. That is, given topology $\Mi$ and start state
$\hiddenstate_{i,0}$, the probability of \emph{observed} data $D_{0:T}$ 
and \emph{hidden} dynamics $S_{0:T+1}$ can be calculated. For the purposes of
inference, the combination of observed and hidden sequences is our data
$\mathbf{D}=(D_{0:T},S_{0:T+1})$.

\subsection{Inferring Transition Probabilities}

The first step is to infer transition probabilities for a single {uHMM} or
topological \eM\ $\Mi$. As noted above, we must assume a start state
$\hiddenstate_{i,0}$ so that edge counts $n(\hiddenstate_{i},\ms \vert
\hiddenstate_{i,0})$ can be obtained from $D_{0:T}$. This requirement means
that the inferred transition probabilities also depend on the assumed start
state. At a later stage, when comparing model topologies, we demonstrate that
the uncertainty in start state can be averaged over.

The set $\{ \theta_i \}$ of parameters to estimate consists of those transition
probabilities defined to be neither one nor zero by the assumed topology:
$\theta_{i} = \{ 0 < p(\ms \vert \hiddenstate_{i}, \hiddenstate_{i,0}) < 1:
\hiddenstate_{i} \in \HiddenStateSet_{i}^{*},
\hiddenstate_{i,0} \in \HiddenStateSet_{i} \}$, where 
$\HiddenStateSet_{i}^{*} \subseteq \HiddenStateSet_{i}$ is the subset of hidden 
states with more than one outgoing edge. The resulting likelihood follows
directly from \eqnref{eqn:likelihood_given_ss}:
\begin{equation} \label{eqn:param:likelihood}
\Pr(\mathbf{D} \vert \theta_{i}, \hiddenstate_{i,0} ,M_{i})  =
  \prod_{\hiddenstate_{i} \in \HiddenStateSet_{i}}
  \prod_{\meassymbol \in \MeasAlphabet}
  p(\meassymbol \vert \hiddenstate_{i}, \hiddenstate_{i,0})^{
    n(\hiddenstate_{i},\meassymbol \vert \hiddenstate_{i,0})
    } ~,
\end{equation}
We note that the set of transition probabilities used in the above expression
are unknown when doing statistical inference. However, we can still write the
probability of the observed data given a setting for these unknown values, as
indicated by the notation for the likelihood:
$\Pr(\mathbf{D} \vert \theta_{i}, \hiddenstate_{i,0} ,M_{i})$. Although not
made explicit above, there is also a possibility that the likelihood
vanishes for some, or all, start states if the observed data is not compatible
with the topology. For example, if we attempt to use \figref{fig:uHMM}(d) for
the observed data in \eqnref{eg:observed_data} we find that neither
$\hiddenstate_{i,0}=A$ nor $\hiddenstate_{i,0}=B$ leads to viable paths for the
observed data, resulting in zero likelihood.

For later use, we denote the number of times a hidden state is visited
by $n(\hiddenstate_{i} \bullet \vert \hiddenstate_{i,0}) = \sum_{\meassymbol
\in \MeasAlphabet} n(\hiddenstate_{i},\meassymbol \vert \hiddenstate_{i,0})$.

Equation (\ref{eqn:param:likelihood}) exposes the Markov nature of the dynamics
on the hidden states and suggests adapting the methods we previously developed
for Markov chains \cite{Stre07a}. Said simply, states that corresponded there
to histories of length $k$ for Markov chain models are replaced by a hidden
state $\hiddenstate_{i}$. Mirroring the earlier approach, we employ a conjugate
prior for transition probabilities. This choice means that the posterior
distribution has the same form as the prior, but with modified parameters. In
the present case, the conjugate prior is a product of Dirichlet distributions:
\begin{eqnarray}
  \Pr(\theta_{i} \vert \hiddenstate_{i,0}, \Mi)
  & = &
  \prod_{\hiddenstate_{i} \in \HiddenStateSet_{i}^{*}} \Biggl\{
  \frac{\Gamma(\alpha(\hiddenstate_{i} \bullet \vert \hiddenstate_{i,0}))}{
    \prod_{\meassymbol \in \MeasAlphabet} 
    \Gamma(\alpha(\hiddenstate_{i} \meassymbol \vert \hiddenstate_{i,0}))}
  \Biggr.
  \nonumber \\
  & \times & \delta \Biggl( 1 - \sum_{\meassymbol \in \MeasAlphabet}
              p(\meassymbol \vert \hiddenstate_{i}, \hiddenstate_{i,0}) \Biggr)
  \label{eqn:param:prior} \\
  & \times & \Biggl.
  \prod_{\meassymbol \in \MeasAlphabet}
  p(\meassymbol \vert \hiddenstate_{i}, \hiddenstate_{i,0})^{
    \alpha(\hiddenstate_{i} \meassymbol \vert \hiddenstate_{i,0})-1
  }
  \Biggr\} \nonumber
  ~,
\end{eqnarray}
where $\alpha(\hiddenstate_{i} \bullet \vert \hiddenstate_{i,0}) = 
\sum_{\meassymbol \in \MeasAlphabet} \alpha(\hiddenstate_{i} \meassymbol \vert 
\hiddenstate_{i,0})$. In the examples to follow we take
$\alpha(\hiddenstate_{i} \meassymbol \vert \hiddenstate_{i,0}) = 1$ for all
parameters of the prior.  This results in a uniform density over the
simplex for all transition probabilities to be inferred, irrespective of start 
state \cite{Wilks1962}.

The product of Dirichlet distributions includes transition probabilities
only from hidden states in $\HiddenStateSet_{i}^{*}$ because these states have
more than one outgoing edge. For transition probabilities from states
$\sigma_{i} \not \in \HiddenStateSet_{i}^{*}$ there is no need for an 
explicit prior because the transition probability must be zero or one by 
definition of the {uHMM} topology.  As a result, the prior expectation for 
transition probabilities is:
\begin{equation} \label{eqn:param:prior_mean}
   \boldsymbol{E}_{\mbox{prior}}
   \left[ p(\meassymbol \vert \hiddenstate_{i}, \hiddenstate_{i,0}) \right] =
     \frac{\alpha(\hiddenstate_{i} \meassymbol \vert \hiddenstate_{i,0})}{
           \alpha(\hiddenstate_{i} \bullet \vert \hiddenstate_{i,0})}
  ~,
\end{equation}
for states $\hiddenstate_{i} \in \HiddenStateSet_{i}^{*}$.

Next, we employ Bayes' Theorem to obtain the posterior distribution for the 
transition probabilities given data and prior assumptions. In this context, 
it takes the form:
\begin{equation} \label{eqn:param:Bayes}
  \Pr(\theta_{i} \vert \mathbf{D}, \hiddenstate_{i,0} , \Mi) =
    \frac{\Pr(\mathbf{D} \vert \theta_{i}, \hiddenstate_{i,0} , \Mi)
          \Pr(\theta_{i} \vert \hiddenstate_{i,0}, \Mi)
          }{\Pr(\mathbf{D} \vert \hiddenstate_{i,0}, \Mi)} ~.
\end{equation}
The terms in the numerator are already specified above as the likelihood 
and the prior, Eqs. (\ref{eqn:param:likelihood}) and (\ref{eqn:param:prior}),
respectively.

The normalization factor in \eqnref{eqn:param:Bayes} is called the
\emph{evidence}, or \emph{marginal likelihood}. This term integrates the
product of the likelihood and prior with respect to the set of transition
probabilities
$\theta_{i}$:
\begin{eqnarray} 
  \Pr(\mathbf{D} \vert \hiddenstate_{i,0} , \Mi)
  & = &
  \int \, d\theta_{i} \, 
    \Pr(\mathbf{D} \vert \theta_{i}, \hiddenstate_{i,0} , \Mi)
    \Pr(\theta_{i} \vert \hiddenstate_{i,0} , \Mi)
  \nonumber \\
  & = &
  \prod_{\hiddenstate_{i} \in \HiddenStateSet_{i}^{*}} \Biggl\{
  \frac{\Gamma(\alpha(\hiddenstate_{i} \bullet \vert \hiddenstate_{i,0}))}{
         \prod_{\meassymbol \in \MeasAlphabet} 
         \Gamma(\alpha(\hiddenstate_{i} \meassymbol \vert \hiddenstate_{i,0}))}
  \label{eqn:param:evidence} \\
  & \times &
  \frac{\prod_{\meassymbol \in \MeasAlphabet}
         \Gamma(\alpha(\hiddenstate_{i} \meassymbol \vert \hiddenstate_{i,0})
                + n(\hiddenstate_{i} \meassymbol \vert \hiddenstate_{i,0}))}{
        \Gamma(\alpha(\hiddenstate_{i} \bullet \vert \hiddenstate_{i,0})
                +n(\hiddenstate_{i} \bullet \vert \hiddenstate_{i,0}))}
  \Biggr\} \nonumber
  ~,
\end{eqnarray}
resulting in the average of the likelihood with respect to the prior.  In 
addition to normalizing the posterior distribution
(Eq. (\ref{eqn:param:Bayes})),
the evidence is important in our subsequent applications of Bayes' Theorem.
In particular, the quantity is central to the model selection to follow and is
used to (i) determine the start state given the model and (ii) compare model
topologies.

As discussed above, conjugate priors result in a posterior distribution of the
same form, with prior parameters modified by observed counts:
\begin{align} \label{eqn:param:posterior}
  P(\theta_{i} \vert \mathbf{D}, & \hiddenstate_{i,0} , \Mi) \nonumber \\
  & =
  \prod_{\hiddenstate_{i} \in \HiddenStateSet_{i}^{*}} \Biggl\{
  \frac{\Gamma(\alpha(\hiddenstate_{i} \bullet \vert \hiddenstate_{i,0})
               +n(\hiddenstate_{i} \bullet \vert \hiddenstate_{i,0}))}{
        \prod_{\meassymbol \in \MeasAlphabet}
        \Gamma(\alpha(\hiddenstate_{i} \meassymbol \vert \hiddenstate_{i,0})
        +n(\hiddenstate_{i} \meassymbol  \vert \hiddenstate_{i,0}))}
  \Biggr.
  \nonumber \\
  & ~~~ \times \delta \Biggl( 1 - \sum_{\meassymbol \in \MeasAlphabet}
                      p(\meassymbol \vert \hiddenstate_{i}, \hiddenstate_{i,0}) 
                    \Biggr) \\
  & ~~~ \times \Biggl.
  \prod_{\meassymbol \in \MeasAlphabet}
  p(\meassymbol \vert \hiddenstate_{i}, \hiddenstate_{i,0} )^{
   \alpha(\hiddenstate_{i} \meassymbol \vert \hiddenstate_{i,0})
   +n(\hiddenstate_{i} \meassymbol \vert \hiddenstate_{i,0})-1
   }
  \Biggr\} \nonumber
  ~.
\end{align}
Comparing Eqs. (\ref{eqn:param:prior}) and (\ref{eqn:param:posterior})---prior
and posterior, respectively---shows that the distributions are very similar:
$\alpha(\hiddenstate_{i} \meassymbol \vert \hiddenstate_{i,0})$ (prior only) 
is replaced by $\alpha(\hiddenstate_{i} \meassymbol \vert \hiddenstate_{i,0}) +
n(\hiddenstate_{i} \meassymbol \vert \hiddenstate_{i,0})$ (prior plus data).
Thus, one can immediately write down the posterior mean for the transition
probabilities:
\begin{align}
   \boldsymbol{E}_{\mbox{post}}
   & \left[ p(\meassymbol \vert \hiddenstate_{i}, \hiddenstate_{i,0} )  \right]
   \nonumber \\
   & = \frac{\alpha(\hiddenstate_{i} \meassymbol\vert \hiddenstate_{i,0})
            + n(\hiddenstate_{i} \meassymbol\vert \hiddenstate_{i,0})}{
            \alpha(\hiddenstate_{i} \bullet\vert \hiddenstate_{i,0})
            + n(\hiddenstate_{i} \bullet\vert \hiddenstate_{i,0})}
  ~,
\label{eqn:param:posterior_mean}
\end{align}
for states $\hiddenstate_{i} \in \HiddenStateSet_{i}^{*}$. As with the prior, 
probabilities for transitions from states
$\hiddenstate_{i} \notin \HiddenStateSet_{i}^{*}$ are zero or one, as
defined by the model topology.

Notably, the posterior mean for the transition probabilities does not
completely specify our knowledge since the uncertainty, reflected in functions
of the posterior's higher moments, can be large. These moments are available
elsewhere \cite{Wilks1962}. However, using methods detailed below, we employ
sampling from the posterior at this level, as well as other inference levels,
to capture estimation uncertainty.

\subsection{Inferring Start States}

The next task is to calculate the probabilities for each start state given a
proposed machine topology and observed data. Although we are not typically
interested in the actual start state, introducing this latent variable is
necessary to develop the previous section's analytic methods. And, in any case,
another level of Bayes' Theorem allows us to average over uncertainty in start
state to obtain the probability of observed data for the topology, independent
of start state.

We begin with the evidence $\Pr(\mathbf{D} \vert \hiddenstate_{i,0} ,\Mi)$
derived in \eqnref{eqn:param:evidence} to estimate transition probabilities.
When determining the start state, the evidence (marginal likelihood) from
inferring transition probabilities becomes the likelihood for start state
estimation.  As before, we apply Bayes' Theorem, this time with unknown start
states, instead of unknown transition probabilities:
\begin{equation} \label{eqn:startstate:posterior}
  \Pr(\hiddenstate_{i,0} \vert \mathbf{D}, \Mi) =
    \frac{\Pr(\mathbf{D} \vert \hiddenstate_{i,0} ,\Mi) 
          \Pr(\hiddenstate_{i,0} \vert \Mi) }{
          \Pr(\mathbf{D} \vert \Mi)} ~.
\end{equation}
This calculation requires defining a prior over start states 
$\Pr(\hiddenstate_{i,0} \vert \Mi)$. In practice, setting start states as 
equally probable \emph{a priori} is a sensible choice in light of the larger
goal of structural inference. The normalization $\Pr(\mathbf{D} \vert \Mi)$,
or evidence, at this level follows by averaging over the uncertainty in
$\hiddenstate_{i,0}$:
\begin{equation} \label{eqn:startstate:evidence}
  \Pr( \mathbf{D} \vert \Mi) = 
    \sum_{\hiddenstate_{i,0} \in \HiddenStateSet_{i}}
     \Pr(\mathbf{D} \vert \hiddenstate_{i,0} ,\Mi) 
     \Pr(\hiddenstate_{i,0} \vert \Mi) ~.
\end{equation}

The result of this calculation no longer explicitly depends on start states or
transition probabilities. The uncertainty created by these unknowns has been
averaged over, producing a very useful quantity for comparing different
topologies: $\Pr(\mathbf{D} \vert \Mi)$. However, one must not forget that
inferring transition and start state probabilities underlies the structural
comparisons to follow.  In particular, the priors set at the levels of
transition probabilities and start states can impact the structures detected
due to the hierarchical nature of the inference:
$\Pr(\mathbf{D} \vert \theta_{i}, \hiddenstate_{i,0}, \Mi) \rightarrow
\Pr(\mathbf{D} \vert \hiddenstate_{i,0}, \Mi) \rightarrow
\Pr(\mathbf{D} \vert \Mi)$.

\subsection{Inferring Model Topology}

So far, we inferred transition probabilities and start states for a given model
topology. Now, we are ready to compare different topologies in a set
$\Mset$ of candidate models. As with inferring start states given a
topology, we write down yet another version Bayes' Theorem, except one for
model topology:
\begin{equation} \label{eqn:structure:posterior}
  \Pr(\Mi \vert \mathbf{D}, \Mset) =
    \frac{\Pr(\mathbf{D} \vert \Mi , \Mset) \Pr(\Mi \vert \Mset)}{
    \Pr(\mathbf{D} \vert \Mset) }
	~,
\end{equation}
writing the likelihood as $\Pr(\mathbf{D} \vert \Mi , \Mset)$ to make the nature
of the conditional distributions clear. This is exactly the same, however, as
the evidence derived above in Eq. (\ref{eqn:startstate:evidence}):
$\Pr(\mathbf{D} \vert \Mi) = \Pr(\mathbf{D} \vert \Mi , \Mset)$. Equality holds
because nothing in calculating the previous evidence term directly depends on
the \emph{set of models} considered. The evidence $\Pr(\mathbf{D} \vert
\Mset)$, or normalization term, in \eqnref{eqn:structure:posterior} has the
general form:
\begin{equation} \label{eqn:structure:evidence}
  \Pr(\mathbf{D} \vert \Mset) =
    \sum_{M_{j} \in \Mset} \Pr(\mathbf{D} \vert M_{j}, \Mset) 
    \Pr(M_{j} \vert \Mset) ~.
\end{equation}

To apply \eqnref{eqn:structure:posterior} we must first provide an explicit
prior over model topologies. One general form, tuned by single parameter 
$\beta$, is:
\begin{equation} \label{eqn:structure:prior}
  \Pr(\Mi \vert \Mset) =
    \frac{\exp\left(- \beta \phi(\Mi) \right)}{
    \sum_{M_{j} \in \Mset} \exp\left(- \beta \phi(M_{j}) \right) }
  ~,
\end{equation}
where $\phi(\Mi)$ is some desired function of model topology. In the examples
to follow we use the number of causal
states---$\phi(\Mi) = \vert \Mi \vert$---thereby penalizing for model size.
This is particularly important when a short data series is being investigated.
However, setting $\beta=0$ removes the penalty, making all models in $\Mset$
\emph{a priori} equally likely. It is important to investigate the effects of
choosing a specific $\beta$ for a given set of candidate topologies. Below,
we first demonstrate the effect of choosing $\beta = 0$, $2$, or $4$. After
that, however, we employ $\beta=4$ since this value, in combination with the
set of one- to five-state binary-alphabet topological \eMs, produces a
preference for one- and two-state machines for short data series and still
allows for inferring larger machines with only a few thousand symbols.
Experience with this $\beta$ shows that it is structurally conservative.

In the examples we explore two approaches to using the results of structural
inference. The first takes into account all model topologies in the set
considered, weighted according to the posterior distribution given in
\eqnref{eqn:structure:posterior}. The second selects a single model
$M_{\mbox{map}}$ that is the \emph{maximum a posteriori} (MAP) topology:
\begin{equation} \label{eqn:structure:map}
  M_{\mbox{map}} = \argmax_{M_{i} \in \Mset} \Pr(\Mi \vert \mathbf{D}, \Mset)
  ~.
\end{equation}
The difference between these methods is most dramatic for short data series.
Also, using the MAP topology often underestimates the uncertainty in functions
of the model parameters; which we discuss shortly. Of course, since one throws
away any number of comparable models, estimating uncertainty in any quantity
that explicitly depends on the model topology cannot be done properly if MAP
selection is employed. However, we expect some will want or need to use a
single model topology, so we consider both methods.

\begin{figure}
  \centering
  \includegraphics{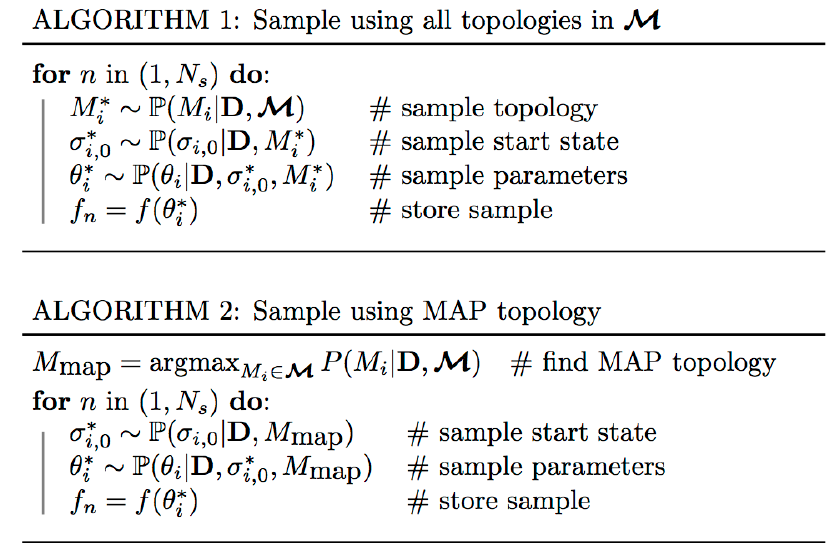}
\caption{Pseudocode for generating $N_{s}$ samples of a function $f(\theta_i)$
  of model parameters $\{\theta_i\}$. Algorithm 1 samples a topology each time
  through the loop, whereas Algorithm 2 uses the MAP topology for all
  iterations. The sampling at each stage allows for the creation of a set of
  samples $\{ f_{n} \}$ that accurately reflect the many sources of uncertainty
  in the posterior distribution.
  }
\label{alg:sample:set}
\end{figure}

\subsection{Estimating Functions of Model Parameters}
\label{sec:sampling}

A primary goal in inference is estimating functions that depend on an inferred
model's parameters. We denote this $f(\theta_{i})$ to indicate the dependence
on transition probabilities. Unfortunately, substituting the posterior mean for
the transition probabilities into some function of interest does not provide 
the desired expectation. In general, obtaining analytic expressions for the 
posterior mean of desired functions is quite difficult; see, for example,
\cite{Wolp1995,Yuan1997}. Deriving expressions for the uncertainty in the 
resulting estimates is equally involved and typically not done; although see 
\cite{Wolp1995}. 

Above, the inference method required inferring transition probabilities, start
state, and topology. Function estimation, as a result, should take into account
all these sources of uncertainty. Instead of deriving analytic expressions for
posterior means (if possible), we turn to numerical methods to estimate function
means and uncertainties in great detail.  We do this by repeatedly sampling from
the posterior
distribution at each level to obtain a sample \eM\ and evaluating the function
of interest for the sampled parameter values. The algorithms in
\figref{alg:sample:set} detail the process of sampling $f(\theta_{i})$ using
all candidate models $\Mset$ (Algorithm~1) or the single $M_{\mbox{\tiny MAP}}$
model (Algorithm~2). Given a set of samples of the function of interest, any
summary statistic can be employed. In the examples, we generate
$N_{s}= 50,000$ samples from which we estimate a variety of properties. More
specifically, these samples are employed to estimate the posterior mean and
the 95\%, equal-tailed, credible interval (CI) \cite{GelmanBook1995}.  This
means there is a 5\% probability of samples being outside the specified
interval, with equal probability of being above or below the interval.
Finally, a Gaussian kernel density estimation (Gkde) is used to visualize the
posterior density for the functions of interest.

The examples demonstrate estimating process randomness and structure from data
series using the two algorithms introduced above. For a known \eM\ topology
$\Mi$, with specified transition probabilities 
$\{ p(\meassymbol \vert \hiddenstate_{i}) \}$, these properties are quantified
using the entropy rate $\hmu$ and statistical complexity $\Cmu$, respectively.  
The entropy rate is:
\begin{equation}
\hmu = - \sum_{\hiddenstate_{i} \in \HiddenStateSet_{i}} p(\hiddenstate_{i})
     \sum_{\meassymbol \in \MeasAlphabet}
     p(\meassymbol \vert \hiddenstate_{i})
     \log_{2} p(\meassymbol \vert \hiddenstate_{i})
\label{eqn:entropy_rate}
\end{equation}
and the statistical complexity is:
\begin{equation}
\Cmu = - \sum_{\hiddenstate_{i} \in \HiddenStateSet_{i}}
     p(\hiddenstate_{i}) \log_{2} p(\hiddenstate_{i})
	 ~.
\label{eqn:statistical_complexity}
\end{equation}
In these expressions, the $p(\hiddenstate_{i})$ are the asymptotic state
probabilities determined by the left eigenvector (normalized in probability)
of the internal Markov chain transition matrix
$T = \sum_{\ms \in \MeasAlphabet} T^{(x)}$.
Of course, $\hmu$ and $\Cmu$ are also functions of the model topology and 
transition probabilities, so these quantities provide good examples of how
to estimate functions of model parameters in general.

\section{Examples}
\label{sec:Examples}

We divide the examples into two parts. First, we demonstrate inferring
transition probabilities and start states for a known topology. Second, we
focus on inferring \eM\ topology using the set of all binary, one- to five-state
topological \eMs, consisting of $36,660$ candidates; see 
Table \ref{tab:topeM}. We use the convergence of estimates for the 
information-theoretic values $\hmu$ and $\Cmu$ to monitor structure
discovery. However, estimating model parameters is at the core of the later
examples and so we start with this procedure.

For each example we generate a single data series $D_{0:T}$ of length 
$T=2^{17}$. When analyzing convergence, we consider subsamples $D_{0:L}$ of
lengths $L=2^{i}$, using $i=0,1,2,\ldots,17$. For example, a four-symbol
sequence starting at the first data point is designated
$D_{0:4}=\ms_{0}\ms_{1}\ms_{2}\ms_{3}$. The overlapping
analysis of a single data series gives insight into convergence for the
inferred models and for the statistics estimated.

\subsection{Estimating Parameters}

\subsubsection{Even Process}
\label{sec:even:params}

\begin{figure}
\centering
\includegraphics{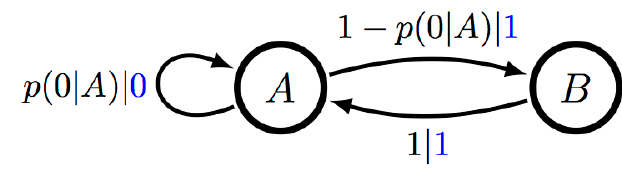}
\caption{State-transition diagram for the Even Process's \eM\ topology.
  The ``true'' value of the unspecified transition probability is
  $p(0 \vert A) = 1/2$. For this topology, 
  $\HiddenStateSet_{\mbox{even}} = \{A, B\}$ and 
  $\HiddenStateSet_{\mbox{even}}^{*} = \{A\}$ because state $B$ has only one
  outgoing transition.
  }
\label{fig:even}
\end{figure}

\begin{figure}
\centering
\includegraphics{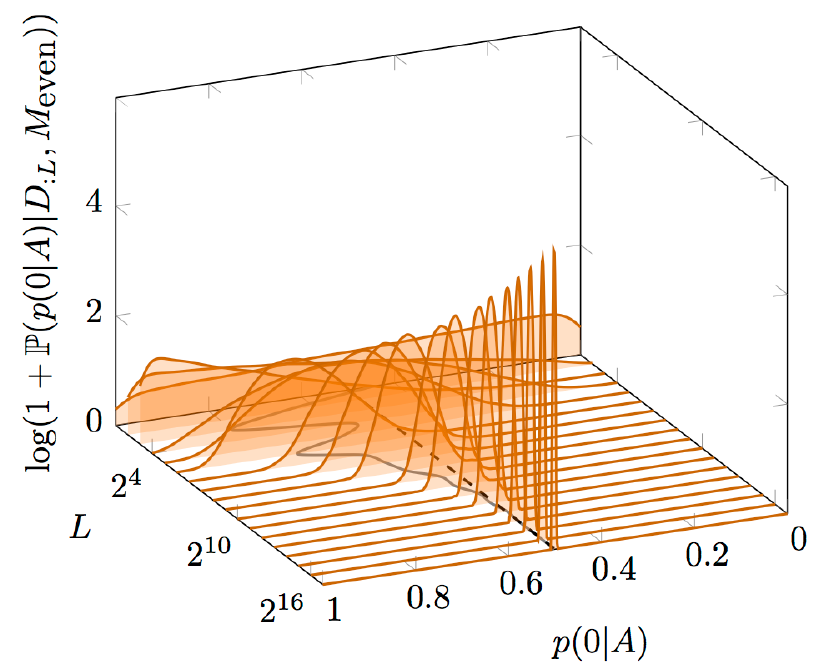}
\caption{(Color online) Convergence of posterior density
  $\Pr(p(0\vert A) \vert D_{0:L}, M_{\mbox{even}})$ as a function of subsample
  length $L=2^{i}$, $i=0,1,2,\ldots,17$. Each posterior density plot uses a
  Gaussian kernel density estimator with $50,000$ samples from the posterior.
  The true value of $p(0\vert A)=1/2$ appears as a dashed line and the
  posterior mean as a solid line.
  }
\label{fig:even:p0gAconvergence}
\end{figure}

We first explore a single example of inferring properties of a known data
source using Eqs. (\ref{eqn:param:likelihood})-(\ref{eqn:param:posterior}).
We generate a data series from the Even Process and then, using the correct
topology (Fig. \ref{fig:even}), we infer start states and transition
probabilities and estimate the entropy rate and statistical complexity.
We do not concentrate on this level of inference in subsequent examples,
preferring to focus instead on model topology and its representation of the
unknown process structure. Nonetheless, the procedure detailed here underlies
all of the examples.

\begin{figure*}
  \includegraphics[]{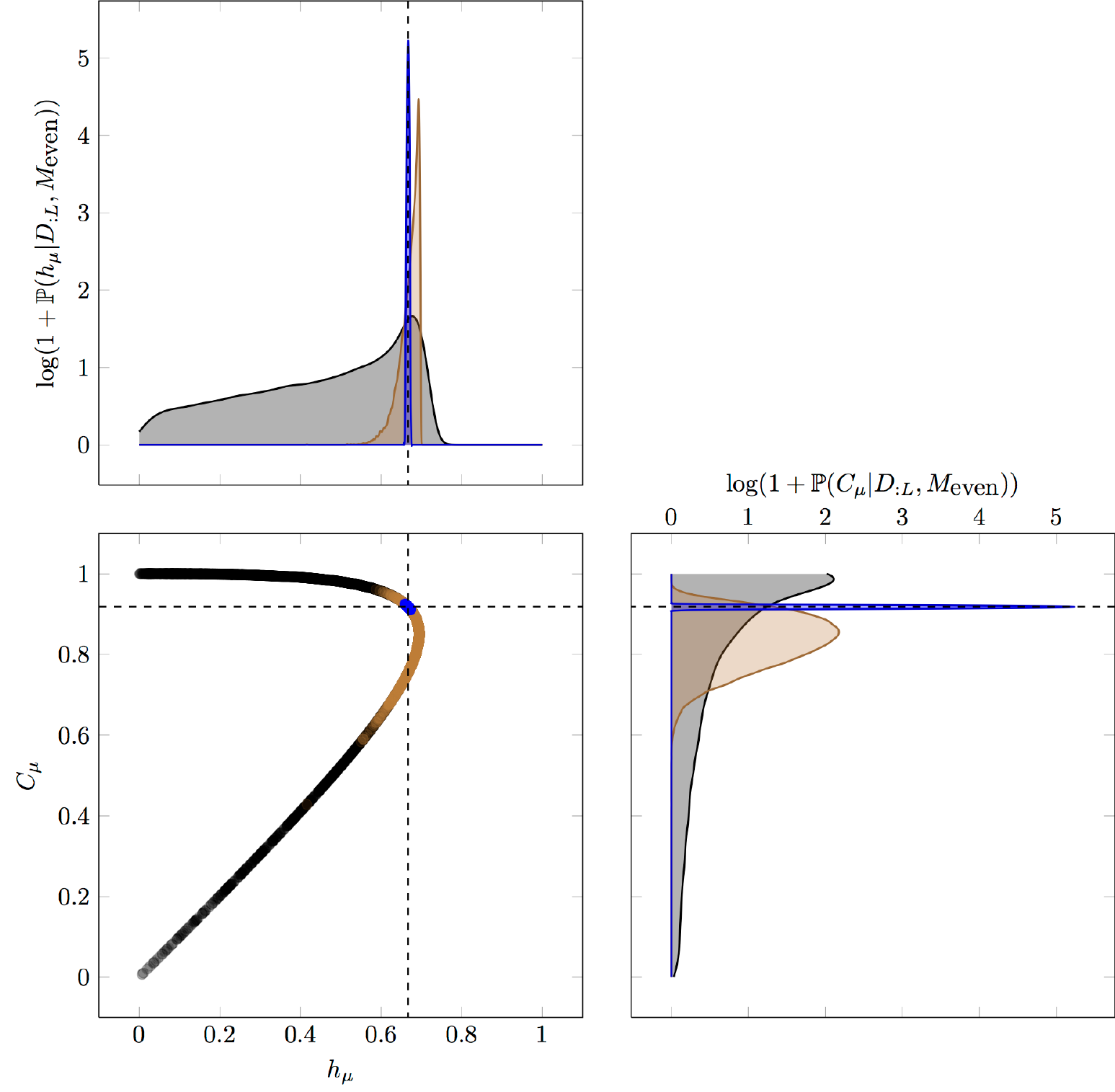}
\caption{Convergence of randomness ($\hmu$) and structure ($\Cmu$) calculated
  with transition probabilities and start states estimated from Even Process 
  data, assuming the correct topology. $50,000$ samples were taken from the
  joint posterior $\Pr(\hmu,\Cmu \vert D_{0:L}, M_{\mbox{even}})$.
  (Lower left) A subsample of size $5,000$ for data sizes $L=1$ (black),
  $L=64$ (brown), and $L=16,384$ (blue). Gaussian kernel density estimates
  (using all $50,000$ samples) of the marginal distributions
  $\Pr(\hmu \vert D_{0:L}, M_{\mbox{even}})$ (top) and
  $\Pr(\Cmu \vert D_{0:L}, M_{\mbox{even}})$ (right) for the same values of $L$.
  Dashed lines indicate the true values of $\hmu$ and $\Cmu$ for the
  Even Process.
  }
  \label{fig:even:hmuCmu}
\end{figure*}

The Even Process is notable because it has infinite Markov order. This means no
finite-order Markov chain can reproduce its word distribution \cite{Stre07a}.
It can be finitely modeled, though, with a finite-state unifilar HMM---the
\eM\ of \figref{fig:even}. A single data series was generated using the Even
Process \eM\ with $p(0\vert A)=1/2$. The start state was randomized before
generating sequence data of length $T=2^{17}$. As it turned out, the initial
segment was $D_{0:T}=100\ldots$, indicating that the unknown start state was
$B$ on that realization. This is so because the first symbol is a $1$, which can
be generated starting in either state $A$ or $B$, but the sequence $10$ is only
possible by starting at node $B$. 

Next, we estimate the transitions from the generated data series using
length-$L$ subsamples $D_{0:L} = \ms_{0} \ms_{1} \ldots \ms_{L-1}$ to track
convergence. Although the mean and other moments of the Dirichlet posterior
can be calculated analytically \cite{Wilks1962}, we sample values using
Algorithm 2 in 
\figref{alg:sample:set}.  However, in this example we employ $M_{\mbox{even}}$ 
instead of $M_{\mbox{map}}$ because we are focused on the model parameters 
for a known topology. The posterior density for each subsample $D_{0:L}$ is 
plotted in \figref{fig:even:p0gAconvergence} using Gaussian kernel density 
estimation (Gkde). The true value of $p(0\vert A)$ is shown as a black, dashed
line and the posterior mean as a solid, gray line. (Both lines connect values
evaluated at each length $L=2^{0},2^{1},\ldots 2^{17}$.) The convergence of the
posterior density to the correct value of $p(0\vert A)=1/2$ with increasing
data size is clear and, moreover, the true value is always in a region of
positive probability.

For our final example using a known topology we estimate $\hmu$ and $\Cmu$ from
the Even Process data. This illustrates estimating these functions of model
parameters when the \eM\ topology is known but there is uncertainty in start
state and transition probabilities. As above, we use Algorithm 2 in 
\figref{alg:sample:set} and employ the known machine structure. We sample start
states and transition probabilities, followed by calculating $\hmu$ and 
$\Cmu$---via Eqs. (\ref{eqn:entropy_rate}) and
(\ref{eqn:statistical_complexity}), respectively---to build a posterior
density for these quantities.  

Figure \ref{fig:even:hmuCmu} presents the joint distribution for $\Cmu$ and
$\hmu$ along with the Gkde estimation of their marginal densities. Samples
from the joint posterior distribution are plotted in the lower left panel for
subsample lengths $L=1,64,$ and $16,384$. Only $5,000$ of the available
samples are displayed in this panel to minimize the graphic's size.
The marginal densities for $\hmu$ (top panel) and $\Cmu$ (right panel) 
are plotted using a Gkde with all $50,000$ samples. Small data size
($L=1$, indicated by black points) samples allow a wide range of structure 
and randomness constrained only by the Even Process \eM\ topology. The range
of $\hmu$ and $\Cmu$ reflect the flat priors set for start states and
transition probabilities.  We note that a uniform prior distribution over
transition probabilities and start states does not produce a uniform 
distribution over $\hmu$ or $\Cmu$.  Increasing the size of the data subsample
to $L=64$ (brown points) results in a considerable reduction in the uncertainty
for both functions.  For this amount of data, the possible values of
entropy rate and statistical complexity curve around the true value in the
$\hmu-\Cmu$ plane and result in a shifted peak for the marginal density for
$\hmu$. For subsample length $L=16,384$ (blue points) the estimates of both
functions of model parameters converge to the true values, indicated by the
black, dashed lines.

\subsection{Inferring Process Structure}

We are now ready to demonstrate BSI's efficacy for structural inference
via a series of increasingly complex processes, monitoring convergence using
data subsamples up to a length of $L = 2^{17}$. In this, we determine the
number of hidden states, number of edges connecting them, and symbols output
on each transition. As discussed above, we use the set of topological \eMs\ as
candidates because an efficient and exhaustive enumeration is available.

For comparison, we first explore the organization of the prior over the set of
candidate \eMs\ using intrinsic informational coordinates---the process entropy
rate $\hmu$ and statistical complexity $\Cmu$. We focus on their joint
distribution, as induced by various settings of the prior parameter $\beta$.
The results lead us to use $\beta=4$ for the subsequent
examples. This value creates a preference for small models when little data is
available but allows for a larger number of states when reasonable amounts
of data support it.

We establish the BSI's effectiveness by inferring the structure of a
finite-order Markov process, an infinite-order Markov process, and an
infinite memory process. Again, the proxy for convergence is estimating
structure and randomness as a function of the data subsample length $L$.
Comparing these quantities' posterior distributions with their prior
illustrates uncertainty reduction as more data is analyzed.

\subsubsection{Priors for Structured Processes}

\begin{figure*}
  \includegraphics[]{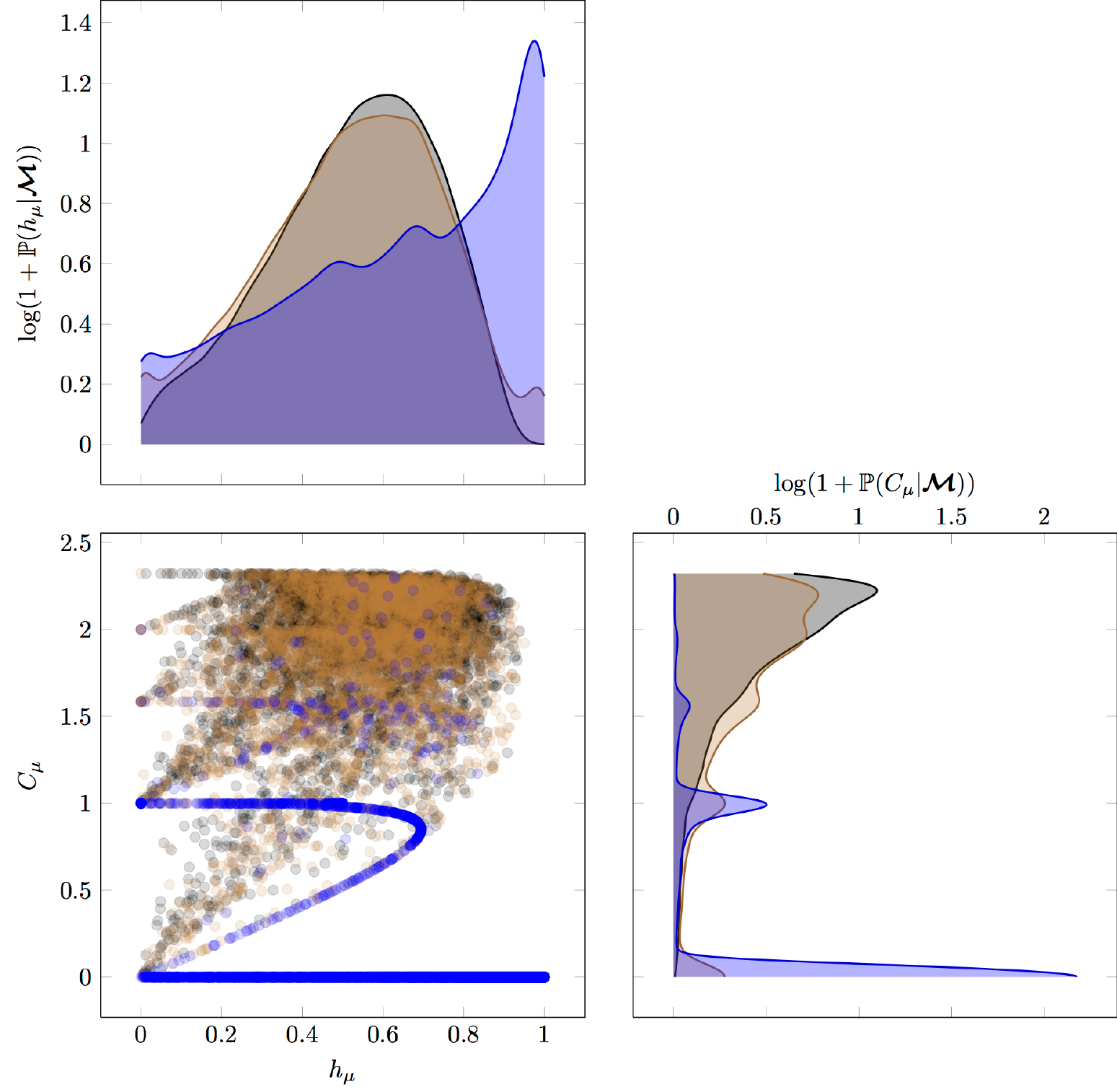}
\caption{
  Model prior dependence on penalty parameter $\beta$:
  $50,000$ samples were taken from the joint prior
  $\Pr(\hmu,\Cmu \vert \Mset)$ using all binary-alphabet, topological \eMs\ 
  with $1-5$ states and parameters: $\beta=0$ (black), $\beta=2$ (brown), and
  $\beta=4$ (blue).
  (Lower left) A subsample of size $5,000$ from the joint distribution is shown
  for each value of $\beta$. A Gaussian kernel density estimation, using all
  $50,000$ samples for each value of $\beta$, of the marginal distributions
  $\Pr(\hmu \vert \Mset)$ (top) and $\Pr(\Cmu \vert \Mset)$ (right).
  }
\label{fig:models_prior}
\end{figure*}

Here, we use a prior over all binary-alphabet, topological \eMs\ with one to
five states. (Recall Table \ref{tab:topeM}.) We denote the set of topological
\eMs\ detailed in Table \ref{tab:topeM} as $\Mset$. Equation
(\ref{eqn:structure:prior}) allows specifying a preference for smaller \eMs\ by
setting $\beta > 0$ and defining the function of model structure to be the
number of states: $\phi(M_{i}) = \vert M_{i} \vert$. Beyond setting this
explicitly, there is an inherent bias to smaller models inversely proportional
to the parameter space dimension. The parameter space is that of the estimated
transition probabilities. Its dimension is the number of states with more than
one out-going transition. However, candidate \eM\ topologies with many states
and few transitions result in a small parameter space and so may be assigned
high probability for short data series. In addition, the prior over topologies
must take into account the increasing number of candidates as the
number of states increases. Setting $\beta$ sufficiently high so that large
models are not given high probability under these conditions is reasonable,
as we would like to approach structure estimates ($\Cmu$) monotonically from
below, as data size increases.

Figure \ref{fig:models_prior} plots samples from the resulting joint prior for
$(\hmu, \Cmu)$ as well as the corresponding Gkde for marginal densities of both
quantities. The data are generated by using the method of Sec.
\ref{sec:sampling} and replacing the posterior density with the prior density.
Specifically, rather than sampling a topology $M_{i}$ from $\Pr(M_{i} \vert D,
\Mset)$, we sample from $\Pr(M_{i} \vert \Mset)$.  Similar substitutions are
made at each level, using the distributions that do not depend on observed
data, resulting in samples from the prior.  Each color in the figure reflects
samples using all \eMs\ in $\Mset$ with different values for the prior
parameter:  $\beta=0$ (black), $\beta=2$ (brown) and $\beta=4$ (blue). While
$\beta=0$ has many samples at high $\Cmu$, reflecting the large number of
five-state \eMs, increasing to $\beta=2$ results in noticeable bands in the
$\hmu-\Cmu$ plane and peaks at $\Cmu = \log_{2} 1$, $\Cmu = \log_{2} 2$, $\Cmu
= \log_{2} 3$ bits, and so on. This reflects the fact that larger $\beta$ makes
smaller machines more likely. As a consequence, the emergence of patterns due
to one-, two-, and three-state topologies is seen. Setting $\beta=4$ shows a
stronger \emph{a priori} preference for one- and two-state machines, reflected
by the strong peaks at $\Cmu=0$ bits and $\Cmu = 1$ bit.  Interestingly, the
prior distribution over $\hmu$ and $\Cmu$ is quite similar for $\beta=0$ and
$2$, with more distributional structure due to smaller \eMs\ at $\beta=2$.
However, the prior distribution for $\hmu$ and $\Cmu$ is quite different for
$\beta=4$, creating a strong preference for one- and two-state topologies.
This results in an \emph{a priori} preference for low $\Cmu$ and high $\hmu$
that, as we demonstrate shortly, is modified for moderate amounts of data.  We
employ $\beta=4$ as a reasonable value in all subsequent examples.  In
practice, sensitivity to this choice should be tested in each application to
verify that the resulting behavior is appropriate. We suggest small, nonzero
values as reasonable starting points. As always, sufficient data makes the
choice relatively unimportant for the resulting inference.

\subsubsection{Markov Example: The Golden Mean Process}

\begin{figure}
\centering
\includegraphics{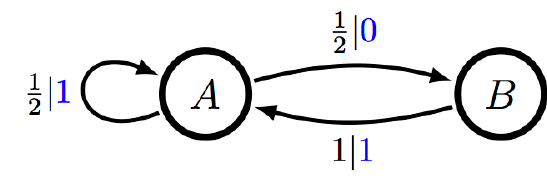}
\caption{Golden Mean Process's \eM.
  }
\label{fig:goldenmean}
\end{figure}

The first example of structural inference explores the Golden Mean Process,
pictured in \figref{fig:goldenmean}. Although it is illustrated as an {HMM} in
the figure, it is effectively a Markov chain with no hidden states: observing
a $1$ corresponds to state $A$, whereas observing $0$ means the process is in
state $B$.  
Previously, we showed that this data source can be inferred using the model 
class of $k$th order Markov chains, as expected \cite{Stre07a}. However, the
Golden Mean Process is also a member of the class of binary-alphabet,
topological \eMs\ considered here.  As a result, structural inference from
Golden Mean data is an example of in-class modeling.

We proceed using the approach laid out above for the Even Process transition
probabilities and start states. We generated a single data series by
randomizing the start state and creating a symbol sequence of length $T=2^{17}$
using the Golden Mean Process \eM. As above, we monitor the convergence using
subsamples $D_{0:L}=x_{0} x_{1} \ldots x_{L-1}$ for lengths $L=2^{i}$,
$i=0,1,\ldots 17$. The candidate machines $\Mset$ consist of all $36,600$ 
\eM\ topologies in Table \ref{tab:topeM}. Estimating $\hmu$ and $\Cmu$ aids
in monitoring convergence of inferred topology and related properties to the
correct values. In addition, we provide supplementary tables and figures,
using both $\Mset$ and the \emph{maximum a posteriori} model
$M_{\mbox{\tiny MAP}}$
at each data length $L$, to give a detailed view of structural inference.  

\begin{figure*}
  \includegraphics[]{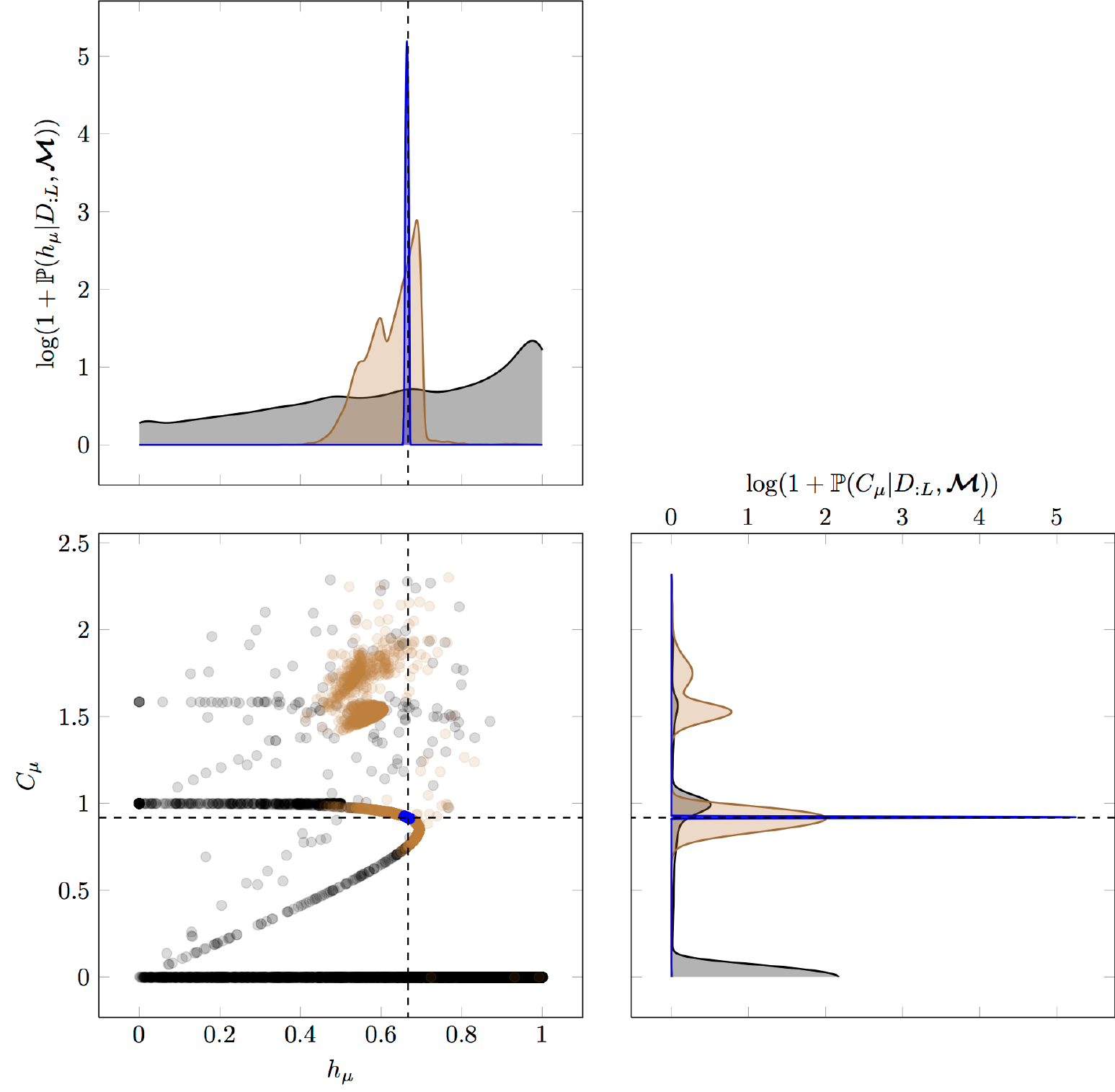}
\caption{Convergence of randomness ($\hmu$) and structure ($\Cmu$) calculated
  with model topologies, transition probabilities, and start states estimated
  from Golden Mean Process data, using all one- to five-state topological \eMs.
  $50,000$ samples were taken from the joint posterior distribution
  $\Pr(\hmu,\Cmu \vert D_{0:L}, \Mset)$.
  (Lower left) Subsample of size $5,000$ for data sizes $L=1$ (black),
  $L=64$ (brown), and $L=16,384$ (blue).
  Gaussian kernel density estimates of the marginal distributions (using all
  $50,000$ samples) $\Pr(\hmu \vert D_{0:L}, \Mset)$ (top) and
  $\Pr(\Cmu \vert D_{:L}, \Mset)$ (right) for the same values of $L$.
  Dashed lines indicate the true values of $\hmu$ and $\Cmu$
  for the Golden Mean Process.
  }
  \label{fig:gm:hmuCmu}
\end{figure*}

Figure \ref{fig:gm:hmuCmu} plots samples from the joint posterior over
$(\hmu, \Cmu)$, as well as their marginal distributions, for three subsample
lengths. As in \figref{fig:even:hmuCmu}, we consider $L=1$ (black), $L=64$
(brown), and $L=16,384$ (blue).  However, this example employs
the full set $\Mset$ of candidate topologies. For small data size ($L=1$) the
distribution closely approximates the prior distribution for $\beta=4$, as it
should. At data size $L=64$, the samples of both the $\hmu$ and $\Cmu$ are
still broad, resulting in 
multimodal behavior with considerable weight given to both two- and three-state 
topologies.  Consulting Table S2 in the supplementary material, we see that 
this is the shortest length that selects the correct topology for the Golden 
Mean Process (denoted n2k2id5 in Table S2). For smaller $L$, the single-state, 
two-edge topology is preferred (denoted n1k2id3). 
However, the probability of the correct model is only {78.7}\%, leaving a 
substantial probability for alternative candidates. The uncertainty is 
further reflected in the large credible interval for $\Cmu$ provided by 
the complete set of models $\Mset$ (see Table S1), ranging from $0.8235$ bits
as the lower bound to $1.797$ bits as the upper bound. However, by subsample
length $L=16,384$ the probability of the correct topology is {99.998}\%, given
the set of candidate machines $\Mset$, and estimates of both $\hmu$ and $\Cmu$
have converged to accurately reflect the correct values.

In addition to Tables S1 and S2, the supplementary materials provide Fig.
S1 showing the Gkde estimates of both $\hmu$ and $\Cmu$ using $\Mset$ and
$M_{\mbox{\tiny MAP}}$ as a function of subsample length. The four panels
clearly show the convergence of estimates to the correct values as $L$
increases. For long data series, there is little difference between the
inference made using the \emph{maximum a posteriori} (MAP) model and the posterior
over the entire candidate set. However, this is not true for short time series,
where using the full set more accurately captures the uncertainty in estimation
of the information-theoretic quantities of interest. We note that the $\Cmu$
estimates approach the true value from below, preferring small topologies when
there is little data and selecting the correct, larger topology only as
available data increases. This desired behavior results from setting $\beta=4$
for the prior over $\Mset$. Setting $\beta=2$, shown in Fig. S2, does not
have this effect. This value of $\beta$ is insufficient to overcome the large
number of three-, four-, and five-state \eMs. Finally, Fig. S3 plots samples
from the joint posterior of $\hmu$ and $\Cmu$ using only the MAP model for
subsample lengths $L=1,64$, and $16,384$. This should be compared with
\figref{fig:gm:hmuCmu} where the complete set $\Mset$ is used. Again, there is
a substantial difference for short data series and much in common for larger
$L$.

Before moving to the next example, let's briefly return to consider start-state
inference. The data series generated to test inferring the Golden Mean Process
started with the sequence $D_{0:T}=1110 \ldots$. We note that the correct start
state, which happens to be state $A$ in that realization, cannot be inferred
and has lower probability than state $B$ due to the process's structure: 
$\Pr(\hiddenstate_{\mbox{gm},0}=A \vert D_{0:T}=1110 \ldots, M_{\mbox{gm}} ) 
\approx 0.3328$ using \eqnref{eqn:startstate:posterior}.  The reason for the
inability to discern the start state is straightforward. Consulting
\figref{fig:goldenmean}, we can see that the string $1110$ can be
produced beginning in both states $A$ and $B$. On the one hand, assuming
$\hiddenstate_{\mbox{gm},0}=A$, the state path would be $AAAAB$ with
probability $p(1\vert A)^{3} p(0\vert A) = (1/2)^{4}$.  On the other hand,
assuming $\hiddenstate_{\mbox{gm},0}=B$, the state path is $BAAAB$ with
probability $p(1\vert B) p(1\vert A)^{2} p(0\vert A) = 1 \times (1/2)^{3}$.
The only difference in the probabilities is a factor of $p(1\vert A)=1/2$
versus $p(1\vert B)=1$ resulting in:

\begin{eqnarray*}
  \Pr(\hiddenstate_{i,0}=A \vert D=1110, M_{\mbox{gm}})
  & = &
  \frac{(1/2)^{4}}{(1/2)^{4} + (1/2)^{3}} \\[0.25em]
  & = & 1/3.
\end{eqnarray*}
This calculation agrees nicely with the result stated above, using finite data
and the inference calculations from \eqnref{eqn:startstate:posterior}.

It turns out that any observed data series from the Golden Mean Process that
begins with a $1$ will have this ambiguity in start state.  However, observed
sequences that begin with a $0$ uniquely identify $A$ as the start state since
a $0$ is not allowed leaving state $B$.  Despite this, the correct topology is
inferred and accurate estimates of $\hmu$ and $\Cmu$ are obtained.

\begin{figure*}
  \includegraphics[]{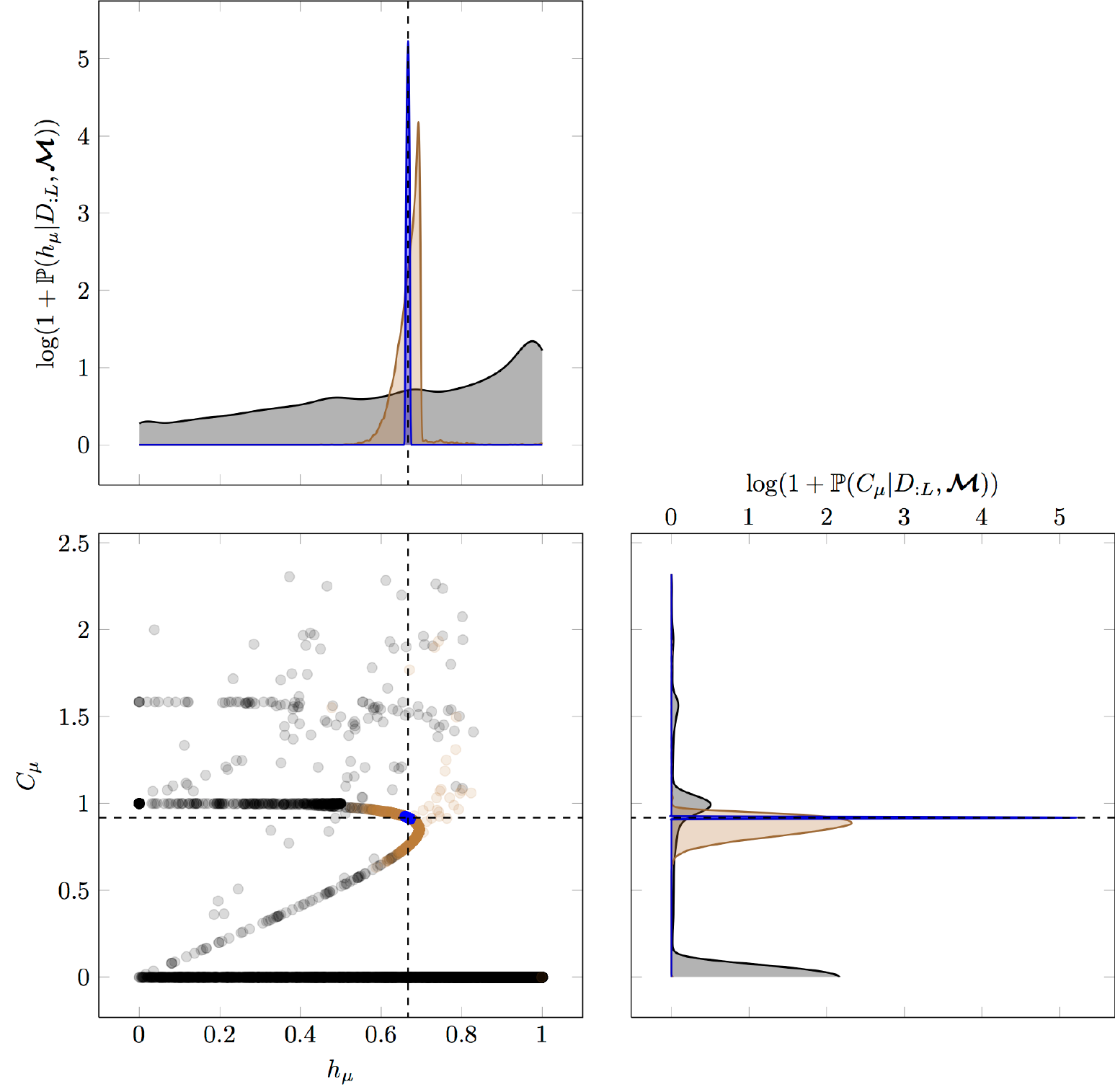}
\caption{Convergence of randomness ($\hmu$) and structure ($\Cmu$) calculated
  with model topologies, transition probabilities, and start states estimated
  from Even Process data, using all one- to five-state topological \eMs.
  $50,000$ samples from the joint posterior
  $\Pr(\hmu,\Cmu \vert D_{:L}, \Mset)$.
  (Lower left) A subsample of $5,000$ for data sizes $L=1$ (black), $L=64$
  (brown), and $L=16,384$ (blue). Gaussian kernel density estimates of the
  marginal distributions (using all $50,000$ samples)
  $\Pr(\hmu \vert D_{:L}, \Mset)$ (top) and
  $\Pr(\Cmu \vert D_{0:L}, \Mset)$ (right) are shown for the same values of
  $L$.
  Dashed lines indicate the true values of $\hmu$ and $\Cmu$ for the
  Even Process.
  }
\label{fig:even:hmuCmu:structure}
\end{figure*}

\subsubsection{Infinite-order Markov Example: The Even Process}

Next, we consider inferring the structure of the Even Process using the same set
of binary-alphabet, one- to five-state, topological \eMs. To be clear, this
example differs from Sec.~\ref{sec:even:params}, where the correct topology
was assumed. Now, we explore Even Process structure using $\Mset$. As noted
above, the Even Process is an infinite-order Markov process and inference 
requires the set of topological \eMs\ considered here. (However, see
out-of-class inference of the Even Process using $k$th-order Markov chains
in \cite{Stre07a}.) As a result, this is an example of in-class inference 
since the Even Process topology is contained within the set $\Mset$. As with
the previous example, a single data series was generated from the Even Process.

Figure \ref{fig:even:hmuCmu:structure} shows samples from the posterior
distribution over $(\hmu,\Cmu)$ using three subsample lengths $L=1, 64$, and
$16,384$ as before. An equivalent plot using only the MAP model is provided in
the supplementary materials for comparison; see Fig. S6. Again, for short data
series the samples mirror the prior distribution as they should. (See black
points for $L=1$.) At subsample length $L=64$ the values of $\hmu$ and $\Cmu$
are much more tightly delineated. Comparing samples for the Golden Mean Process
in \figref{fig:gm:hmuCmu} shows that there is much less uncertainty in
structure for the Even Process at this data size. Consulting Table S4, the MAP
topology for this value of $L$ already identifies the correct topology (denoted
n2k2id7) and assigns a probability of 99.41\%.  This high probability is
reflected by the smaller spread, when compared with the Golden Mean example, of
the samples of $\hmu$ and $\Cmu$.  At subsample length $L=16,384$ the
probability of the correct topology has grown to {99.998}\%. Estimates of both
$\hmu$ and $\Cmu$ are also very accurate, with small uncertainties, at this
$L$; see Table S3.

The supplementary materials provide Figs. S4 and S5 to show the convergence of
the posterior densities for $\hmu$ and $\Cmu$ as a function of subsample
length. Figure S4 shows estimates using both $\Mset$ and $M_{\mbox{\tiny MAP}}$
for $\beta=4$. Whereas, Fig. S5 demonstrates the effects of using a small
penalty ($\beta=2$) for model size. As seen with the Golden Mean Process, the
difference is most apparent at small data sizes.  At large $L$, the difference
between using the complete set $\Mset$ of models versus the MAP model is minor,
as is the effect of choosing $\beta=4$ or $\beta=2$.  However, at small data
sizes the choices impact the resulting inference. In particular, the choice of
$\beta=4$ allows the inference machinery to approach the correct $\Cmu$ from
below whereas the choice of $\beta=2$ approaches $\Cmu$ from above; see Figs.
S4 and S5. This behavior, which we believe is desirable, is similar to the
inference dynamics observed for the Golden Mean Process, further strengthening
the apparent suitability of using $\beta=4$.

Unlike the previous example, the start state for the correct structure is
inferred with little data. In this example, the data series begins with the 
symbols $D_{0:T}=10\ldots$, which can only be generated from state $B$.  So,
at $L=2$ the start state for the correct topology is determined, but it takes
more data---$32$ symbols in this case---for this structure to become the most
probable in the set considered.  

\subsubsection{Out-of-Class Structural Inference: The Simple Nonunifilar Source}

\begin{figure}
\centering
\includegraphics{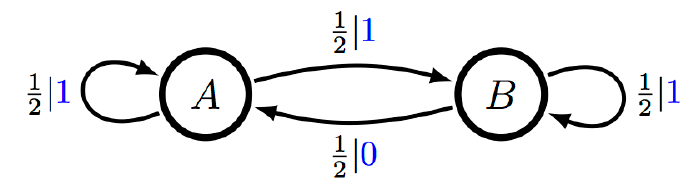}
\caption{The Simple Nonunifilar Source.
  }
\label{fig:sns}
\end{figure}

\begin{figure*}
  \includegraphics[]{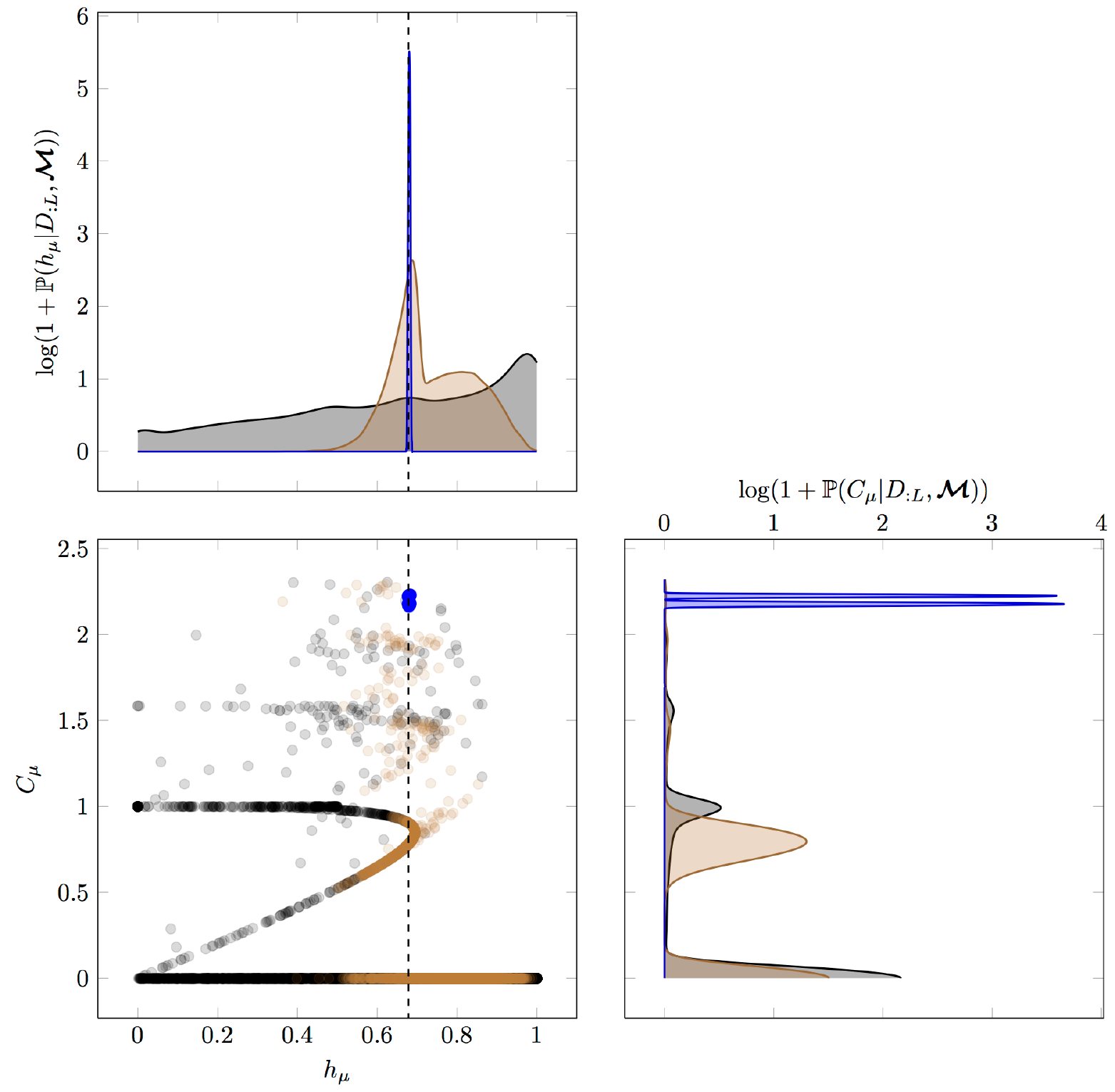}
\caption{Convergence of randomness ($\hmu$) and structure ($\Cmu$) calculated
  with model topologies, transition probabilities, and start states estimated
  from Simple Nonunifilar Source data, using all one- to five-state topological
  \eMs. Sample sizes, colors and line types mirror those in previous figures.
  }
  \label{fig:sns:hmuCmu:structure}
\end{figure*}

The Simple Nonunifilar Source (SNS) is our final and most challenging example
of structural inference due its being out-of-class. The SNS is not only
infinite-order Markov, any unifilar presentation requires a infinite number of
states. In particular, its \eM, the minimal unifilar presentation, has a
countable infinity of causal states \cite{Crut92c}. We can see the difference
between the SNS and previous processes by inspecting state $A$, where both
out-going edges emit a symbol `1'. (See \figref{fig:sns} for a
\HMM\ presentation that is not an \eM.)
This makes the SNS a nonunifilar topology, as the name suggests. Importantly,
even if we assume a start state, there is no longer a single, unique path
through the hidden states for an observed output data series. This is
completely different from  the unifilar examples previously considered, where
an assumed start state and observed data series either determined a unique path
through hidden states or was disallowed.  As a result, the inference tools
developed here cannot use the HMM topology of Fig. \ref{fig:sns}. Concretely,
this class of representation breaks our method for counting transitions. 

Our goal, though, is to use the set of unifilar, topological \eMs\ at our
disposal to infer properties of the Simple Nonunifilar Source. (One reason to
do this is that unifilar models are \emph{required} to calculate $\hmu$.)
Typical data series generated by the SNS model are accepted by many of the
unifilar topologies in $\Mset$ and a posterior distribution over these models
can be calculated.  As with previous examples, we demonstrate estimating
$\hmu$ and $\Cmu$ for the data source. Due to the nonunifilar nature of the
source, we expect $\Cmu$ estimates to increase with the size of the available
data series. However, the ability to estimate $\hmu$ accurately is unclear a
priori. Of course, in this example we cannot find the correct model topology
because infinite structures are not contained in $\Mset$.

Figure \ref{fig:sns:hmuCmu:structure} presents the joint posterior for 
$(\hmu, \Cmu)$ for three subsample lengths. As previously, a single data series
of length $2^{17}$ is generated using the SNS and analysis of subsamples
$D_{0:L}$ are employed to demonstrate convergence. The short subsample
(L=1, black points) is predictably uninteresting, reflecting the the prior
distribution over models.  For subsamples shorter than $L = 64$ the MAP model
is the single-state, two-edge topology. (Denoted n1k2id3 in Table S6.) At
$L = 64$ the Golden Mean Process topology becomes most probable with a
posterior probability of {53.01}\%. The probability of the single-state
topology is still {43.98}\%, though, resulting in $\Cmu$'s strongly bimodal
marginal posterior observed for $L=64$. (See \figref{fig:sns:hmuCmu:structure}
brown points, right panel.) Bimodality also appears in the marginal posterior
for $\hmu$, with the largest peak coming from the two-state topology and the
high entropy rates being contributed by the single-state model.  At large
data size ($L=16,384$, blue points) $\hmu$ has converged on the true value,
while $\Cmu$ has sharp, bimodal peaks due to many nearly equally probable
five-state topologies. Consulting Table S6, we see that the MAP structure for
this value of $L$ has five states (denoted n5k2id22979, there) and a low
posterior probability of only {8.63}\%. Further investigation reveals that
there are four additional \eM\ topologies (making a total of five) with similar
posterior probability. These general details persist for longer subsamples
sequences including the complete data series at length $2^{17}$. Although
estimating $\hmu$ converges smoothly, the inference of structure as reflected
by $\Cmu$ does not show signs of graceful convergence.

We provide supplementary plots in Figs. S7 and S8 that show the convergence 
of $\hmu$ and $\Cmu$ using $\Mset$ and $M_{\mbox{\tiny MAP}}$ for prior
parameters $\beta=4$ and $\beta=2$, respectively. Again, the choice of $\beta$
matters most at small data sizes.  While the $\Cmu$ estimate increases as
function of $L$ for $\beta=4$, the use of $\beta=2$ results in posterior means
for $\Cmu$ that first decrease as function of $L$, then increase.  Again, this 
supports the use of $\beta=4$ for this set of binary-alphabet, topological 
\eMs. The need to employ the complete model set $\Mset$ versus the MAP topology
is most evident at small data sizes; as was also seen in previous examples.
However, the $\Cmu$ inference in this example is more complicated due to the
large number of five-state topologies with roughly equal probability. The MAP 
method selects just one model, of course, and so cannot represent
the posterior distribution's bimodal behavior. Given that the data source is
out-of-class, this trouble is perhaps not surprising. Figure S9 shows samples
from the joint posterior of $(\hmu,\Cmu)$ using only the MAP topology. Using
the latter also suffers from requiring one to select a single exemplar topology
for a posterior distribution that is simply not well represented by a single
\eM.

\section{Discussion}

The examples demonstrated structural inference of unifilar \HMM s using the
set of one- to five-state, binary-alphabet, topological \eMs. We found that
in-class examples, including the Golden Mean and Even Processes, were
effectively and efficiently discovered. That is, the correct topology was accorded
the largest posterior probability and estimates of information coordinates
$\hmu$ and $\Cmu$ were accurate. However, we found that a sufficiently large
value of $\beta$, providing the model size penalty, was key to a conservative
structural inference. Conservative means that $\Cmu$ estimates approach
the true value from below, effectively counteracting the increasing number
of topologies with larger state sets. For the out-of-class example,
given by the Simple Nonunifilar Source, these broader patterns held true.
However, structure could not be captured as reflected in the increasing number
of states inferred as a function of data length.  Also, many topologies had 
relevant posterior probability for the SNS data, reflecting a lack of consensus
and a large degeneracy with regard to structure. This resulted in a multimodal
posterior distribution for $\Cmu$ and a MAP model with very low posterior
probability.

One of the surprises was the number of \emph{accepting} topologies for a given
data set.  By this we mean the number of candidate structures for which the
data series of interest had a valid path through hidden states, resulting in
nonzero posterior probability. In many ways, this aspect of structural
inference mirrors grammatical inference for deterministic finite automaton
(DFA) \cite{Lang1998,delaHiguera2005}.  In the supplementary material we
provide plots for the three processes considered above showing the number of
accepting topologies in the set of one- to five-state \eMs\ used for $\Mset$.
(See Supplemental Fig. S10.)  For all of these topologies, a rapid decline in
the number of accepting topologies occurs for the first $2^{6}$ to $2^{7}$
symbols, followed by a plateau at a set of accepting topologies.  For smaller
topologies, which come from the model class under consideration, this pattern
makes sense.  Often, the smaller topology is embedded within a larger set of
states, some of which are never used. For out-of-class examples like the SNS
this behavior is less transparent.  The rejection of a data series by a given
topology provides a first level of filtering by assigning zero posterior
probability to the structure due to vanishing likelihood of the data given the
model. For the examples given above, of the $36,660$ possible topologies,
$6,225$ accepted Golden Mean data, $3,813$ topologies accepted Even Process
data, and $6,225$ accepted SNS data when the full data series was considered.

In all of the examples the data sources were stationary, so that statistics did
not change over the course of the data series. This is important because
stationarity is built into the model class definition employed: the model
topology and transition probabilities did not depend on time. However, given a
general data series with unknown properties, it is unwise to assume
stationarity holds. How can this be probed? One method is to subdivide the data
into overlapping segments of equal length. Given these, inference
using $\Mset$ or $M_{\mbox{\tiny MAP}}$ should return similar results for each
segment.  For in-class data sources like the Even and Golden Mean Processes,
the true model should be returned for each data subsegment.  For out-of-class,
but stationary models like the Simple Nonunifilar Source, the true topology
cannot be returned, but a consistent model within $\Mset$ should be returned
for each data segment.

However, one form of relatively simple nonstationarity---a structural
change-point problem such as switching between the Golden Mean and Even
Processes---can be detected by BSI applied to subsegments. The inferred
topology for early segments returns the Golden Mean topology and later segments
return the Even topology. Notably, the inferred topology using all of the data
or a subsegment overlapping the switch returns a more complicated model
topology reflecting both structures. Of course, detection of this behavior
requires sufficient data and slow switching between data sources.

In a sequel we compare BSI to alternative structural inference methods. The
range of and differences with these is large and so a comparison demands its
own venue. Also, the sequel addresses expanding the model candidates beyond
the set of topological \eMs\ to the full set of unifilar \HMM s. A necessary
step before useful comparisons can be explored.

\section{Conclusion}

We demonstrated effective and efficient inference of topological \eMs\ using a
library of candidate structures and the tools of Bayesian inference. Several
avenues for further development are immediately obvious.  First, as just noted,
using full unrestricted \eMs---allowing models outside the set of topological
\eMs---is straightforward. This will provide a broad array of candidates within
the more general class of unifilar \HMM s. In the present setting, by way of
contrast, processes with full support (all words allowed) can map only to the
single-state topology. Second, refining the eminently parallelizable Bayesian
Structural Inference algorithms will allow them to take advantage of large
compute clusters and cloud computing to dramatically expand the number of
candidate topologies considered. For comparison, the current implementation
uses nonoptimized Python on a single thread. This configuration (running on
contemporary Linux compute node) takes between $0.6$ and $1.6$ hours, depending
on the number of accepting topologies, to calculate the posterior distribution
over the $36,660$ candidates for a data series of length $2^{17}$. An
additional $10$ to $20$ minutes is needed to generate the $50,000$ samples from
the posterior to estimate functions of model parameters, like $\hmu$ and $\Cmu$.

We note that the methods of Bayesian Structural Inference can be applied to any
set of unifilar \HMM s and, moreover, they do not have to employ a large,
enumerated library.  For example, a small set of candidate fifty-state
topologies could be compared for a given data series.  This ability opens the
door to automated methods for generating candidate structures.  Of course, as
always, one must keep in mind that all inferences are then conditioned on the,
possibly limited or inappropriate, set of model topologies chosen.

Finally, let's return to the scientific and engineering problem areas cited in
the introduction that motivated structural inference in the first place.
Generally, Bayesian Structural Inference will find application in fields, such
as those mentioned, that rely on finite-order Markov chains or the broader
class of (nonunifilar) \HMM s. It will also find application in areas requiring
accurate estimates of various system statistics. The model class considered
here (\eMs) consists of a novel set of topologies and usefully allows one to
estimate both randomness and structure using $\hmu$ and $\Cmu$. Two of the most
basic informational measures. As a result, we expect Bayesian Structural
Inference to find an array of applications in bioinformatics, linguistics, and
dynamical systems.

\section*{Acknowledgments}

The authors thank Ryan James and Chris Ellison for helpful comments and
implementation advice. Partial support was provided by ARO grants
W911NF-12-1-0234 and W911NF-12-1-0288.

\bibliography{chaos}

\pagebreak
\onecolumngrid
\appendix
\renewcommand{\thetable}{S\arabic{table}}
\renewcommand{\thefigure}{S\arabic{figure}}
\setcounter{table}{0}
\setcounter{figure}{0}

\begin{center}
{\huge \bf Supplementary Material}

% Title
\vskip3em
%\begin{flushleft}
{\Large
\textbf{\ourTitle}
}

% Author names, affiliations and corresponding author email.
\vskip1em
Christopher C.~Strelioff and James P. Crutchfield
%\end{flushleft}
\end{center}

%%
%%
%%
% \clearpage

\section{Overview}

The supplementary materials provide tables and figures that lend an in-depth
picture of the Bayesian Structural Inference examples. Unless otherwise noted,
all analyses presented here use the same single data series and parameter
settings detailed in the main text. Please use the main text as the primary
guide.

The first three sections address the Golden Mean, Even, and SNS processes.
Each provides a table of estimates of $\hmu$ and $\Cmu$ using the complete set
$\Mset$ of one- to five-state \eMs\ denoted. Estimates are given for each
subsample length $L=2^{i}$, where $i=0,1,2,\ldots, 17$, as in the main text.
To be clear, this means that we analyze subsamples
$D_{0:L} = x_{0} x_{1} x_{2} \ldots x_{L-1}$ using different initial segments of
a single long data series, allowing for a consistent view of estimate
convergence. For both information-theoretic quantities, we list the posterior
mean and equal-tailed, 95\% credible interval (CI) constructed using the
{2.5}\% and {97.5}\% quantiles estimated from $50,000$ samples of the posterior
distribution. The CI
is denoted by parenthesized number pairs. A second table provides the same
estimates of $\hmu$ and $\Cmu$ using only the $M_{\tiny \mbox{MAP}}$ model.
As a result, this table no longer reflects uncertainty in model topology,
which may be small or large depending on the data and subsample length under
consideration.  An additional column in this second table provides the MAP
topology along with its posterior probability. The latter is denoted in
parentheses.

In addition to the tables of estimates, figures demonstrate the convergence of
$\hmu$ and $\Cmu$ marginal posterior distributions as a function subsample
length
$L$. In this, we consider the difference between posteriors using the complete
set $\Mset$ of candidate models and those that only employ the MAP topology.
This set of figures also illustrates the difference between $\beta=4$ and
$\beta=2$. (We use different data, but still a single time series, for the
$\beta=2$ example.)  In all plots the marginal posterior distribution for the
quantity of interest is estimated using a Gaussian kernel density estimation
(Gkde) of the density using $50,000$ samples from the appropriate density.
If there is little or no variation in the samples the Gkde fails and no density
is drawn. This happens, for example, when the MAP topology has one state, and
$\Cmu=0$, for small data sizes. Posterior samples are valid, however, and
posterior mean and credible interval can be provided (see tables).

\begin{figure}[b]
\begin{minipage}[t]{0.48\textwidth}
  \includegraphics[]{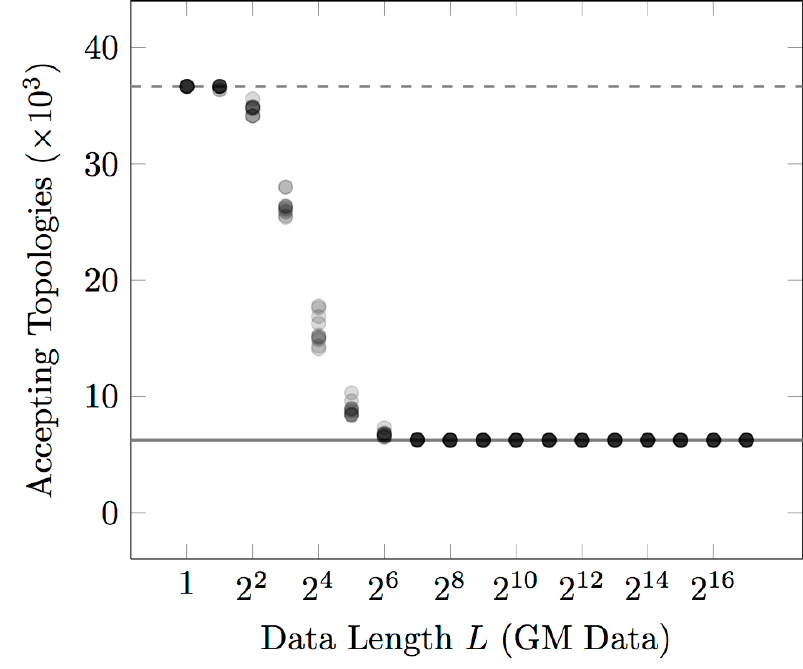}
  \vskip1em
  \includegraphics[]{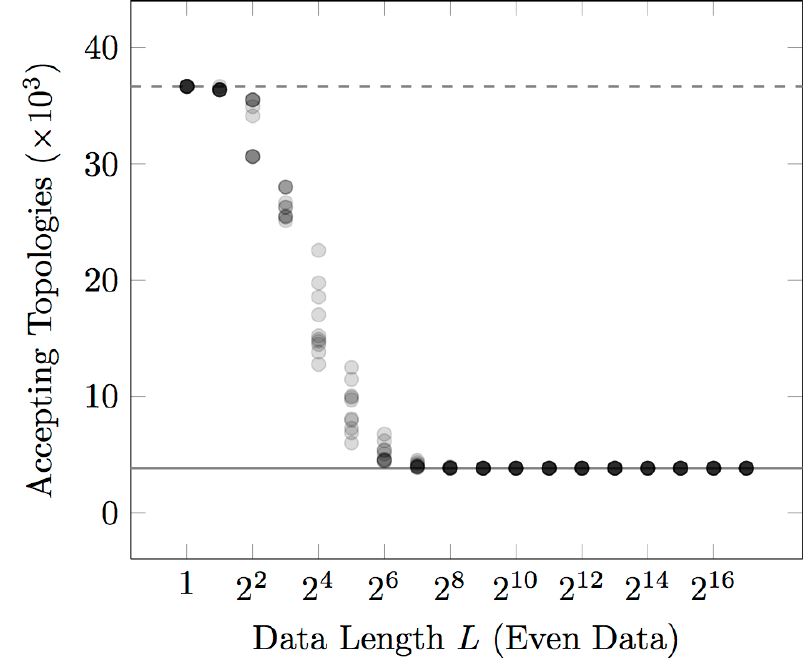}
\end{minipage}
\hfill
\begin{minipage}[t]{0.48\textwidth} 
  \includegraphics[]{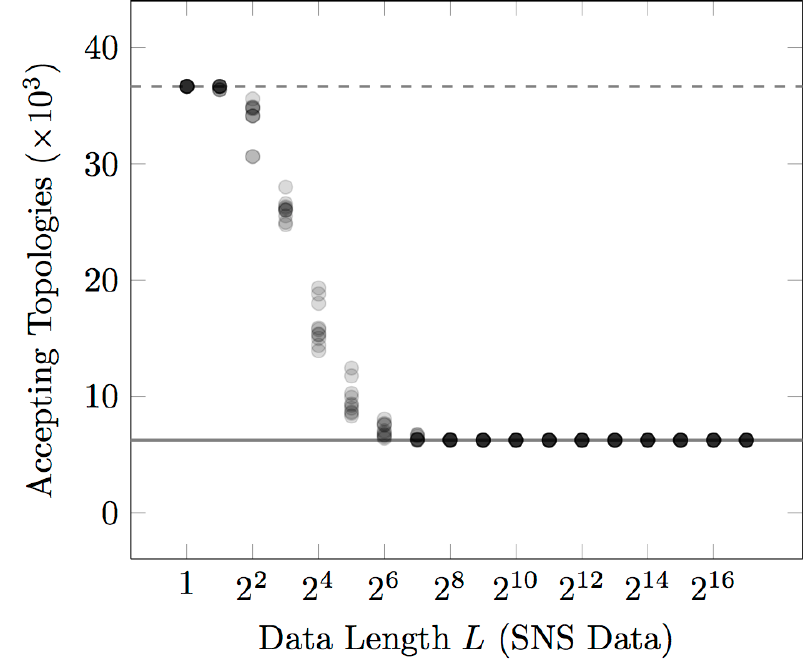}
\end{minipage}
\caption{Number of accepting topologies for each of example processes as a
  function of subsample length $L$. For each, a set of ten data series was
  created and subsamples of length $L$ were analyzed to determined the number
  of binary-alphabet, one- to five-state topological \eMs\ that had at least
  one valid path for that length. (This would result in nonzero likelihood
  and posterior probability.) For each data series, a gray point is plotted.
  Overlapping gray points, created by multiple data series with the same number
  of accepting topologies at the given 
  value of $L$ generate a darker gray or black point.  The horizontal 
  lines indicate the total number of candidate structures ($36,000$, gray 
  dashed line) and the asymptotic number of accepting topologies (solid, gray 
  line).  For the Even Process (bottom, left panel), $3,813$ topologies were 
  asymptotically accepting whereas the Golden Mean Process (top, left panel) 
  and Simple Nonunifilar Source (top, right panel) both had $6,225$.
  \label{fig:accept:top}
  }
\end{figure}

Section \ref{sec:AcceptTopos} plots the number of accepting topologies as a
function of subsample length for each of the example data sources in 
Fig. \ref{fig:accept:top}. The panels demonstrate that there are many valid
candidate topologies for a given data series, even when subsamples
of considerable length are available.

Finally, Sec. \ref{sec:MAPTopos} illustrates all topologies that met the MAP
criterion for the data sources considered. Notably, there are not many 
structures to consider despite the large number of topologies that accept the 
data.

\clearpage
\section{Golden Mean Process: Structural Inference}
\label{sec:GMP}

\begin{table*}[b]
\caption{\label{tab:gm:all}Inference of Golden Mean Process properties using 
$\Mset$, $\beta=4$.}
\begin{ruledtabular}
\begin{tabular}{rcc}
L & $\hmu$ & $\Cmu$ \\ 
1 & 6.767e-01 (3.682e-02,9.994e-01) & 1.467e-01 (0.000e+00,1.333e+00) \\ 
2 & 6.400e-01 (6.662e-02,9.990e-01) & 1.074e-01 (0.000e+00,1.089e+00) \\ 
4 & 7.771e-01 (2.760e-01,9.996e-01) & 1.146e-01 (0.000e+00,1.000e+00) \\ 
8 & 7.753e-01 (3.557e-01,9.994e-01) & 1.441e-01 (0.000e+00,1.000e+00) \\ 
16 & 7.941e-01 (4.751e-01,9.976e-01) & 1.128e-01 (0.000e+00,9.469e-01) \\ 
32 & 7.697e-01 (5.221e-01,9.773e-01) & 2.564e-01 (0.000e+00,1.556e+00) \\ 
64 & 6.440e-01 (5.207e-01,6.942e-01) & 1.052e+00 (8.235e-01,1.797e+00) \\ 
128 & 6.575e-01 (5.953e-01,6.930e-01) & 9.209e-01 (8.667e-01,9.590e-01) \\ 
256 & 6.684e-01 (6.311e-01,6.917e-01) & 9.128e-01 (8.740e-01,9.437e-01) \\ 
512 & 6.718e-01 (6.477e-01,6.889e-01) & 9.107e-01 (8.835e-01,9.338e-01) \\ 
1024 & 6.622e-01 (6.428e-01,6.780e-01) & 9.217e-01 (9.048e-01,9.369e-01) \\ 
2048 & 6.618e-01 (6.483e-01,6.736e-01) & 9.225e-01 (9.107e-01,9.333e-01) \\ 
4096 & 6.587e-01 (6.490e-01,6.678e-01) & 9.253e-01 (9.172e-01,9.329e-01) \\ 
8192 & 6.645e-01 (6.582e-01,6.704e-01) & 9.203e-01 (9.143e-01,9.259e-01) \\ 
16384 & 6.643e-01 (6.599e-01,6.685e-01) & 9.205e-01 (9.164e-01,9.245e-01) \\ 
32768 & 6.647e-01 (6.615e-01,6.676e-01) & 9.202e-01 (9.173e-01,9.231e-01) \\ 
65536 & 6.662e-01 (6.640e-01,6.682e-01) & 9.188e-01 (9.167e-01,9.208e-01) \\ 
131072 & 6.670e-01 (6.655e-01,6.684e-01) & 9.180e-01 (9.165e-01,9.194e-01) \\ 
\end{tabular}
\end{ruledtabular}
\end{table*}

\begin{table*}[b] 
\caption{\label{tab:gm:map}Inference of Golden Mean Process properties using 
$M_{\mbox{MAP}}$, $\beta=4$.}
\begin{ruledtabular}
\begin{tabular}{rccr}
L & $\hmu$ & $\Cmu$ & MAP Topology\\ 
1 & 7.221e-01 (9.729e-02,9.996e-01) & 0.000e+00 (0.000e+00,0.000e+00) & n1k2id3 (8.570e-01)\\ 
2 & 6.603e-01 (6.849e-02,9.992e-01) & 0.000e+00 (0.000e+00,0.000e+00) & n1k2id3 (8.954e-01)\\ 
4 & 8.116e-01 (3.066e-01,9.997e-01) & 0.000e+00 (0.000e+00,0.000e+00) & n1k2id3 (8.896e-01)\\ 
8 & 8.129e-01 (3.811e-01,9.995e-01) & 0.000e+00 (0.000e+00,0.000e+00) & n1k2id3 (8.600e-01)\\ 
16 & 8.141e-01 (4.787e-01,9.981e-01) & 0.000e+00 (0.000e+00,0.000e+00) & n1k2id3 (8.795e-01)\\ 
32 & 8.134e-01 (5.668e-01,9.830e-01) & 0.000e+00 (0.000e+00,0.000e+00) & n1k2id3 (7.324e-01)\\ 
64 & 6.636e-01 (5.842e-01,6.942e-01) & 9.061e-01 (8.188e-01,9.622e-01) & n2k2id5 (7.873e-01)\\ 
128 & 6.577e-01 (5.962e-01,6.929e-01) & 9.198e-01 (8.666e-01,9.583e-01) & n2k2id5 (9.971e-01)\\ 
256 & 6.684e-01 (6.316e-01,6.918e-01) & 9.125e-01 (8.736e-01,9.433e-01) & n2k2id5 (9.987e-01)\\ 
512 & 6.717e-01 (6.477e-01,6.889e-01) & 9.108e-01 (8.836e-01,9.338e-01) & n2k2id5 (9.994e-01)\\ 
1024 & 6.621e-01 (6.429e-01,6.781e-01) & 9.217e-01 (9.046e-01,9.369e-01) & n2k2id5 (9.997e-01)\\ 
2048 & 6.617e-01 (6.481e-01,6.735e-01) & 9.226e-01 (9.108e-01,9.335e-01) & n2k2id5 (9.998e-01)\\ 
4096 & 6.588e-01 (6.491e-01,6.677e-01) & 9.253e-01 (9.172e-01,9.328e-01) & n2k2id5 (9.999e-01)\\ 
8192 & 6.645e-01 (6.582e-01,6.705e-01) & 9.202e-01 (9.143e-01,9.259e-01) & n2k2id5 (1.000e+00)\\ 
16384 & 6.643e-01 (6.599e-01,6.685e-01) & 9.205e-01 (9.164e-01,9.245e-01) & n2k2id5 (1.000e+00)\\ 
32768 & 6.646e-01 (6.616e-01,6.677e-01) & 9.202e-01 (9.173e-01,9.231e-01) & n2k2id5 (1.000e+00)\\ 
65536 & 6.662e-01 (6.640e-01,6.682e-01) & 9.188e-01 (9.167e-01,9.208e-01) & n2k2id5 (1.000e+00)\\ 
131072 & 6.670e-01 (6.655e-01,6.684e-01) & 9.180e-01 (9.165e-01,9.194e-01) & n2k2id5 (1.000e+00)\\ 
\end{tabular}
\end{ruledtabular}
\end{table*}

\begin{figure}[b]
\begin{minipage}{0.48\textwidth}
  \includegraphics[]{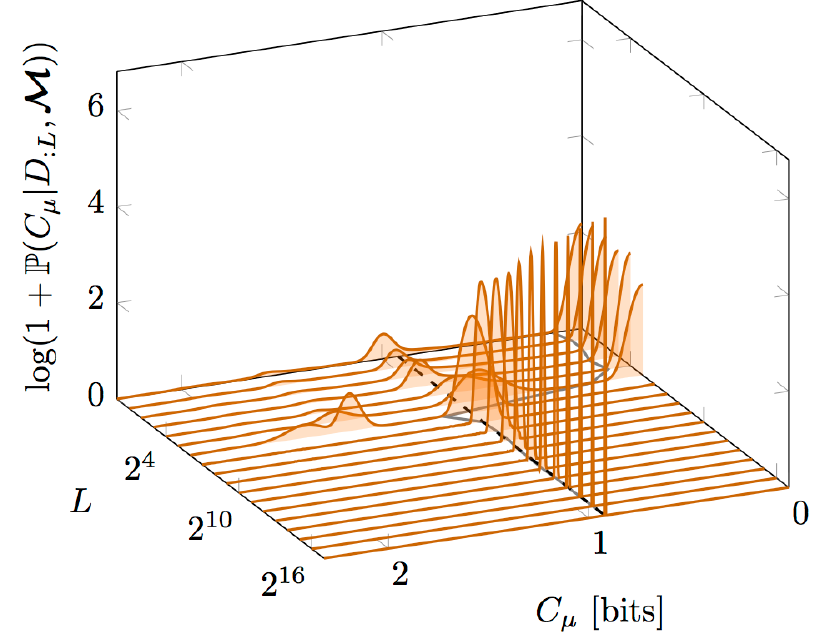}
  \vskip1em
  \includegraphics[]{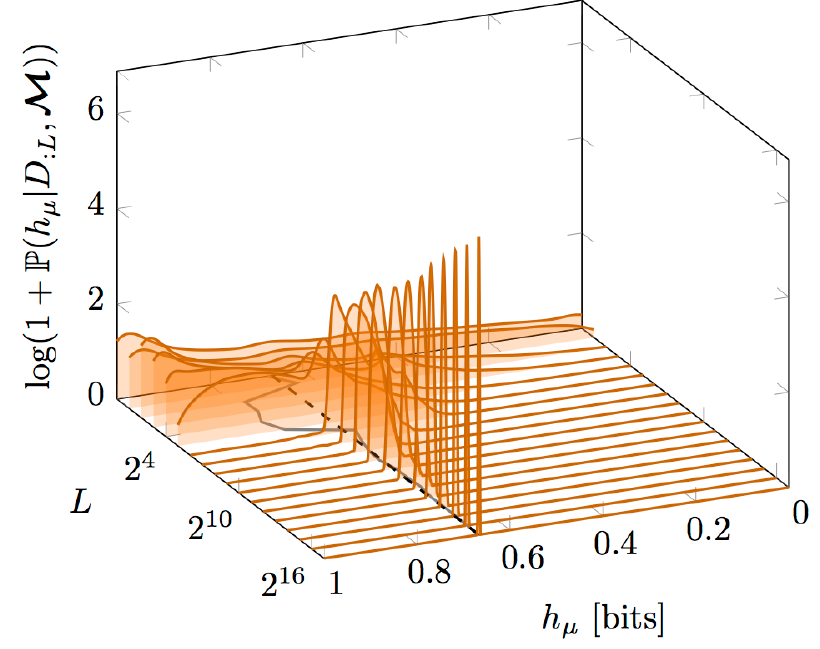}
\end{minipage}
\hfill
\begin{minipage}{0.48\textwidth} 
  \includegraphics[]{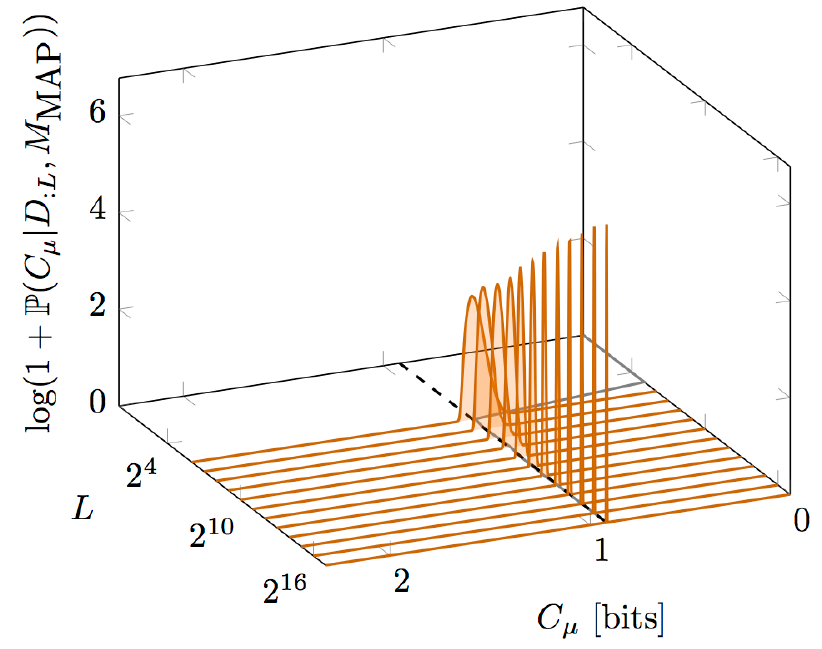}
  \vskip1em
  \includegraphics[]{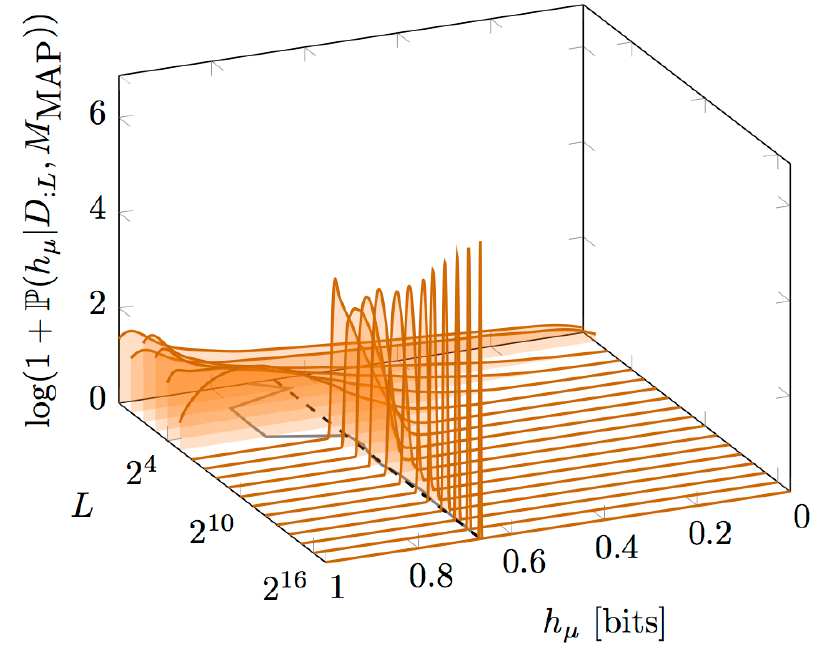}
\end{minipage}
  \caption{Golden Mean Process, $\beta=4$: Convergence of the posterior 
  densities for $\Cmu$ (top) and $\hmu$ (bottom) as a function of subsample 
  length $L$ using the set of all topological \eMs\ with 1-5 states $\Mset$ 
  (left column) and the maximum a posteriori model $M_{\mbox{MAP}}$ (right 
  column).  In each panel, the black, dashed line indicates the true value and 
  the gray, solid line shows the posterior mean.}
  \label{fig:gm:struct:hmuCmu:converge}
\end{figure}

\begin{figure}[b]
\begin{minipage}{0.48\textwidth}
  \includegraphics[]{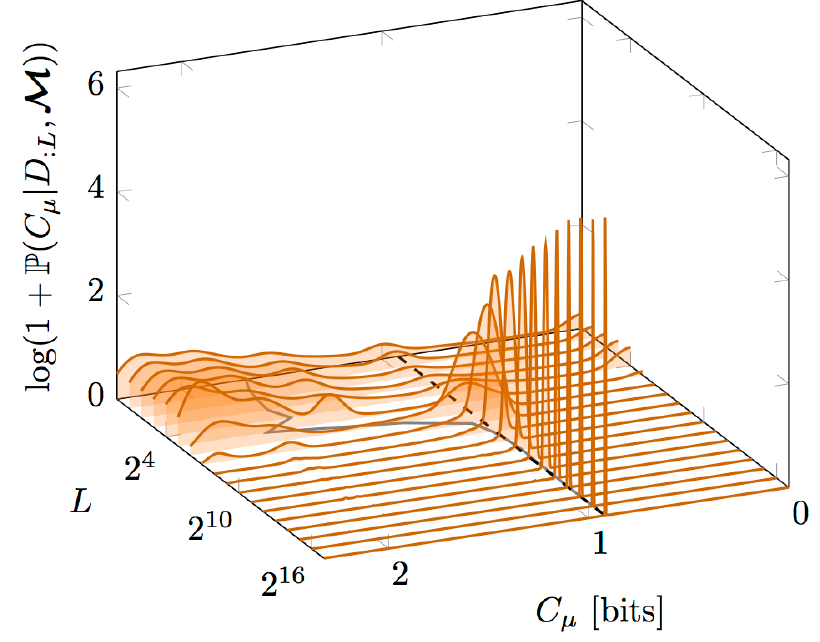}
  \vskip1em
  \includegraphics[]{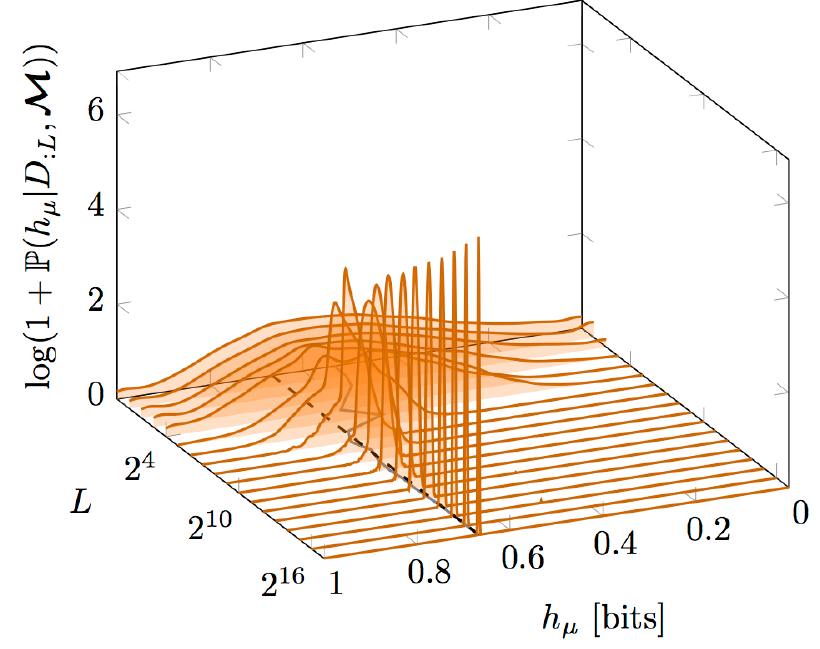}
\end{minipage}
\hfill
\begin{minipage}{0.48\textwidth} 
  \includegraphics[]{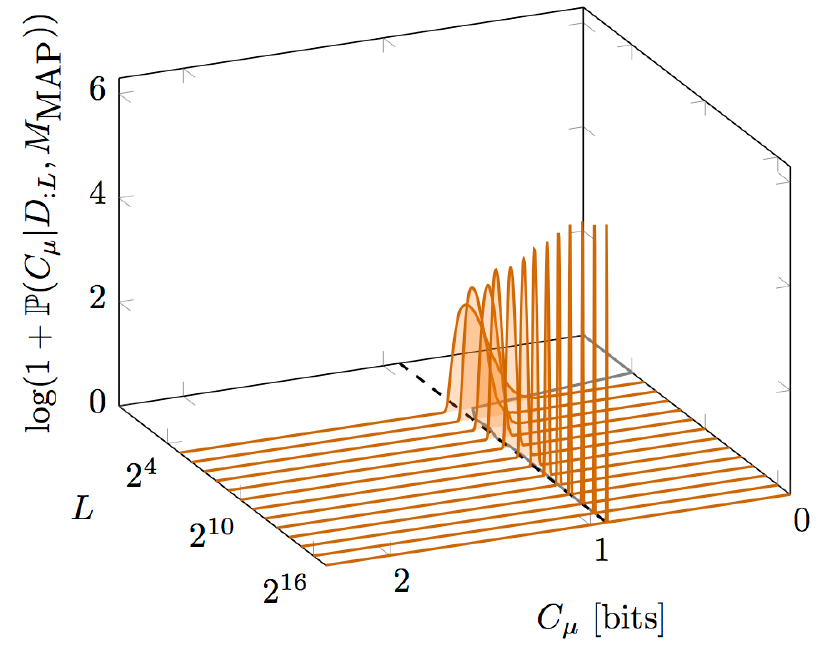}
  \vskip1em
  \includegraphics[]{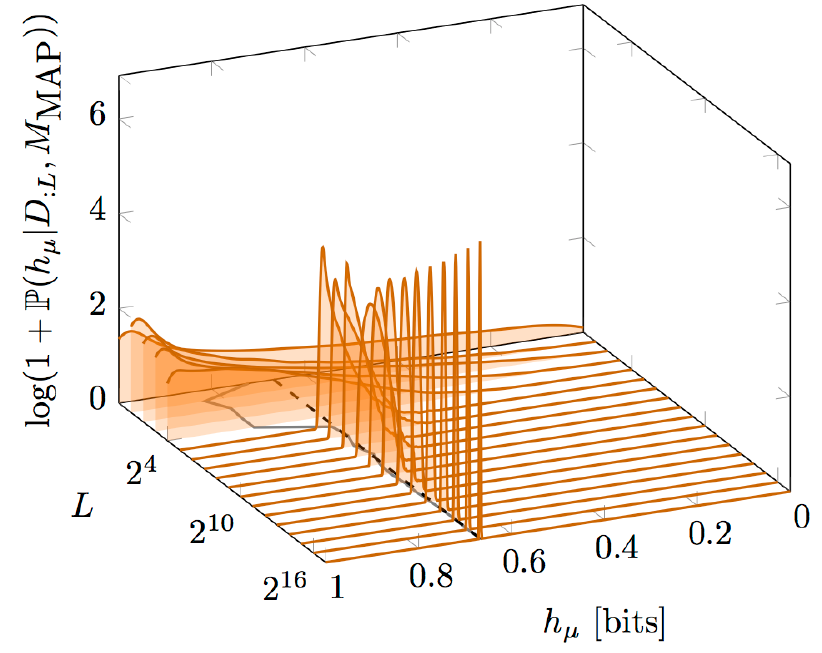}
\end{minipage}
  \caption{Golden Mean Process, $\beta=2$: Convergence of the posterior 
  densities for $\Cmu$ (top) and $\hmu$ (bottom) as a function of subsample 
  length $L$ using the set of all topological \eMs\ with 1-5 states $\Mset$ 
  (left column) and the maximum a posteriori model $M_{\mbox{MAP}}$ (right 
  column).  In each panel, the black, dashed line indicates the true value and 
  the gray, solid line shows the posterior mean.  Contrast these panels with 
  those in Figure \ref{fig:gm:struct:hmuCmu:converge}, where the penalty 
  for structure is higher.}
  \label{fig:gm:struct:hmuCmu:converge:beta2}
\end{figure}

\begin{figure}[b]
  \includegraphics[]{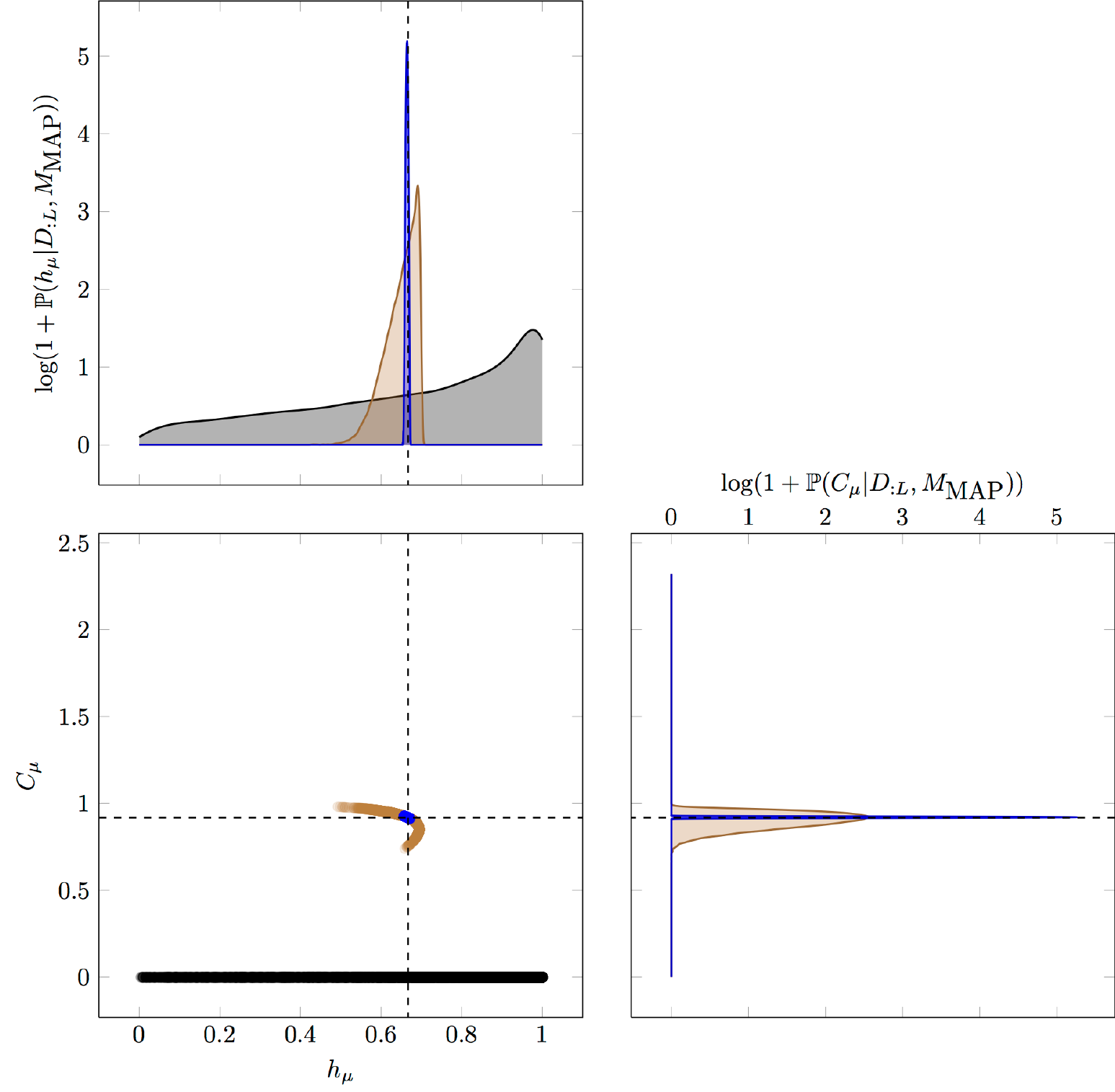}
  \caption{Golden Mean Process: Joint distribution samples using the 
  MAP model at the given lengths instead of the full set of candidate models.  
  Colors correspond to data subsample length, as in previous plots.  The MAP 
  topology for $L=1$ (black) has one state and $\Cmu=0$, as indicated by the 
  samples in the $\hmu-\Cmu$ plane. No Gkde approximation of these samples is 
  provided due to this complete lack of variation.}
  \label{fig:gm:hmuCmu:joint}
\end{figure}

\clearpage
\section{Even Process: Structural Inference}
\label{sec:EP}

\begin{table*}[b]
\caption{\label{tab:even:all}Inference of Even Process properties using 
$\Mset$, $\beta=4$.}
\begin{ruledtabular}
\begin{tabular}{rcc}
L & $\hmu$ & $\Cmu$ \\ 
1 & 6.777e-01 (3.811e-02,9.994e-01) & 1.480e-01 (0.000e+00,1.388e+00) \\ 
2 & 7.414e-01 (0.000e+00,9.997e-01) & 2.222e-01 (0.000e+00,1.528e+00) \\ 
4 & 7.697e-01 (2.359e-01,9.996e-01) & 1.191e-01 (0.000e+00,1.224e+00) \\ 
8 & 8.572e-01 (4.097e-01,9.998e-01) & 1.249e-01 (0.000e+00,1.422e+00) \\ 
16 & 8.235e-01 (4.751e-01,9.998e-01) & 3.080e-01 (0.000e+00,9.454e-01) \\ 
32 & 6.457e-01 (4.655e-01,9.616e-01) & 6.909e-01 (0.000e+00,8.961e-01) \\ 
64 & 6.804e-01 (6.276e-01,6.942e-01) & 8.746e-01 (7.675e-01,9.464e-01) \\ 
128 & 6.824e-01 (6.453e-01,6.942e-01) & 8.854e-01 (8.166e-01,9.359e-01) \\ 
256 & 6.783e-01 (6.485e-01,6.939e-01) & 8.993e-01 (8.568e-01,9.333e-01) \\ 
512 & 6.679e-01 (6.422e-01,6.868e-01) & 9.151e-01 (8.890e-01,9.374e-01) \\ 
1024 & 6.756e-01 (6.602e-01,6.874e-01) & 9.069e-01 (8.875e-01,9.243e-01) \\ 
2048 & 6.700e-01 (6.581e-01,6.801e-01) & 9.144e-01 (9.016e-01,9.260e-01) \\ 
4096 & 6.666e-01 (6.578e-01,6.744e-01) & 9.181e-01 (9.096e-01,9.263e-01) \\ 
8192 & 6.704e-01 (6.647e-01,6.757e-01) & 9.142e-01 (9.080e-01,9.202e-01) \\ 
16384 & 6.666e-01 (6.623e-01,6.707e-01) & 9.183e-01 (9.141e-01,9.225e-01) \\ 
32768 & 6.660e-01 (6.629e-01,6.689e-01) & 9.189e-01 (9.160e-01,9.219e-01) \\ 
65536 & 6.657e-01 (6.635e-01,6.677e-01) & 9.193e-01 (9.172e-01,9.213e-01) \\ 
131072 & 6.658e-01 (6.643e-01,6.672e-01) & 9.192e-01 (9.177e-01,9.206e-01) \\ 
\end{tabular}
\end{ruledtabular}
\end{table*}

\begin{table*}[b] 
\caption{\label{tab:even:map}Inference of Even Process properties using 
$M_{\mbox{MAP}}$, $\beta=4$.}
\begin{ruledtabular}
\begin{tabular}{rccr}
L & $\hmu$ & $\Cmu$ & MAP Topology\\ 
1 & 7.226e-01 (1.003e-01,9.996e-01) & 0.000e+00 (0.000e+00,0.000e+00) & n1k2id3 (8.570e-01)\\ 
2 & 8.426e-01 (3.541e-01,9.998e-01) & 0.000e+00 (0.000e+00,0.000e+00) & n1k2id3 (7.893e-01)\\ 
4 & 8.100e-01 (2.982e-01,9.997e-01) & 0.000e+00 (0.000e+00,0.000e+00) & n1k2id3 (8.721e-01)\\ 
8 & 9.027e-01 (5.764e-01,9.999e-01) & 0.000e+00 (0.000e+00,0.000e+00) & n1k2id3 (8.626e-01)\\ 
16 & 9.517e-01 (7.735e-01,9.999e-01) & 0.000e+00 (0.000e+00,0.000e+00) & n1k2id3 (6.023e-01)\\ 
32 & 6.316e-01 (4.650e-01,6.941e-01) & 7.152e-01 (4.825e-01,8.861e-01) & n2k2id7 (9.434e-01)\\ 
64 & 6.802e-01 (6.282e-01,6.942e-01) & 8.728e-01 (7.690e-01,9.445e-01) & n2k2id7 (9.941e-01)\\ 
128 & 6.823e-01 (6.456e-01,6.942e-01) & 8.845e-01 (8.165e-01,9.351e-01) & n2k2id7 (9.973e-01)\\ 
256 & 6.783e-01 (6.483e-01,6.939e-01) & 8.991e-01 (8.566e-01,9.334e-01) & n2k2id7 (9.989e-01)\\ 
512 & 6.681e-01 (6.426e-01,6.869e-01) & 9.149e-01 (8.887e-01,9.370e-01) & n2k2id7 (9.995e-01)\\ 
1024 & 6.757e-01 (6.604e-01,6.873e-01) & 9.068e-01 (8.878e-01,9.241e-01) & n2k2id7 (9.997e-01)\\ 
2048 & 6.700e-01 (6.581e-01,6.801e-01) & 9.143e-01 (9.017e-01,9.260e-01) & n2k2id7 (9.999e-01)\\ 
4096 & 6.666e-01 (6.579e-01,6.744e-01) & 9.181e-01 (9.096e-01,9.262e-01) & n2k2id7 (9.999e-01)\\ 
8192 & 6.704e-01 (6.647e-01,6.757e-01) & 9.142e-01 (9.080e-01,9.202e-01) & n2k2id7 (1.000e+00)\\ 
16384 & 6.666e-01 (6.623e-01,6.707e-01) & 9.183e-01 (9.141e-01,9.224e-01) & n2k2id7 (1.000e+00)\\ 
32768 & 6.660e-01 (6.629e-01,6.689e-01) & 9.189e-01 (9.160e-01,9.219e-01) & n2k2id7 (1.000e+00)\\ 
65536 & 6.657e-01 (6.635e-01,6.678e-01) & 9.193e-01 (9.172e-01,9.213e-01) & n2k2id7 (1.000e+00)\\ 
131072 & 6.658e-01 (6.642e-01,6.672e-01) & 9.192e-01 (9.177e-01,9.207e-01) & n2k2id7 (1.000e+00)\\ 
\end{tabular}
\end{ruledtabular}
\end{table*}

\begin{figure}[b]
\begin{minipage}{0.48\textwidth}
  \includegraphics[]{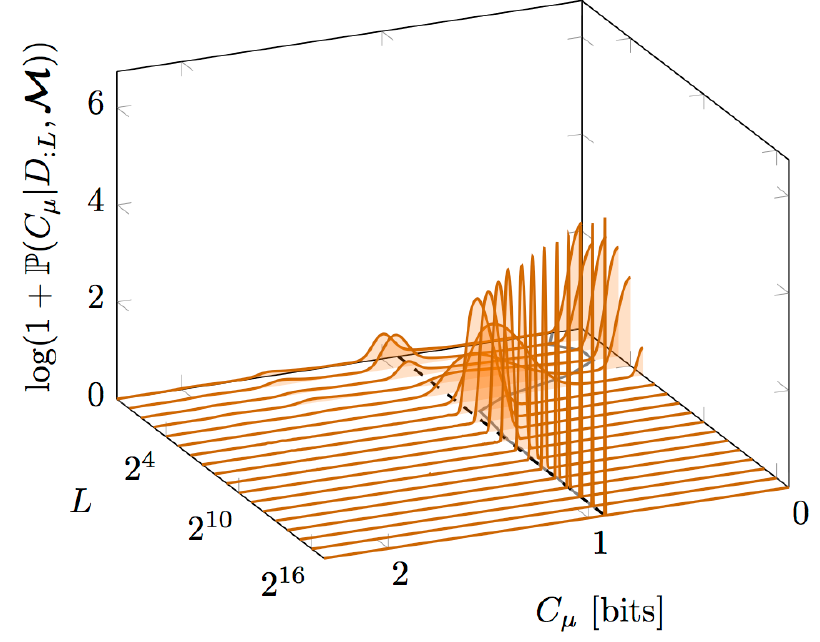}
  \vskip1em
  \includegraphics[]{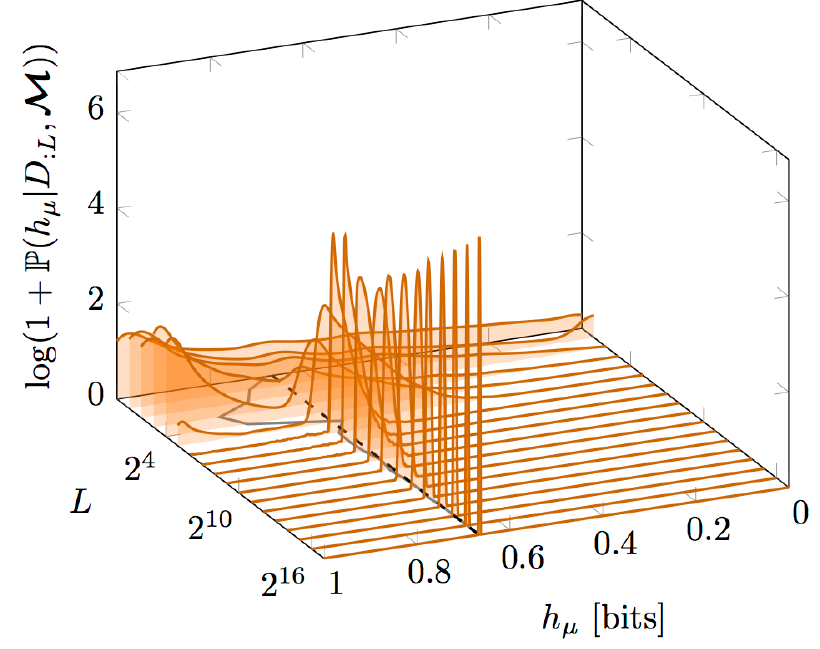}
\end{minipage}
\hfill
\begin{minipage}{0.48\textwidth} 
  \includegraphics[]{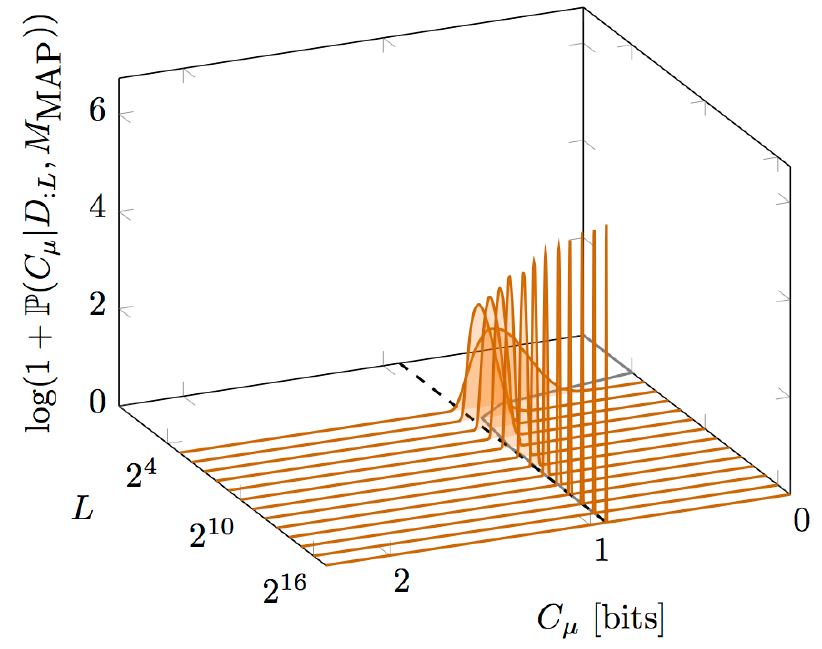}
  \vskip1em
  \includegraphics[]{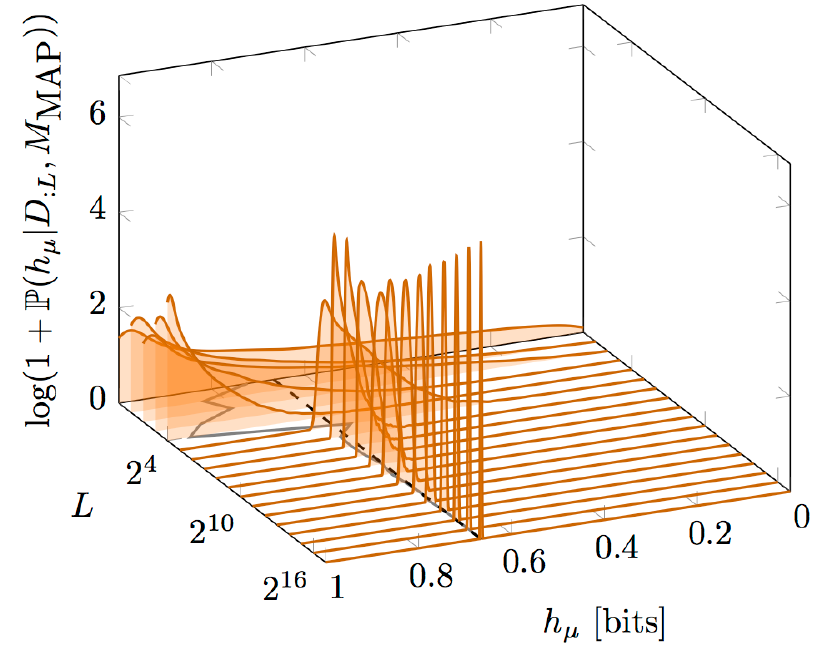}
\end{minipage}
  \caption{Even Process, $\beta=4$: Convergence of the posterior densities for 
  $\Cmu$ (top) and $\hmu$ (bottom) as a function of subsample length $L$ using 
  the set of all topological \eMs\ with one- to five-states $\Mset$ (left 
  column) and the maximum a posteriori model $M_{\mbox{MAP}}$ (right column).  
  In each panel, the black, dashed line indicates the true value and the gray, 
  solid line shows the posterior mean.}
  \label{fig:even:struct:hmuCmu:converge}
\end{figure}

\begin{figure}[b]
\begin{minipage}{0.48\textwidth}
  \includegraphics[]{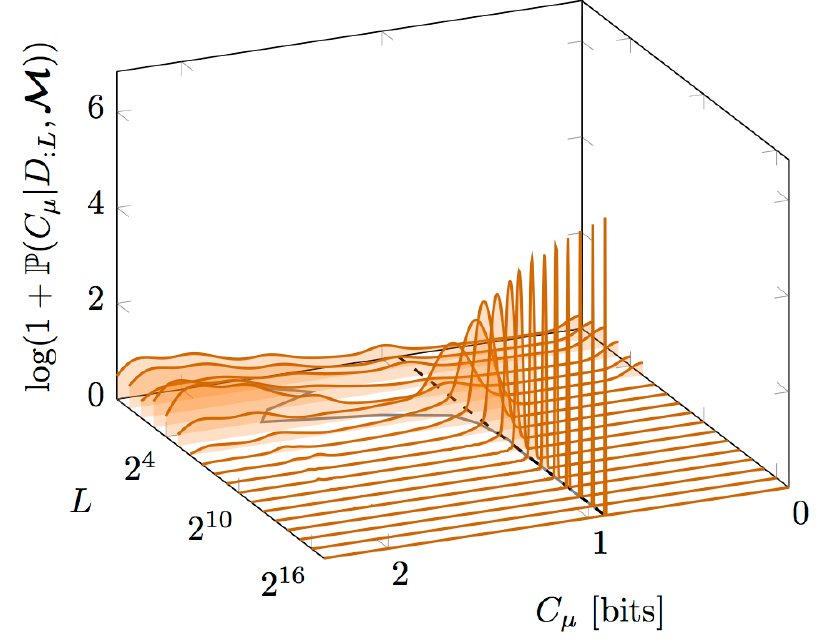}
  \vskip1em
  \includegraphics[]{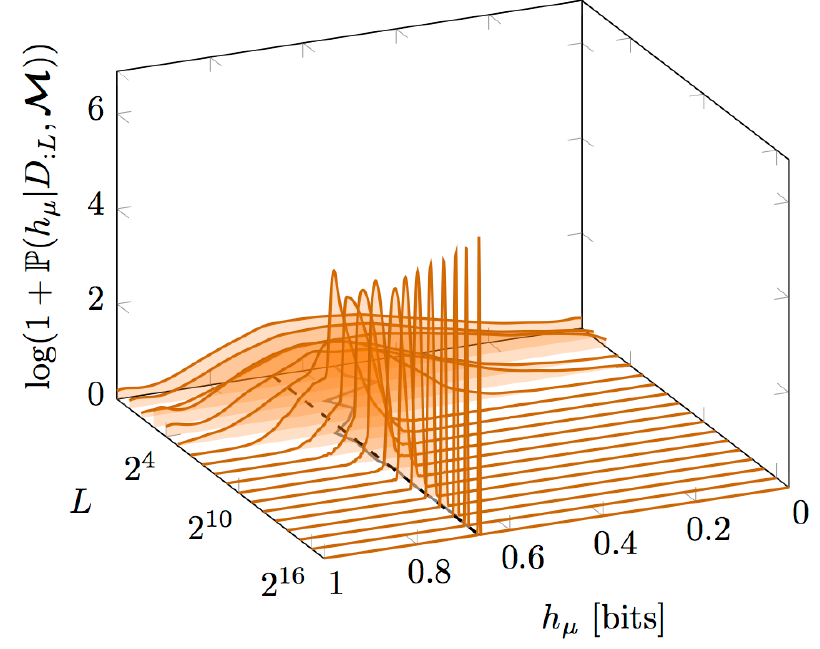}
\end{minipage}
\hfill
\begin{minipage}{0.48\textwidth} 
  \includegraphics[]{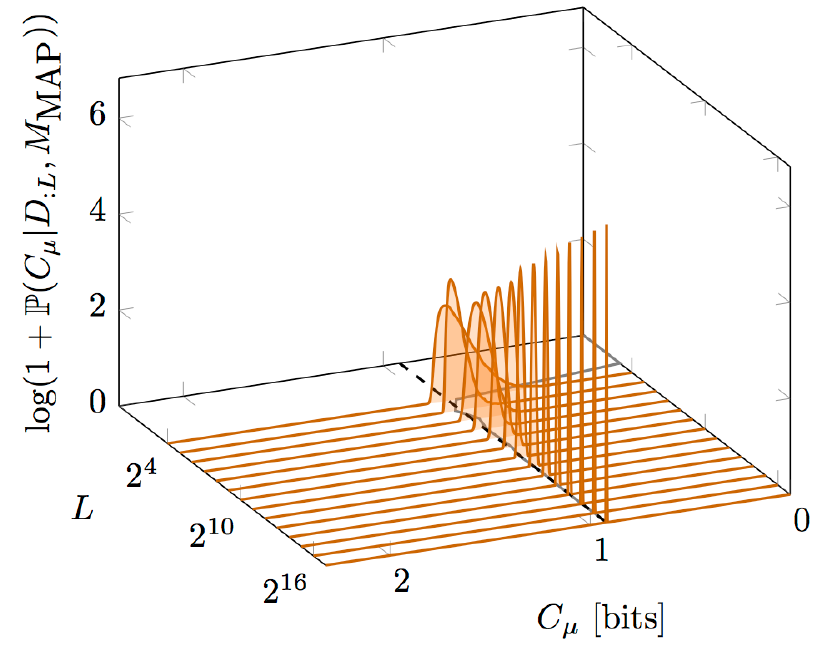}
  \vskip1em
  \includegraphics[]{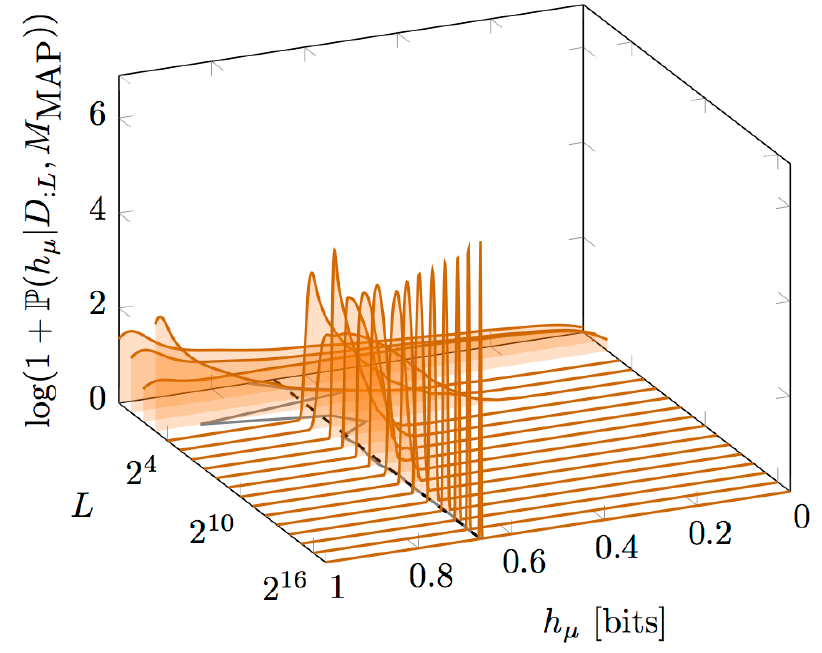}
\end{minipage}
  \caption{Even Process, $\beta=2$: Convergence of the posterior densities for 
  $\Cmu$ (top) and $\hmu$ (bottom) as a function of subsample length $L$ using 
  the set of all topological \eMs\ with one- to five-states $\Mset$ (left 
  column) and the maximum a posteriori model $M_{\mbox{MAP}}$ (right column).  
  In each panel, the black, dashed line indicates the true value and the gray, 
  solid line shows the posterior mean.}
  \label{fig:even:struct:hmuCmu:converge:beta2}
\end{figure}

\begin{figure}[b]
  \includegraphics[]{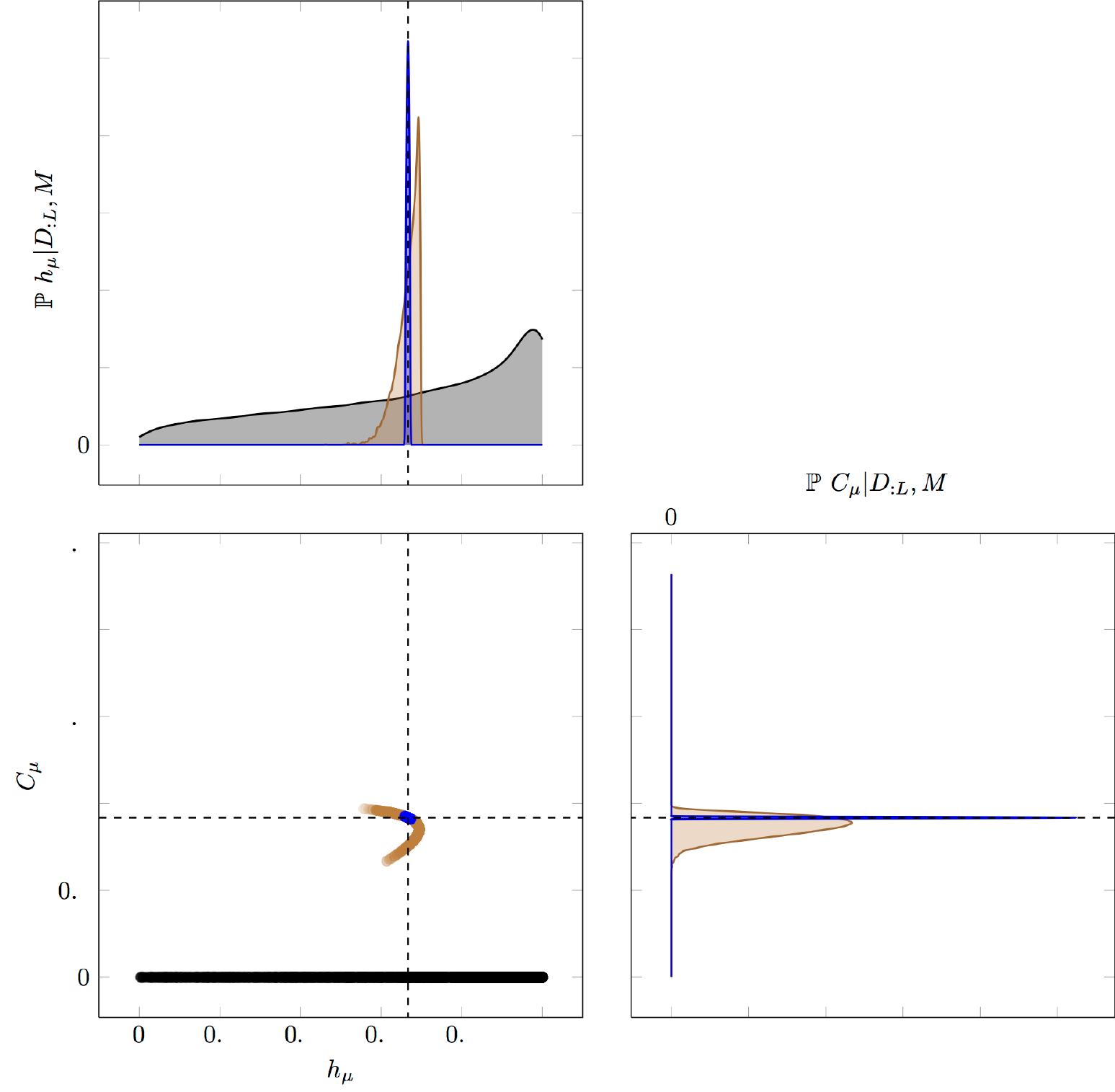}
  \caption{Even Process: Samples of the joint distribution using the 
  MAP model at the given lengths instead of the full set of candidate models.  
  Colors correspond to data subsample length, as in previous plots.  The MAP 
  topology for $L=1$ (black) has one state and $\Cmu=0$, as indicated by the 
  samples in the $\hmu-\Cmu$ plane. No Gkde approximation of these samples is 
  provided due to this complete lack of variation.}
  \label{fig:even:hmuCmu:joint}
\end{figure}

%%
%%%%%%%%%%%%%%%%%%%%%%%%%%%%%%%%%%%%%%%%%%%%%%%%%%%%%%%%%%%%%%%%%%%%%%%%%%%%%%%%
%%
\clearpage
\section{SNS Process: Structural Inference}
\label{sec:SNS}

\begin{table*}[b]
\caption{\label{tab:sns:all}Inference of SNS Process properties using $\Mset$,
$\beta=4$.}
\begin{ruledtabular}
\begin{tabular}{rcc}
L & $\hmu$ & $\Cmu$ \\ 
1 & 6.780e-01 (3.817e-02,9.993e-01) & 1.483e-01 (0.000e+00,1.325e+00) \\ 
2 & 7.425e-01 (0.000e+00,9.997e-01) & 2.207e-01 (0.000e+00,1.525e+00) \\ 
4 & 7.698e-01 (2.398e-01,9.997e-01) & 1.207e-01 (0.000e+00,1.225e+00) \\ 
8 & 7.781e-01 (3.449e-01,9.994e-01) & 1.326e-01 (0.000e+00,1.357e+00) \\ 
16 & 7.952e-01 (2.702e-01,9.994e-01) & 3.679e-01 (0.000e+00,2.084e+00) \\ 
32 & 7.555e-01 (4.978e-01,9.605e-01) & 8.161e-02 (0.000e+00,8.579e-01) \\ 
64 & 7.228e-01 (5.935e-01,9.142e-01) & 4.627e-01 (0.000e+00,1.043e+00) \\ 
128 & 6.808e-01 (6.365e-01,6.942e-01) & 8.006e-01 (6.982e-01,8.808e-01) \\ 
256 & 6.756e-01 (6.411e-01,6.937e-01) & 7.801e-01 (7.088e-01,8.407e-01) \\ 
512 & 6.799e-01 (6.562e-01,6.929e-01) & 8.151e-01 (7.419e-01,1.390e+00) \\ 
1024 & 6.849e-01 (6.693e-01,6.931e-01) & 9.021e-01 (7.717e-01,1.757e+00) \\ 
2048 & 6.827e-01 (6.701e-01,6.922e-01) & 1.441e+00 (7.905e-01,2.219e+00) \\ 
4096 & 6.825e-01 (6.756e-01,6.896e-01) & 1.787e+00 (1.673e+00,2.228e+00) \\ 
8192 & 6.828e-01 (6.782e-01,6.874e-01) & 2.002e+00 (1.692e+00,2.233e+00) \\ 
16384 & 6.800e-01 (6.769e-01,6.832e-01) & 2.198e+00 (2.168e+00,2.231e+00) \\ 
32768 & 6.789e-01 (6.766e-01,6.811e-01) & 2.197e+00 (2.170e+00,2.229e+00) \\ 
65536 & 6.784e-01 (6.769e-01,6.800e-01) & 2.199e+00 (2.174e+00,2.228e+00) \\ 
131072 & 6.788e-01 (6.777e-01,6.799e-01) & 2.201e+00 (2.178e+00,2.230e+00) \\ 
\end{tabular}
\end{ruledtabular}
\end{table*}

\begin{table*}[b] 
\caption{\label{tab:sns:map}Inference of SNS Process properties using 
$M_{\mbox{MAP}}$, $\beta=4$.}
\begin{ruledtabular}
\begin{tabular}{rccr}
L & $\hmu$ & $\Cmu$ & MAP Topology\\ 
1 & 7.231e-01 (9.607e-02,9.996e-01) & 0.000e+00 (0.000e+00,0.000e+00) & n1k2id3 (8.570e-01)\\ 
2 & 8.414e-01 (3.462e-01,9.998e-01) & 0.000e+00 (0.000e+00,0.000e+00) & n1k2id3 (7.893e-01)\\ 
4 & 8.086e-01 (2.981e-01,9.997e-01) & 0.000e+00 (0.000e+00,0.000e+00) & n1k2id3 (8.721e-01)\\ 
8 & 8.136e-01 (3.826e-01,9.996e-01) & 0.000e+00 (0.000e+00,0.000e+00) & n1k2id3 (8.829e-01)\\ 
16 & 8.800e-01 (5.927e-01,9.997e-01) & 0.000e+00 (0.000e+00,0.000e+00) & n1k2id3 (7.774e-01)\\ 
32 & 7.665e-01 (5.040e-01,9.641e-01) & 0.000e+00 (0.000e+00,0.000e+00) & n1k2id3 (9.105e-01)\\ 
64 & 6.712e-01 (5.947e-01,6.942e-01) & 7.842e-01 (6.406e-01,8.918e-01) & n2k2id5 (5.301e-01)\\ 
128 & 6.803e-01 (6.370e-01,6.942e-01) & 7.981e-01 (7.021e-01,8.756e-01) & n2k2id5 (9.835e-01)\\ 
256 & 6.755e-01 (6.408e-01,6.937e-01) & 7.786e-01 (7.083e-01,8.393e-01) & n2k2id5 (9.953e-01)\\ 
512 & 6.804e-01 (6.600e-01,6.928e-01) & 7.887e-01 (7.416e-01,8.313e-01) & n2k2id5 (9.721e-01)\\ 
1024 & 6.858e-01 (6.746e-01,6.929e-01) & 8.029e-01 (7.714e-01,8.321e-01) & n2k2id5 (8.989e-01)\\ 
2048 & 6.871e-01 (6.801e-01,6.922e-01) & 8.066e-01 (7.848e-01,8.273e-01) & n2k2id5 (3.419e-01)\\ 
4096 & 6.826e-01 (6.760e-01,6.893e-01) & 1.703e+00 (1.672e+00,1.733e+00) & n4k2id3334 (1.336e-01)\\ 
8192 & 6.834e-01 (6.792e-01,6.877e-01) & 1.709e+00 (1.687e+00,1.730e+00) & n4k2id3334 (6.462e-02)\\ 
16384 & 6.800e-01 (6.769e-01,6.831e-01) & 2.177e+00 (2.166e+00,2.188e+00) & n5k2id22979 (8.630e-02)\\ 
32768 & 6.789e-01 (6.766e-01,6.810e-01) & 2.176e+00 (2.169e+00,2.184e+00) & n5k2id22979 (8.632e-02)\\ 
65536 & 6.784e-01 (6.769e-01,6.799e-01) & 2.178e+00 (2.173e+00,2.184e+00) & n5k2id22979 (8.560e-02)\\ 
131072 & 6.788e-01 (6.777e-01,6.798e-01) & 2.181e+00 (2.177e+00,2.185e+00) & n5k2id22979 (8.539e-02)\\ 
\end{tabular}
\end{ruledtabular}
\end{table*}

\begin{figure}[b]
\begin{minipage}{0.48\textwidth}
  \includegraphics[]{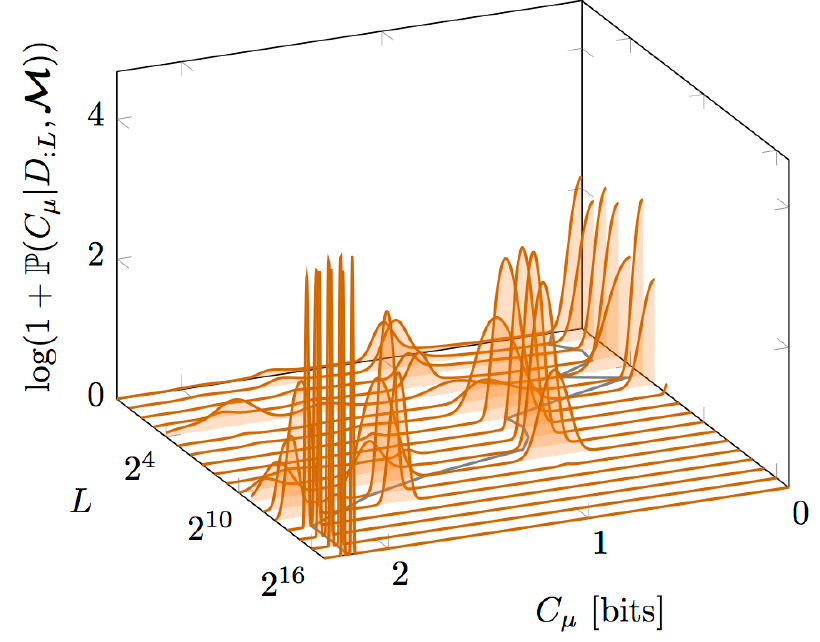}
  \vskip1em
  \includegraphics[]{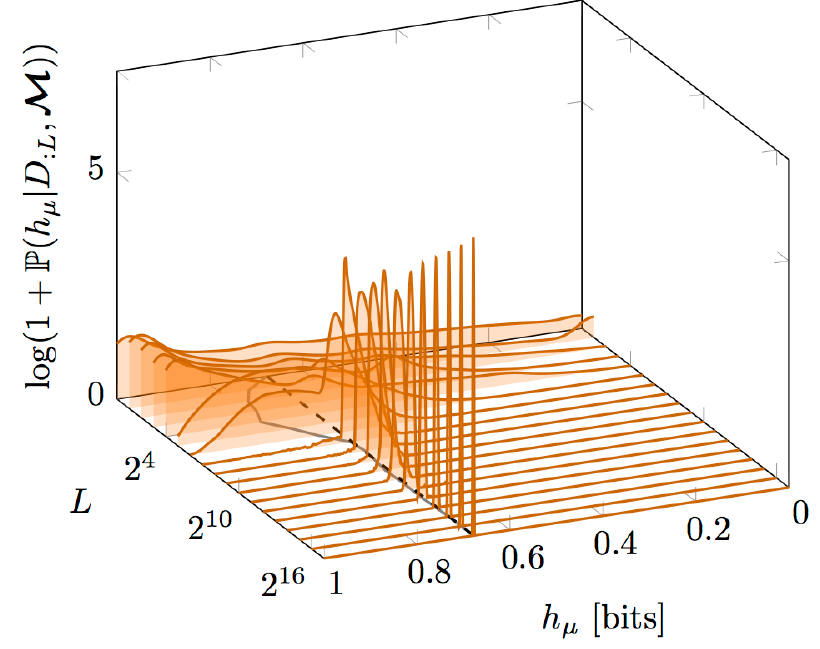}
\end{minipage}
\hfill
\begin{minipage}{0.48\textwidth} 
  \includegraphics[]{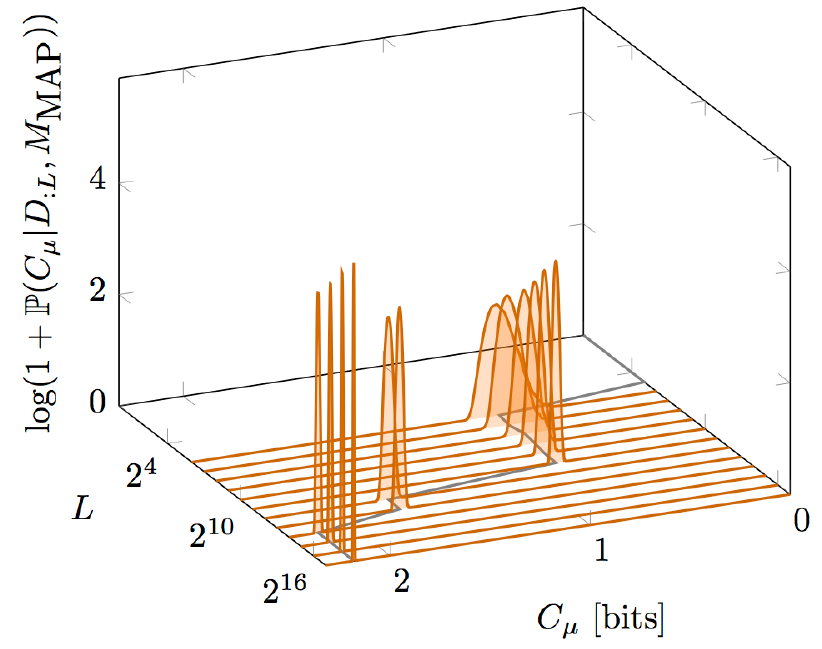}
  \vskip1em
  \includegraphics[]{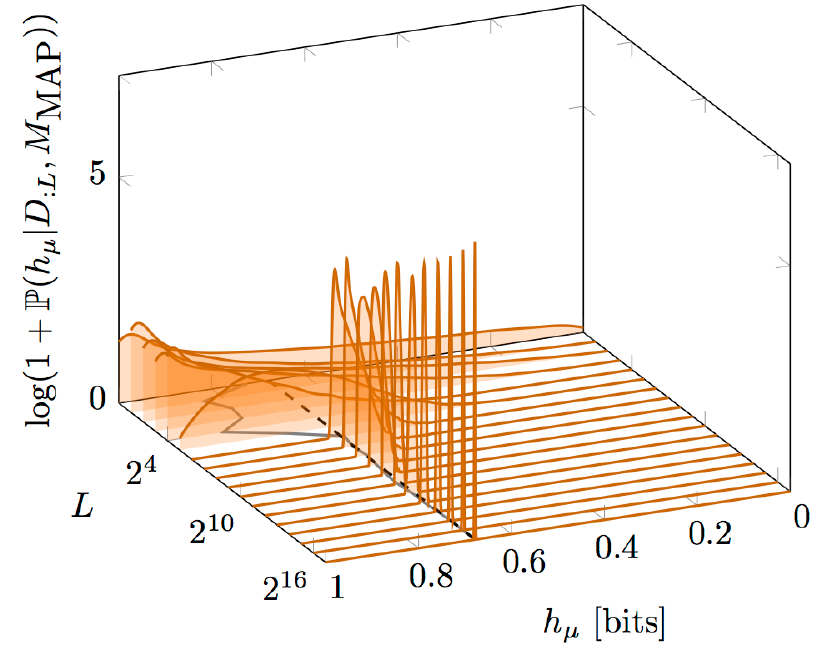}
\end{minipage}
  \caption{Simple Nonunifilar Source, $\beta=4$: Convergence of the posterior 
  densities for $\Cmu$ (top) and $\hmu$ (bottom) as a function of subsample 
  length $L$ using the set of all topological \eMs\ with one- to five-states 
  $\Mset$ (left column) and the maximum a posteriori model $M_{\mbox{MAP}}$ 
  (right column).  In each panel, the black, dashed line indicates the true 
  value and the gray, solid line shows the posterior mean.}
  \label{fig:sns:struct:hmuCmu:converge}
\end{figure}

\begin{figure}[b]
\begin{minipage}{0.48\textwidth}
  \includegraphics[]{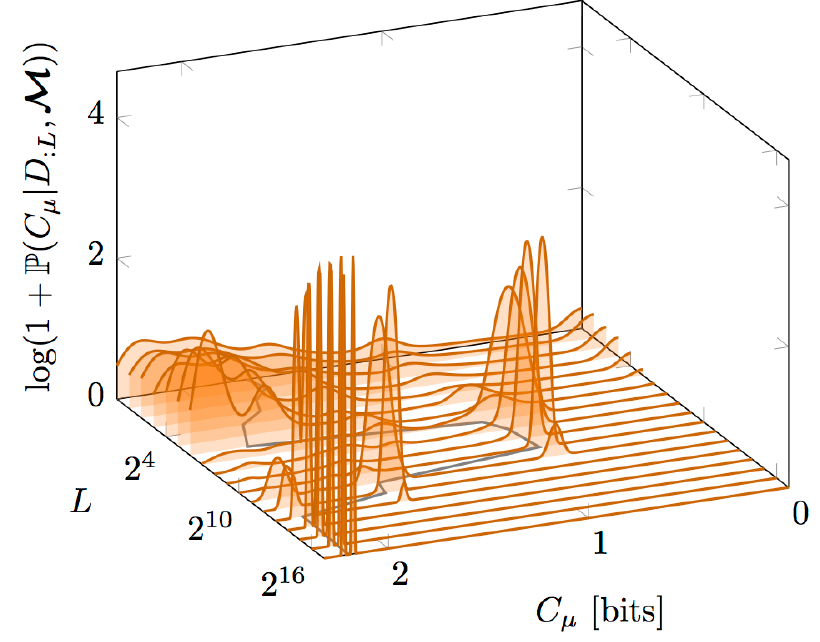}
  \vskip1em
  \includegraphics[]{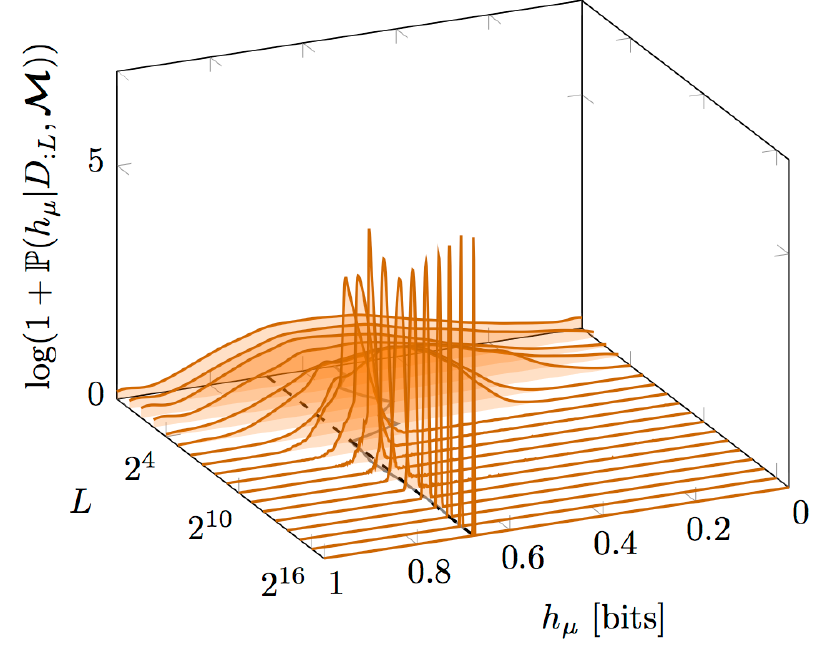}
\end{minipage}
\hfill
\begin{minipage}{0.48\textwidth} 
  \includegraphics[]{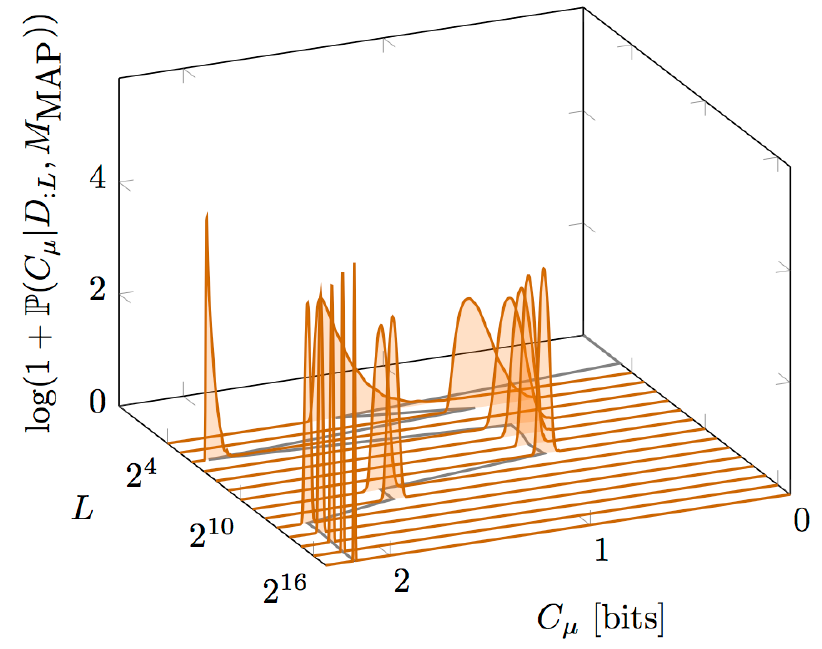}
  \vskip1em
  \includegraphics[]{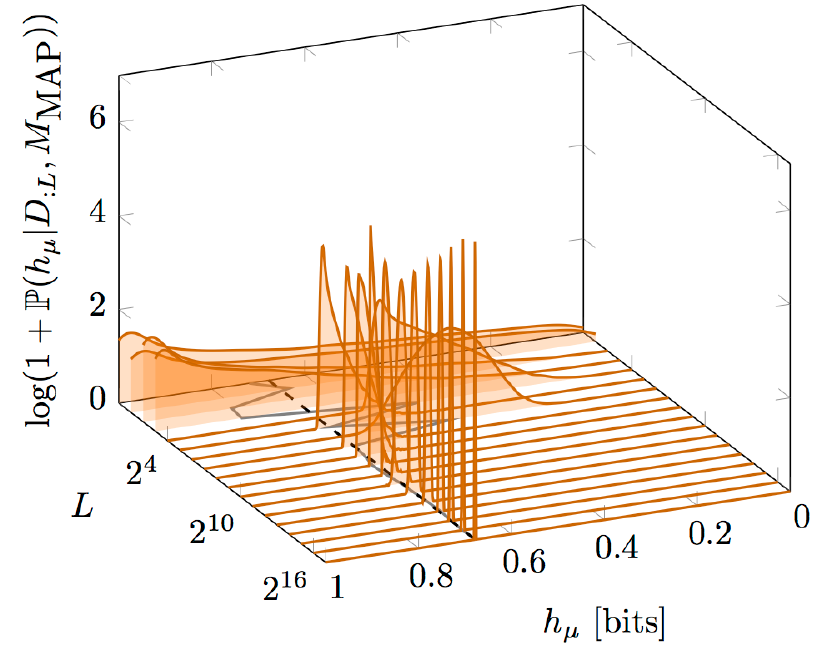}
\end{minipage}
\caption{Simple Nonunifilar Source, $\beta=2$: Convergence of the posterior 
  densities for $\Cmu$ (top) and $\hmu$ (bottom) as a function of subsample 
  length $L$ using the set of all topological \eMs\ with one- to five-states 
  $\Mset$ (left column) and the maximum a posteriori model $M_{\mbox{MAP}}$ 
  (right column).  In each panel, the black, dashed line indicates the true 
  value and the gray, solid line shows the posterior mean.
  }
\label{fig:sns:struct:hmuCmu:converge:beta2}
\end{figure}

\begin{figure}[b]
  \includegraphics[]{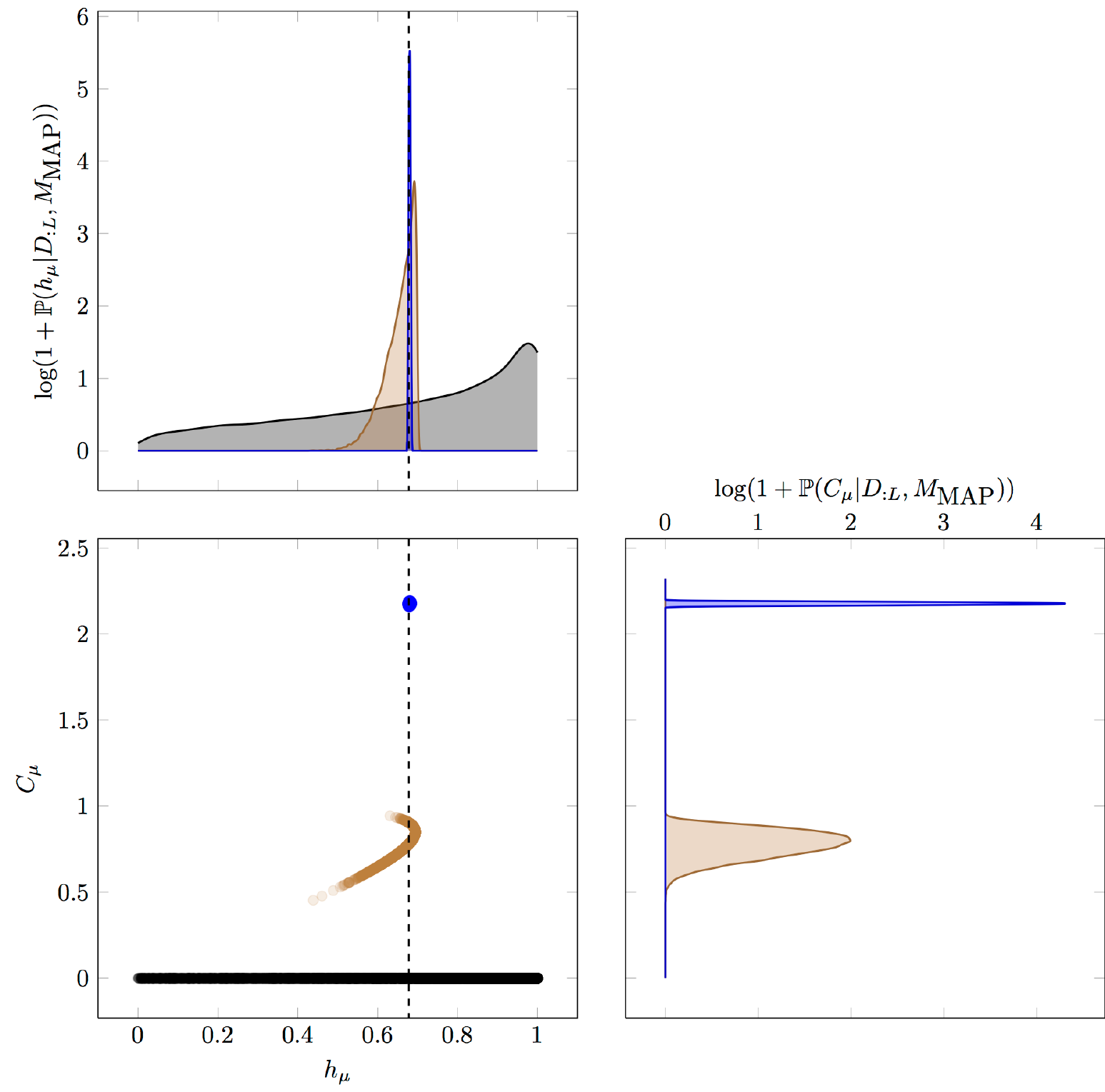}
\caption{Simple Nonunifilar Source: Joint distribution samples using the MAP
  model at the given lengths instead of the full set of candidate models.
  Colors correspond to data subsample length, as in previous plots.  
  The MAP topology for $L=1$ (black) has one state and $\Cmu=0$, as indicated
  by the samples in the $\hmu-\Cmu$ plane. No Gkde approximation of these
  samples is provided due to this complete lack of variation.
  }
\label{fig:sns:hmuCmu:joint}
\end{figure}

\clearpage
\section{Number of Accepting Topologies for Processes}
\label{sec:AcceptTopos}

\clearpage
\section{Maximum a posteriori topologies}
\label{sec:MAPTopos}

\begin{figure}[b]
  \includegraphics[]{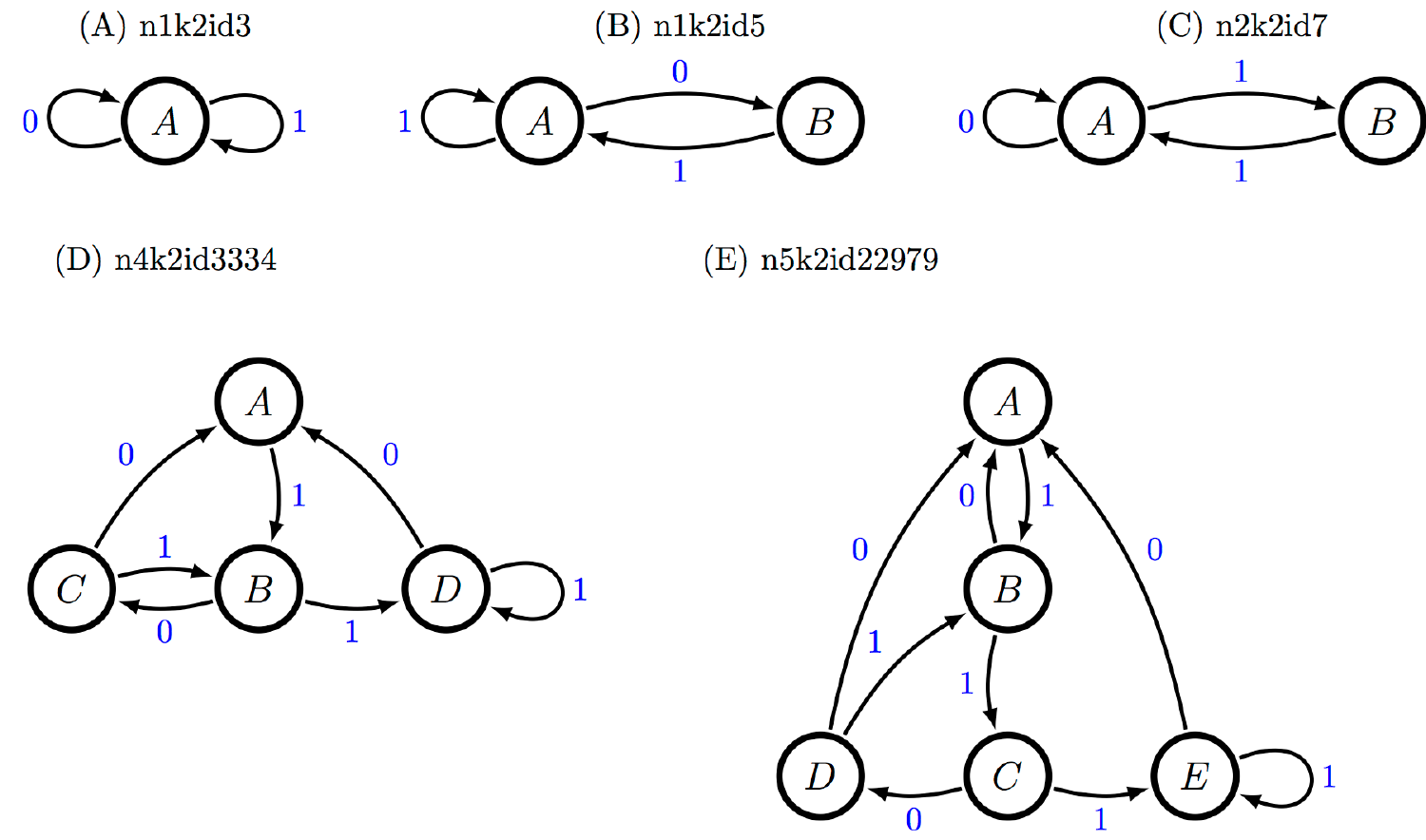}
\caption{\emph{Maximum a posteriori} topologies for the Golden Mean, Even, and
  SNS Process data series. Transitions are only listed with emitted output
  symbol. Transition probabilities are inferred from data for states that 
  have more than one out-going transition. Transitions from states with only
  one out-going arc must have probability one, by definition of the topology.  
  Consult Tables \ref{tab:gm:map}, \ref{tab:even:map}, and \ref{tab:sns:map} to
  see when these structures corresponded to the MAP topologies for the given 
  data sources.
  }
\label{fig:map:topologies}
\end{figure}

Figure \ref{fig:map:topologies} lists all MAP topologies encountered when
inferring \eM\ structure using data from the Even, Golden Mean, and SNS 
Processes. All processes had n1k2id3 (panel A) as the MAP topology for small 
$L$, reflecting a preference for small structures when limited data is 
available. The Golden Mean Process transitioned from n1k2id3 to n2k2id5 
(panel B), the correct structure, at $L=64$, as documented in Table 
\ref{tab:gm:map}.  In a similar manner, the Even Processes changed from n1k2id3 
to n2k2id7 (panel C), the correct structure, at data size $L=32$, as shown in 
Table \ref{tab:even:map}.  The fact that these in-class \eM\ structures 
quickly converge on the correct topology is perhaps expected. Predicting the
sample size at which this occurs, however, is not obvious.

The Simple Nonunifilar Source has a more complicated series of MAP topologies, 
starting with the simple n1k2id3 and progressing through n2k2id5 (panel B), to 
n4k2id3334 (panel D) and, finally, to n5k2id22979 (panel E) at data size
$2^{17}$. Of course, this out-of-class data source cannot be exactly captured
by the set of finite-state unifilar \eMs\ considered here. Nonetheless, we
expect the size of the inferred model to increase if more data from the SNS
were employed and a larger numbers of states were allowed. It is important to
note that the MAP structure in this case has very low posterior probability.
As discussed in the main text, the topology listed is one of five similar
structures with nearly equal posterior probabilities.

%%
%%
%\bibliography{chaos}

%%%
\end{document}